\documentclass{article}

\usepackage{arxiv}

\usepackage{dsfont}
\usepackage{bbold}

\usepackage[utf8]{inputenc} 
\usepackage[T1]{fontenc}    
\usepackage{hyperref}       
\usepackage{url}            
\usepackage{booktabs}       
\usepackage{amsfonts}       
\usepackage{nicefrac}       
\usepackage{microtype}      
\usepackage{lipsum}
\usepackage{graphicx}
\graphicspath{ {./images/} }
\graphicspath{ {./media/} }
\usepackage{caption}
\usepackage{subcaption}

\usepackage{amsmath}
\DeclareMathOperator*{\argmax}{arg\,max}
\DeclareMathOperator*{\argmin}{arg\,min}

\usepackage{amsthm}

\usepackage[bottom]{footmisc}

\makeatletter
\newcommand{\longdash}[1][2em]{%
  \makebox[#1]{$\m@th\smash-\mkern-7mu\cleaders\hbox{$\mkern-2mu\smash-\mkern-2mu$}\hfill\mkern-7mu\smash-$}}
\makeatother
\newcommand{\omitskip}{\kern-\arraycolsep}

\usepackage{mathtools, nccmath}
\DeclarePairedDelimiter{\nint}\lfloor\rceil

\usepackage{braket}

\usepackage{cleveref}

\DeclarePairedDelimiterX\bbraket[2]{\langle}{\rangle}{#1 \delimsize\vert #2}

\usepackage{scalerel,stackengine}
\stackMath
\newcommand\wwidehat[1]{%
\savestack{\tmpbox}{\stretchto{%
  \scaleto{%
    \scalerel*[\widthof{\ensuremath{#1}}]{\kern-.6pt\bigwedge\kern-.6pt}%
    {\rule[-\textheight/2]{1ex}{\textheight}}
  }{\textheight}%
}{0.5ex}}%
\stackon[1pt]{#1}{\tmpbox}%
}
\usepackage{mathrsfs}

\usepackage{diagbox}

\usepackage{makecell}

\usepackage{longtable}

\usepackage[acronym]{glossaries}

\usepackage{fancyvrb}

\usepackage{listings}

\usepackage{changepage}

\usepackage{authblk}

\usepackage{setspace}

\newcommand{\ignore}[1]{}

\makeglossaries

\setacronymstyle{long-short}

\newacronym{rbm}{RBM}{ristricted boltzmann machine}
\newacronym{eca}{ECA}{eigen component analysis}
\newacronym{aeca}{AECA}{approximated eigen component analysis}
\newacronym{veca}{VECA}{vanilla eigen component analysis}
\newacronym{pe}{PE}{pure eigenfeature}
\newacronym{efm}{EFM}{eigenfeature matrix}
\newacronym{ecmm}{ECMM}{eigenfeature-class mapping matrix}
\newacronym{lor}{LoR}{logistic regression}
\newacronym{lir}{LiR}{linear regression}
\newacronym{svm}{SVM}{support vector machine}
\newacronym{ksvm}{KSVM}{kernel support vector machine}
\newacronym{pca}{PCA}{principal component analysis}
\newacronym{tsne}{t-SNE}{t-distributed stochastic neighbor embedding}
\newacronym{knn}{KNN}{k-nearest neighbor}
\newacronym{kmc}{KMC}{k-means clustering}
\newacronym{fnn}{FNN}{fragment neural network}
\newacronym{1d}{1D}{one-dimensional}
\newacronym{2d}{2D}{two-dimensional}
\newacronym{3d}{3D}{three-dimensional}
\newacronym{mse}{MSE}{mean squared error}
\newacronym{keca}{KECA}{kernel eigen component analysis}
\newacronym{ceca}{CECA}{continuous eigen component analysis}
\newacronym{ueca}{UECA}{unsupervised eigen component analysis}
\newacronym{nceca}{NCECA}{nonlinear continuous eigen component analysis}
\newacronym{gan}{GAN}{generative adversarial network}
\newacronym{asd}{ASD}{additive state decomposition}
\newacronym{rbf}{RBF}{radial basis function}
\newacronym{lda}{LDA}{linear discriminative analysis}
\newacronym{qda}{QDA}{quadratic discriminative analysis}
\newacronym{ecan}{ECAN}{eigen component analysis network}
\newacronym{sr}{SR}{softmax regression}
\newacronym{sf}{SF}{softmax function}
\newacronym{ecabgan}{ECAbGAN}{eigen component analysis-based generative adversarial network}
\newacronym{ecanbgan}{ECANbGAN}{eigen component analysis network-based generative adversarial network}
\newacronym{geca}{GECA}{generative eigen component analysis}
\newacronym{gecan}{GECAN}{generative eigen component analysis network}
\newacronym{fcfnn}{FcFNN}{fully connected fragment neural network}
\newacronym{dbfnn}{DbFNN}{degeneracy-based fragment neural network}
\newacronym{dnn}{DNN}{deep neural networks}
\newacronym{cnn}{CNN}{convolutional neural networks}
\newacronym{ica}{ICA}{independent component analysis}
\newacronym{dictl}{DictL}{dictionary learning}
\newacronym{qml}{QML}{quantum machine learning}
\newacronym{rado}{RaDO}{rasing dimension operator}
\newacronym{redo}{ReDO}{reducing dimension operator}
\newacronym{em}{EM}{Expectation-maximization}
\newacronym{pmf}{p.m.f.}{probability mass function}
\newacronym{pdf}{p.d.f.}{probability density function}
\newacronym{relu}{ReLU}{rectified linear unit}

\title{Eigen Component Analysis: A Quantum Theory Incorporated Machine Learning Technique to Find Linearly Maximum Separable Components}

\author[1,2]{Chen Miao}
\author[1,2,3,4]{Shaohua Ma}
\affil[1]{Tsinghua-Berkeley Shenzhen Institute, Tsinghua University, Shenzhen, 518055, China}
\affil[2]{Tsinghua University Shenzhen International Graduate School, Shenzhen, 518055, China}
\affil[3]{Corresponding email: ma.shaohua@sz.tsinghua.edu.cn}
\affil[4]{Lead Contact}
\date{}                     
\begin{document}

\maketitle
\begin{abstract}

For a linear system, the response to a stimulus is often superposed by its responses to other decomposed stimuli. In quantum mechanics, a state is the superposition of multiple eigenstates. Here, by taking advantage of the phase difference, a common feature as we identified in data sets, we propose \gls{eca}, an interpretable linear learning model that incorporates the principle of quantum mechanics into the design of algorithm design for feature extraction, classification, dictionary and deep learning, and adversarial generation, etc. The simulation of \gls{eca}, possessing a measurable \textit{class-label} $\mathcal{H}$, on a classical computer outperforms the existing classical linear models. \Gls{ecan}, a network of concatenated \gls{eca} models, enhances \gls{eca} and gains the potential to be not only integrated with nonlinear models, but also an interface for deep neural networks to implement on a quantum computer, by analogizing a data set as recordings of quantum states. 
Therefore, \gls{eca} and \gls{ecan} promise to expand the feasibility of linear learning models, by adopting the strategy of quantum machine learning to replace heavy nonlinear models with succinct linear operations in tackling complexity.

\end{abstract}




\glsresetall

\section{Introduction}


Machine learning is widely used in areas ranging from chemistry \cite{RN52,RN53,RN54}, biology \cite{RN55,RN56,RN57}, materials \cite{RN2} to medicine \cite{RN58}
\ignore{from the study of graphene \cite{RN2} to stool detection \cite{RN1}}. It has also been used in quantum mechanics \cite{RN3,RN10,RN6,RN5} and quantum chemistry \cite{RN25,RN26}. Quantum mechanics has also inspired many machine learning algorithms \cite{RN4,RN7,RN21,RN16}, which, in turn, facilitate physics growth per se \cite{RN30,RN24}. The entanglement between machine learning and quantum mechanics has started to produce increasing cross-disciplinary breakthrough in physics, chemistry, artificial intelligence and even social sciences \cite{RN51}, and emerged \Gls{qml}, an interdisciplinary field that employs quantum mechanics principles into machine learning. In fact, quantum mechanics share high similarity with machine learning in both of their underlying principles and prediction manners \cite{RN5,RN24}. 

In machine learning, the features of a data set are usually redundant \cite{RN15}. Feature extraction refers to enriching the features of interest and suppressing or discard features out of interest. A number of classical dimension reduction methods has been proposed to learn the similarity or difference among the features of a data set or across multiple data sets.
\Gls{pca} seeks an orthogonal transformation to maximize the variance and separate the data, but it is incapable of exploiting class labels or performing inter-class differentiation. \Gls{lda} takes the advantage of class label to differentiate inter-class data, but it is conditioned on Gaussian distribution. \Gls{tsne} is a nonlinear feature extraction model that finds a low dimensional representation of a high dimensional data, but can suffer false clustering if with low perplexity. 
\gls{ica} decouples a mixed signal into multiple source signals, but it's limited to non-Gaussian data distribution. 

With features extracted, classification can be proceeded in an easier manner. The goal of classification is to assign a class label on a given sample\ignore{\textit{[NeedCitation]}}. If several classes are linearly separable, the classifier is termed linear classifier\ignore{\textit{[NeedCitation]}}. Usually, linear separation refers to identifying one or several hyperplanes to separate the data separate the data. \Gls{lor} finds a hyperplane and converts the distance between a new input and the hyperplane into probability of the data belonging to a class. Taking this one step further, \gls{svm} maximizes the two margins on each side of the hyperplane. \Gls{lda} finds several hyperplanes at once, with each one being similar to the ones found by \gls{lor}. 

Empirically, it's agreed that a linear model is less robust or powerful than a nonlinear model. However, this is not the case for \gls{eca}. By utilizing the linearity of a quantum system that superimpose eigenfeatures (i.e. eigenstates), \gls{eca} functions as a linear model capable to couple with most of the machine learning subjects, including but not limited to classification, generative model, feature extraction, dictionary learning, \gls{dnn}, \gls{cnn} or adversarial generation. 
For example, \Gls{eca} can provide a solution toward image generation by learning from a few coefficient generators, such as normal and uniform distribution. The image generation from known coefficients are analogous to prepare a 'cuisine' by following an established 'recipe'. 

The responses of a linear system to two or more stimuli are superposed. It allows to inspect on the responses to individual stimulus and obtain the overall response to ensembled stimuli by superposition.
The decomposing strategy could also be extended to approximate nonlinear systems, by dividing one composite signal into multiple basis signals in their analysis. A well chosen basis is important in decomposing such a signal. 
For example, the Fourier transform can decompose a signal 
into infinite orthogonal basis signals composed by sine and cosine functions. The overall response is the superposed consequence of responses to individual decomposed signals. In analysis of a basis signal,such as the sinusoid function,
\begin{equation}
f(x) = A\sin(\omega x+\phi), x \in (-2\pi, 2\pi]
\end{equation}
it could be distinguished by its amplitude $A$, frequency $\omega$, and phase shift $\phi$. From a local observer's point of view at $x_0 \in (-2\pi, 2\pi)$, only the amplitude and phase difference could be sensed. Hence, we define two types of separable components, the amplitude component and the phase component. We use \textit{amplitude} and \textit{phase} here to avoid confusion with the concept of \textit{space}, \textit{time} and \textit{frequency} domains in Fourier transform. The amplitude differences is related to \textit{amplitude}, \textit{magnitude}, and \textit{coordinate} differences. 
The phase difference is related to \textit{phase}, \textit{frequency}, \textit{direction} and \textit{eigenvalue}. There exists many classical algorithms to distinguish the coordinate difference or amplitude difference. 




Phase differences, as we surmise, substantially exist among various data sets. Each signal can be decomposed into several phases with varied probabilities. Meanwhile, a phase may be prevalent in some classes but rare in the rest, which suggests differential probability one eigenfeature belonging to each class. Here, we propose \gls{eca}, a quantum theory-based algorithm and focuses on the phase differences. \Gls{eca} identifies the phase differences for a $l$-class data set. Benefiting from \gls{eca}, the tasks such as feature extraction, classification, clustering, and dictionary learning can all be performed in linear models. 

In classical machine learning, the class label in an entire data set follows one Bernoulli or categorical distribution. \Gls{eca} challenges it with a more rational assumption, i.e. $l$ class labels derived from the original class label given by the data set following $l$ independent Bernoulli distributions. In \gls{eca}, first,, a $m$-dimensional data is prepared as a state on $\log(m)$ qubits. Second, measurements on $l$ measurables with $l$ commutative operators are taken on this state in an arbitrary order and the measured results are recorded. Last, optimization is performed on a classical computer based upon the parameters of the operators, probabilities in measurements and ground truth of each prepared state.


We performed classical simulation on a series of data sets, including MNIST data set and two breast cancer data sets, to confirm the speculation that phase difference exists alongside amplitude difference or even prevails in different data sets. 
Some data sets, such as the two aforementioned breast cancer data sets, were failed to be processed by classical linear models, but well separated by \gls{eca}. Apart from data separation, we also proved the significance of \gls{eca} in revealing the hidden features that have been inaccessible by the existing techniques. Finally, we demonstrated the integration of linear and nonlinear models using \gls{ecan}, which implies the broad feasibility of \gls{eca} and \gls{ecan} in line with nonlinear algorithms. 



\section{Results}
    \subsection{Background and eigen component analysis (ECA) mechanism
    }
    
    \begin{figure}
      \centering
      \includegraphics[width=\linewidth]{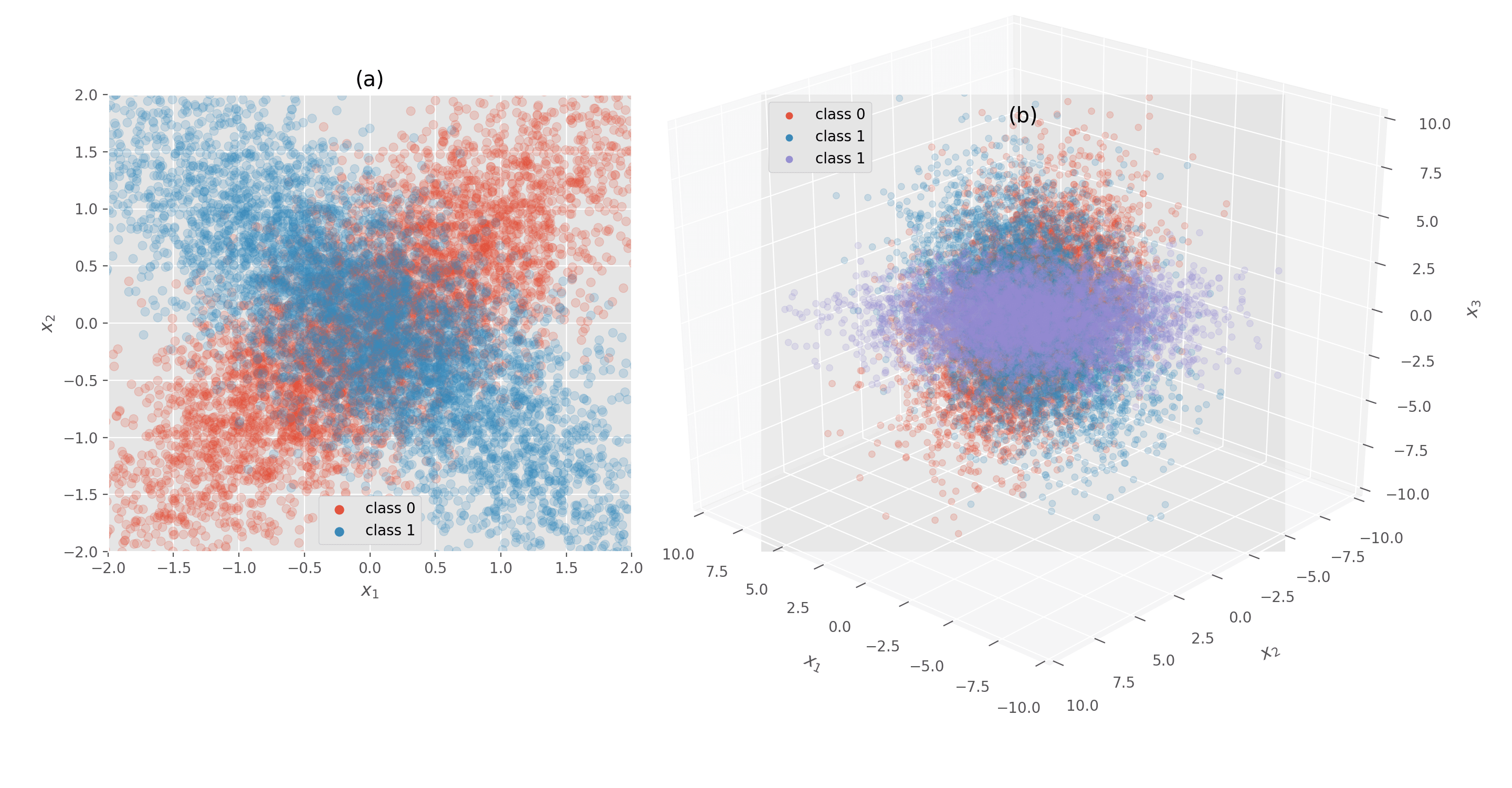}
      \caption{Two artificial intersected data sets. (a) 2D data set. Data in red belong to class $0$ and data in blue belong to class $1$; (b) 3D data set. Data in red belong to class $0$, and data in blue and purple belong to class $1$. }
      \label{fig:2d_3d}
    \end{figure}

    \begin{figure}
      \centering
      \includegraphics[width=\linewidth]{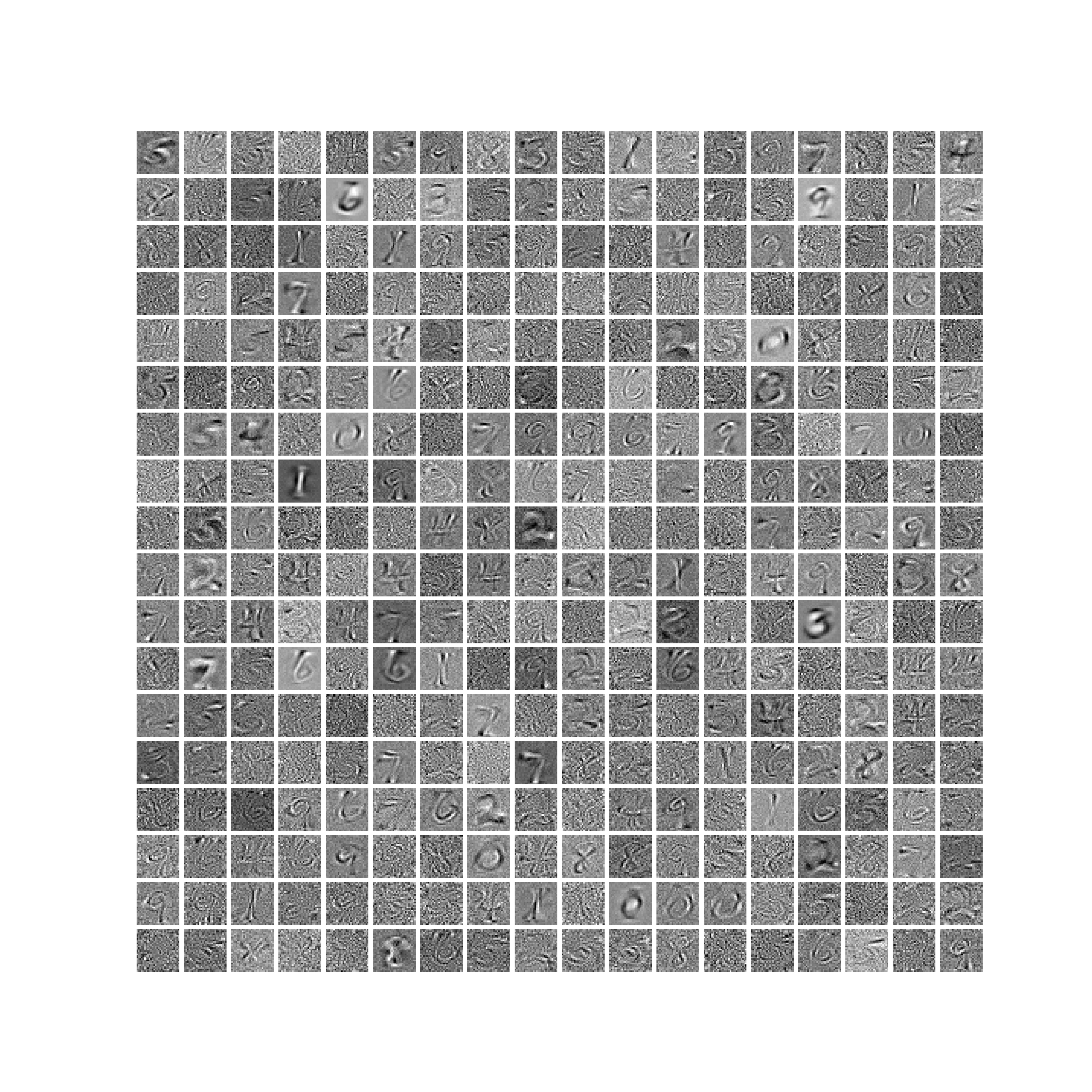}
      \caption{Some 324 pure eigenfeatures (PEs) randomly choosed from all 328 PEs learnt on MNIST data set by approximated eigen component analysis (AECA). The total number of eigenfeatures is 784. }
      \label{fig:mnist_nonoverlap_eigenfeatures}
    \end{figure}

    \begin{figure}
      \centering
      \includegraphics[width=\linewidth]{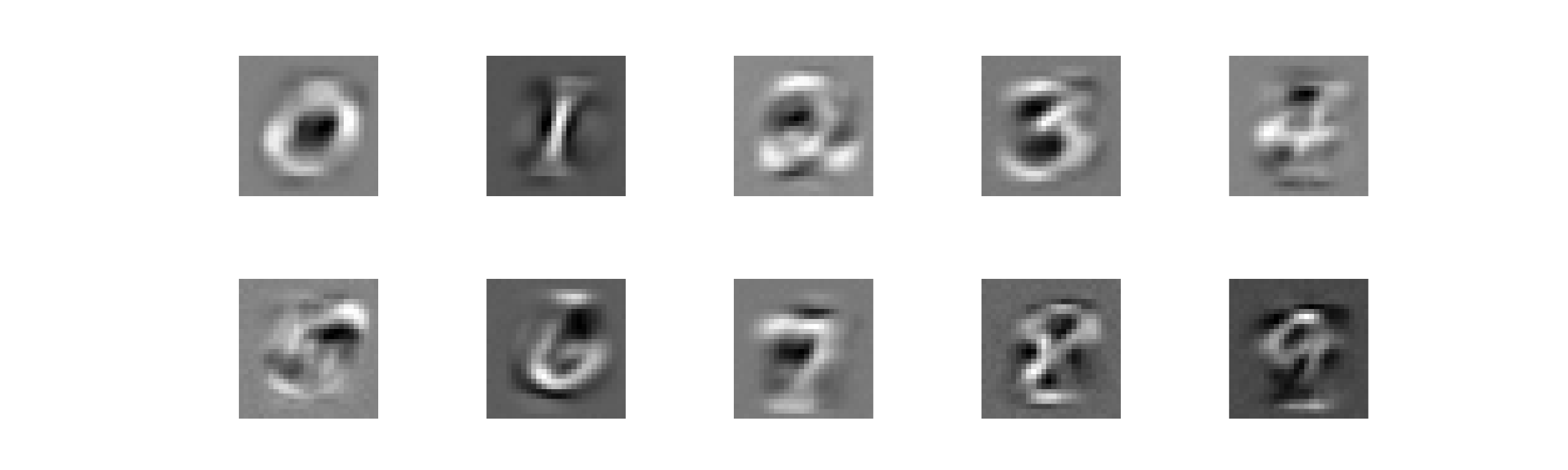}
      \caption{These images are coarsely generated with weighted sum of pure eigenfeatures (PEs) of MNIST data set. For each class, the weight of a PE is the mean projection of all training samples on that PE.}
      \label{fig:mean_weighted_sum_of_pe}
    \end{figure}


    \begin{figure}
      \centering
      \includegraphics[width=\linewidth]{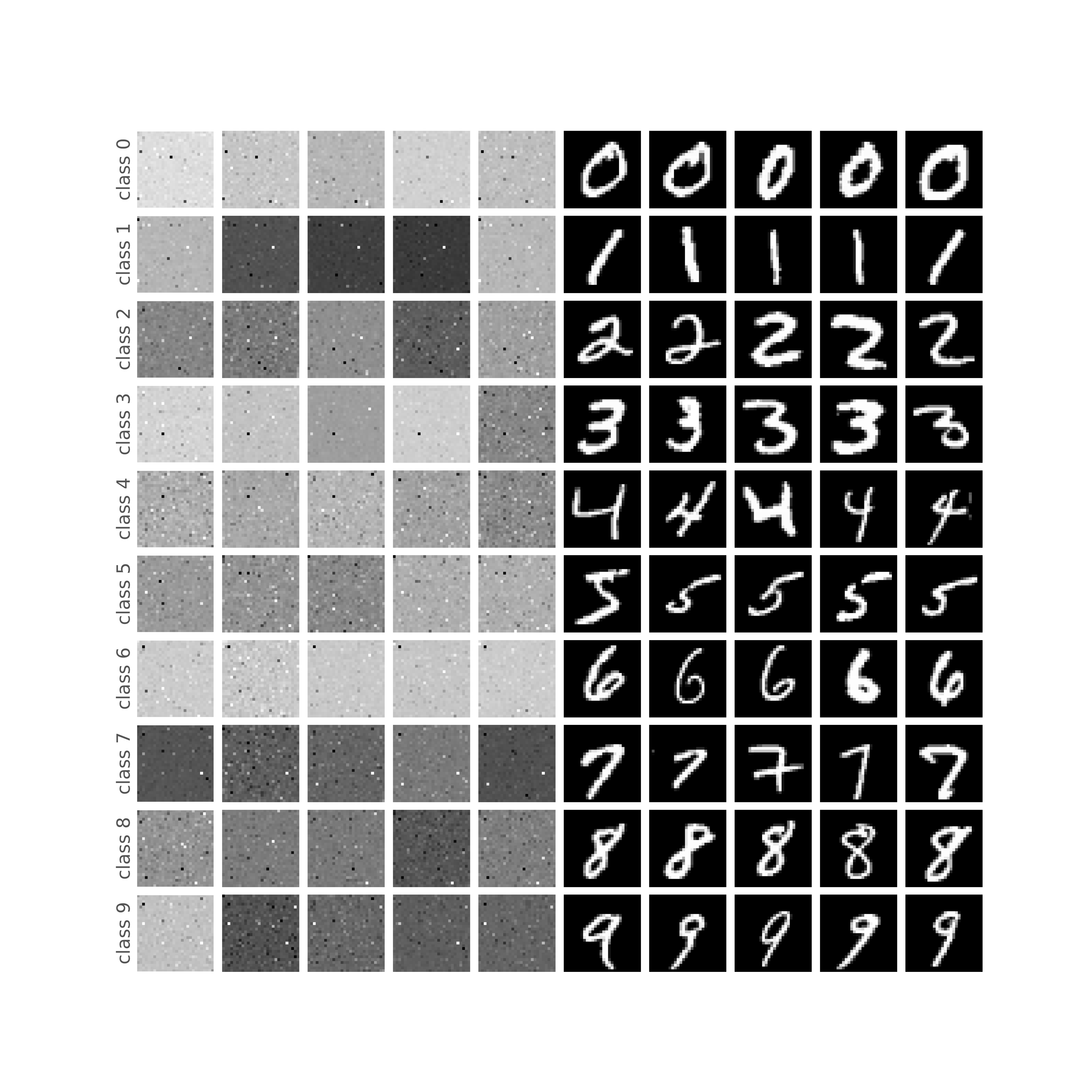}
      \caption{Randomly chosen images from MNIST data set upon basis learnt by approximated eigen component analysis (AECA) and the standard basis. The images on the right half are input images to train AECA. The images on the left are corresponding result of a basis transformation with learnt basis, i.e. eigenfeatures, from right half. The darkest and brightest pixels on the left half are dominant eigenfeatures, which remain constant regardless of digit features on the right.}
      \label{fig:digit_change_of_basis}
    \end{figure}

    First, we clarify the notations used in this paper. A quantum algorithm is intrinsically simple and intuitive, but also abstractive. Its implementation can be simulated on a classical computer. To help understand the concept and verify the simulation, we describe the simulation algorithm in a classical machine learning language, but meanwhile, follow the conventional notations as in quantum mechanics to be consistent with the quantum algorithm, unless specified otherwise. For example, to avoid the confusion caused by using 'observe' and its derivatives, we adopt their meanings as in quantum mechanics in this article. Likewise, the samples in the data set $\{\mathsf{x}^{(n)}\}$ are termed \textit{sample}, \textit{input}, \textit{state}, \textit{recording} or just \textit{vector}. Furthermore, we use notations of convention in real coordinate spaces when no complex numbers are involved. 
    
    It should be paid attention that in quantum algorithms, the eigenvalues of measurable for a qubit or a composite system, are defined as $-1$ and $+1$, representing 'false' and 'true' that if an input or eigenfeature belongs to a class. In classical simulation, the counterparts of these eigenvalues, indicating the class label of an eigenfeature or input vector, are defined as $0$ and $1$. Moreover, the term 'class label' could mean both the original class label given by the data set and the derived class label that if an input belongs to one class. For example, the class label '3' of a sample in a 10-class data set could derive ten class labels, which are '+1' for being a sample from class '3' and '-1' for the corresponding input belonging to classes other than '3'. 
    
    All the sets used in this paper are $0$-indexed. The values $i$, $j$, $k$ index the sample data set (with size $n$), input vector (with size $m$), and class label (with size $l$), respectively. $\{\mathsf{\mathring{x}}^{(i)}\}$ refers to a data set with $n$ samples, where $i=0,1,\cdots,n-1$, together with a finite set $\mathcal{C}$ of class labels $\{k, k=0,1,...,l-1\}$. The corresponding target values compose the set $\{y^{(i)}\}$. Notation $\mathring{(\cdot)}$ refers to an initially non-normalized vector. The data set is then normalized to $\{(\mathsf{x}^{(i)}, y^{(i)}), i=0,1,\cdots,n-1\}$. All the vector in \gls{veca} are normalized and the magnitude information are discarded if without notation $\mathring{(\cdot)}$ or unspecified. The normalized data set is the recordings of states and their measured values. We also denote one indicator function $1\{y=y^{(i)}\}$ and one-hot vector function $\mathsf{y}^{(i)}=\mathbb{1}\{y=y^{(i)}\}$, where $i=0,1,\cdots,n-1$. $\mathsf{y}^{(i)}$ is a $l\times 1$ vector with its $y^{(i)}$-th element being $1$ and otherwise $0$. In addition to that, we denote a one-hot matrix of stacked Bernoulli one-hot vector as 
    
    \begin{equation}
    \mathbb{y}^{(i)} 
    = 
    \begin{bmatrix}
    \mathsf{y}^{(i)} & \widetilde{\mathsf{y}^{(i)}} \\
    \end{bmatrix}
    =
    \begin{bmatrix}
    \mathbb{1}^T\{\mathsf{y}_0=\mathsf{y}_0^{(i)}\} \\
    \mathbb{1}^T\{\mathsf{y}_1=\mathsf{y}_1^{(i)}\} \\
    \vdots \\
    \mathbb{1}^T\{\mathsf{y}_{l-1}=\mathsf{y}_{l-1}^{(i)}\} \\
    \end{bmatrix}_{l \times 2}
    \end{equation}
    
    where the operator $\widetilde{\mathsf{y}}$ takes the one's complement of each element in $\mathsf{y}$. In the classical simulation, for discrete \gls{eca}, i.e. the observed values are discrete, the ket-vectors (or kets) $\ket{\mathsf{x}}$ is the same as $\mathsf{x}$. The probability $p(y|\ket{\mathsf{x}})$ without a superscript specified on $y$ means a vector representation of its \gls{pmf}. The bold font $\mathbf{p}$ indicates a vector of stacked probabilities of independent random variables. The outline font $\mathbb{p}$ indicates a matrix of stacked \gls{pmf} of independent Bernoulli random variables. The element $\mathsf{y}_k$ of $\mathsf{y}$ is a Bernoulli random variable if without a superscript. 
    Thus, with a superscript, $\mathsf{y}^{(i)}$ is one-hot vector, yet without superscript, it is a vector of stacked independent random variables. 
    For some situation, usually in general discussion, the superscript of a numerical value like $y^{(i)}$, $\mathsf{y}^{(i)}$, or $\mathsf{x}^{(i)}$ is omitted for simplicity, when it could be inferred from the context . 
    The subscript and superscript are omitted when there is no risk of ambiguity in the rest of the paper.
    
    In quantum mechanics, a vector representation of an object or state is 'measurable' as long as we know the measurable. We could also predict these measurements, once we know the mathematical expression of the measurable and its state. Likewise, we could abstract a real world state or object, such as a spin or image, as a vector, and construct its measurable, no matter it's momentum or \textit{class-label}.
    
    
    As we addressed, samples in a data set have two types of variance. One is amplitude-based and the other one is phase-based. 
    Our \gls{eca} focuses on identifying the phase-based differences of a data set.
    
    
    

    For a vector, in general, a linear classifier or even some kernel-based classifiers treat each element of the vector as one feature. However, these 'features' may not fully represent the real property of the data. For a vector possessing more complicated structures, all the elements in the vector become necessary to constitute a reliable feature. In other words, it's a choice to express the data on a well-defined basis (see \Cref{fig:2d_3d} (a)) or a standard basis (see \Cref{fig:2d_3d} (b)). Therefore, for a vector $\mathsf{x}$ 
    
    \begin{equation}
    \mathsf{x} =
    \begin{bmatrix}
    x_0 \\
    x_1 \\
    \vdots \\
    x_{m-1} \\
    \end{bmatrix}, 
    \end{equation}
    
    each $x_{i}\;(i=0,1,\cdots,m-1)$ is viewed as a feature of this vector upon a standard basis. 
    For some complicated structures (e.g. the edges of an object in an image) , one single element of $\mathsf{x}$ i.e. a vector in a standard basis, cannot tell the whole story. If we finds an orthogonal basis (i.e. eigenfeatures, see \Cref{fig:mnist_nonoverlap_eigenfeatures}) $\mathcal{B}$, all the vectors could be a unique linear combination of vectors in this basis (see \Cref{fig:mean_weighted_sum_of_pe}). If $\mathcal{B} = \{\ket{\lambda_0}, \ket{\lambda_1},\cdots,\ket{\lambda_{m-1}}\}$, then 
    
    \begin{equation}
    \mathsf{x} = \psi_0 \ket{\lambda_0} + \psi_1 \ket{\lambda_1} + \cdots + \psi_{m-1} \ket{\lambda_{m-1}}
    \end{equation}
    
    in which $\psi_j$ is nomalized coefficients (see \Cref{fig:digit_change_of_basis}), with
    \begin{equation}
    \sum_j \braket{\psi_j|\psi_j} = 1.
    \end{equation}
    
    In quantum mechanics, the state x would collapse on eigenstate $\ket{\lambda_j}$ $(j=0,1,\cdots,m-1)$ with the probability
    
    \begin{equation}
    p_j = \braket{\psi_j|\psi_j}.
    \end{equation}

\begin{figure}
      \centering
      \includegraphics[width=\linewidth]{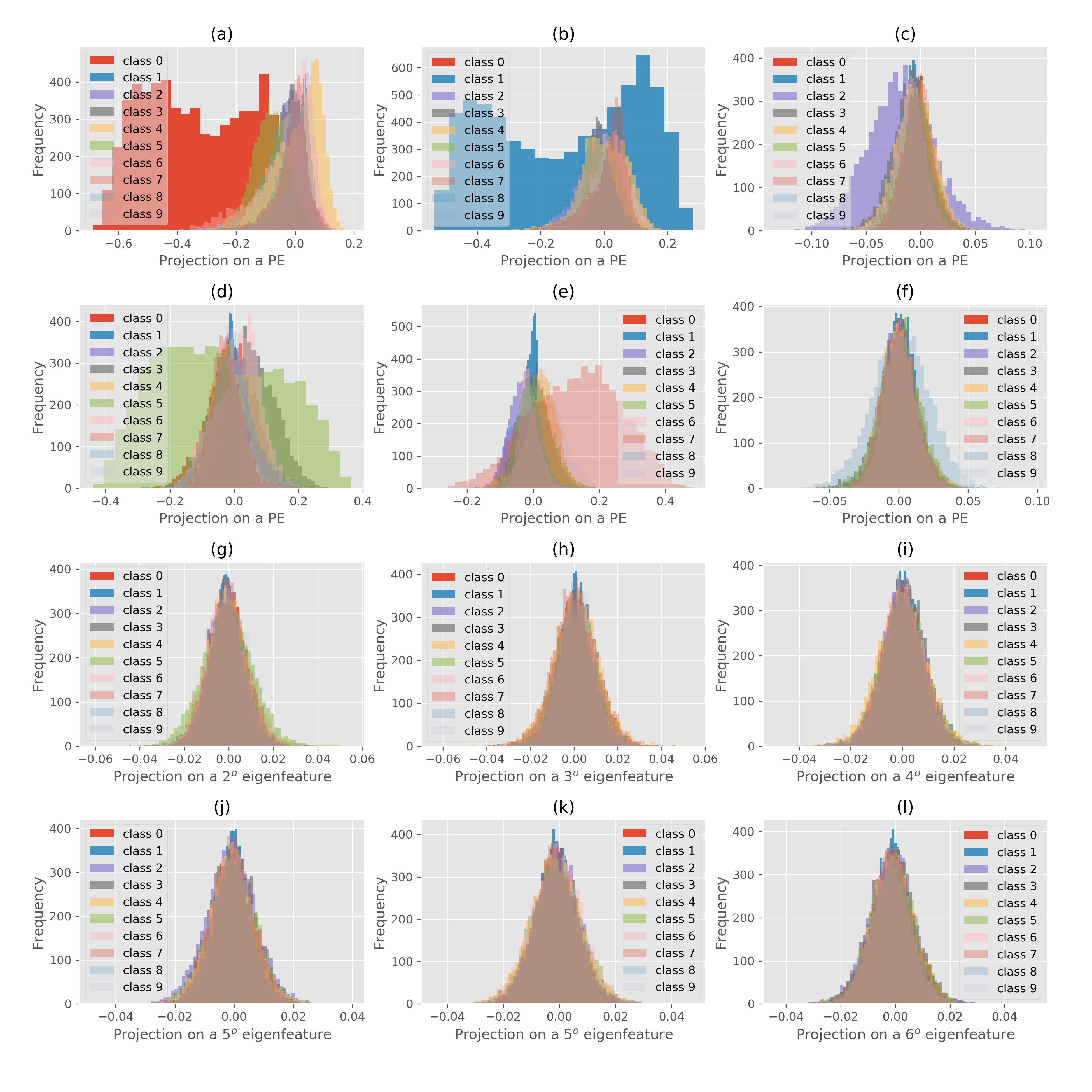}
      \caption{The distribution of projections of all training samples on $12$ eigenfeatures selected from eigenfeatures learnt on MNIST data set using approximated eigen component analysis (AECA). The superscript $k^o$ denotes $k$ degree of overlapping, which means one eigenfeature has been shared by $k$ classes. (a) - (f) Frequency distribution of normalized projections on pure eigenfeatures (PEs) corresponding to the class label '0', '1', '2', '5', '7', and '8' sequentially. That PEs used in (a) - (c) are corresponding to the least degenerated and these in (d) - (f) are corresponding to the most degenerated. (g) - (i) Frequency distributions of randomly choosed $2^o$, $3^o$ and $4^o$ eigenfeatures in turn. (j) - (l) Frequency distributions of the most overlapped three eigenfeatures with $5^o$ for eigenfeature in (g) and (h) and $6^o$ for (i). }
      \label{fig:mnist_proj_dist_ef}
    \end{figure}

    Not only the input vector but also its class label have quantum interpretation. Classical machine learning algorithms usually assume the class labels of input or basis vectors following Bernoulli or categorical distribution, which is true for the eigenvectors in \Cref{fig:mnist_proj_dist_ef} (a)-(f). 
    For a $2$-class data set ( \Cref{fig:2d_3d} (a)), the probabilities of the predictions could be described by a probability matrix
    
    \begin{equation}
    \rho = 
    \begin{bmatrix}
    \rho_0(1-\rho_1) & \rho_0\rho_1 \\
    (1-\rho_0)(1-\rho_1)  & \rho_1(1-\rho_0)\\
    \end{bmatrix}
    \end{equation}
    
    
    in which $\rho_0$ and $\rho_1$ are the probability of an input vector belonging to class '0' and class '1', respectively. Under the classical assumption, the trace of $\rho$ equals $1$, i.e.
    
    \begin{equation}
    Tr(\rho)=1.
    \end{equation}

    If a new input vector unambiguously belongs to one class, then either $p_0$ or $p_1$ equals 1, reaching
    
    \begin{equation}
    \rho^2=\rho.
    \end{equation}
    
    
    
    However, this could be wrong. For the data points in the center of \Cref{fig:2d_3d} (a), it could belong to either class. Meanwhile, many data points in the \gls{2d} space do not belong to any of these two classes. A data set usually occupies a compact space or spans a subspace in the full vector space. For a $l$-class data set, the total number of decisions in the full space could reach $2^l$, but the mutual exclusive decision is only a small subset of size $l$. If we assume that each input vector as well as its basis vectors have considerable possibilities belonging to one or more classes (\Cref{fig:mnist_proj_dist_ef} (g)-(l)), we could choose to inspect each class independently. If we prepare a new input vector as a quantum state, it's assumed an apparatus could be constructed to measure the class label indicating whether an input vector belongs to a to-be-decided class. For the $l$ classes, 
    $l$ sized apparatuses could be built to take measurements on $l$ identical copies of the state. Second, an operator on a $m$-dimensional state vector has $m$ eigenvalues, with each eigenvalue has two possible outcomes, $-1$ ('false') and $+1$ ('true'), to represent if it belongs to a specific class. The candidates of all these eigenvalues of each operator should be arranged to degenerate to $-1$ and $+1$. Taking all classes into account, we should assume all these corresponding operators of these measurables share a complete basis of simultaneous eigenvectors. 
    Hence, for a data set with $l$ classes and $m$-dimensional features, the task of degenerating the data set into two distinctive states is converted into a measurement on $l$ independent systems with $l\log(m)$ qubits in total. 
    In these measurements, for each measurable, the measurement on the whole system could be product of observed values taken on each qubit. For more concurrency, $l \log(m)$ qubits could be used.
    As these operators commute, all these operators share a complete basis of simultaneous eigenvectors but has its own eigenvalues. 
    
    Therefore, in classical simulation, instead of learning which class a vector belongs to, we learn which classes each eigenfeature of a vector belongs to. Furthermore, the decision-making should be conducted independently on each class. Next, we have a mapping table between eigenfeatures and class labels. This leads us to learning an \gls{efm} representing the unitary operator with a complete basis of simultaneous eigenvectors and \gls{ecmm}. \gls{ecmm} bridges the superposition of the probabilities of the class label of each eigenfeature and the class label of the input vector. 
    All we need is to sum up the probabilities of each eigenfeature assigned for all classes independently. Afterward, we could obtain the combined probabilities one vector belonging to each class. For $l$ classes, the probabilities of an input vector belonging to each class is $p_0, p_1, \cdots, p_{l-1}$. The mutual exclusive probability for class 'c' is calculated by 
    
    \begin{equation*}
    p(y^{(i)}=c|\mathsf{x}; \theta)
    = \prod_{k \neq c} (1-p_k)p_c.
    \end{equation*}

    The unitarity of \gls{efm} could guarantee that the difference is kept in a change of basis transformation. For \gls{efm}, the variance of projection on eigenfeature is maximized (see \Cref{fig:digit_change_of_basis}). In the left half of \Cref{fig:digit_change_of_basis}, the bright and dark pixels indicate significant signals, whereas the gray pixels are trivial, enabling for dimension reduction. Also, the stable positioning of these bright and dark pixels among different inputs suggests it being appropriate as a classifier. 
    
    For prediction, we prove that an independent decision can be made without calculating the mutual exclusive probabilities for a data set with two classes. The proof can be easily extended to data sets with multiple classes using mathematical induction. For a given input, the two mutual exclusive probabilities of each class label are 
    
    \begin{equation*}
    p(
    \begin{bmatrix}
    0 \\
    1
    \end{bmatrix}
    |\mathsf{x}^{(i)}; \theta)
    = p_0(1-p_1)
    \end{equation*}
    
    and 
    
    \begin{equation*}
    p(
    \begin{bmatrix}
    1 \\
    0
    \end{bmatrix}
    |\mathsf{x}^{(i)}; \theta)
    = p_1(1-p_0)
    \end{equation*}
    
    Then the proof is given as 
    \begin{proof}
    Without loss of generality, suppose $p_0 > p_1$, such that 
    \begin{equation*}
    \begin{split}
    & p(
    \begin{bmatrix}
    0 \\
    1
    \end{bmatrix}
    |\mathsf{x}^{(i)}; \theta)
    - 
    p(
    \begin{bmatrix}
    1 \\
    0
    \end{bmatrix}
    |\mathsf{x}^{(i)}; \theta) 
    = p_0(1-p_1) - p_1(1-p_0) 
    =  p_0 - p_1 > 0 
    \Rightarrow 
    p(
    \begin{bmatrix}
    0 \\
    1
    \end{bmatrix}
    |\mathsf{x}^{(i)}; \theta)
    >
    p(
    \begin{bmatrix}
    1 \\
    0
    \end{bmatrix}
    |\mathsf{x}^{(i)}; \theta)
    \end{split}
    \end{equation*}
    \end{proof}
    
    Further, we can build a $\tau$-fold \gls{ecan} by concatenating $\tau$ \gls{eca} models. \Gls{ecan} gives \gls{eca} the capability to integrate nonlinear models such as \glspl{dnn}. A dimension operator assuming the nonlinearity, which can be specially designed or a classical \gls{dnn}, is installed between consequent \gls{eca} models. 
    
    \subsection{Related work}
    
    
      The related work include several classical algorithms like \gls{ica}, \gls{dictl}, and also \gls{qml} algorithms. \ignore{Some \gls{qml} algorithms might be too complicated even we cannot fully understand what they are talking about. Our method might be the simplest might-be-complicated \gls{qml} algorithms. }
    
        \subsubsection{Independent component analysis (ICA)}
        \gls{ica} shares the similar goal as \gls{eca}. Both algorithms try to find independent components which could generate the data from some independent sources. \gls{ica} decouples a mixed signal by multiple recorders depending on the varied combination of source signals. The recorders can be regarded as another kind of label because they record the intensity of the source variably. Thus, \gls{ica} could be replaced by \gls{eca} as \gls{eca} is dependent on the most intensive source. The major advantage of \gls{eca} over \gls{ica} is that there is no necessary for prior distribution assumption. 
    
        \subsubsection{Dictionary learning (DictL)}
        \Gls{dictl} is similar to \gls{eca} because they both want to find a sparse representation of the data set. A supervised dictionary learning method presented in \cite{NIPS2008_3448} is like a second cousin of \gls{eca} in appearance. A discriminative task is added to the objective while the reconstruction term is reserved. Likewise, in \gls{veca}, our objective is also to identify the independent eigenfeatures, based on which the data classification is conducted. In comparison, \gls{eca} is easier to train because it comprises less hyperparameters. Moreover, \gls{eca} is less likely for loss of information because it takes all classes and a complete profile of basis into account and preserves the difference. 
    
        \subsubsection{Quantum machine learning (QML)}
        \Gls{qml} is a wide range of machine learning algorithms including quantum computation-based machine learning or inspired and facilitated by quantum mechanics. The method presented in \cite{RN21} assumes a prior for the input feature states and obtains a number of template classes. However, the identities of these template states are ignored. In their method, the need to find some template states which are all linear combination of some pure quantum states. In contrast, in our \gls{eca}, all the inputs are linear combinations of pure quantum states, the identities of which are utilized for further pursuit.
    
    
    \subsection{Preliminary performance test of eigen component analysis (ECA)}
    \label{example}
    Before moving forward to the algorithm, we prepared an example that illustrated two ideal cases with two artificial data sets (see \Cref{fig:2d_3d}). To birth some epiphanies, this informal discussion is based upon \textit{guess} and intuition. 

    Two data sets showed in \Cref{fig:2d_3d}. The data of each class intersected with each other. 
    For the \gls{2d} and \gls{3d} data set , we could \textit{guess} an \gls{efm} (which is an unitary operator) $\overset{\asymp}{P}\in \mathbb{C}^{m\times m}$ of a linear seprator $\overset{\asymp}{H}$ with eigenfeature as column vector and \gls{ecmm} $\overset{\asymp}{\mathfrak{L}} \in \mathbb{R}^{m\times l}$ (of which the elements $\overset{\asymp}{\mathfrak{L}}_{jk} \in \{0,1\}$) which are

    \begin{equation*}
        \overset{\asymp}{P}_{2D} = 
        \begin{bmatrix}
        \frac{1}{\sqrt{2}} & \frac{-1}{\sqrt{2}} \\
        \frac{1}{\sqrt{2}} & \frac{1}{\sqrt{2}} \\
        \end{bmatrix}
        \quad
        \overset{\asymp}{\mathfrak{L}}_{2D} = 
        \begin{bmatrix}
        1 & 0 \\
        0 & 1
        \end{bmatrix},
    \end{equation*}

    and

    \begin{equation*}
        \overset{\asymp}{P}_{3D} = 
        \begin{bmatrix}
        1 & 0 & 0 \\
        0 & 1 & 0 \\
        0 & 0 & 1 \\
        \end{bmatrix}
        \quad
        \overset{\asymp}{\mathfrak{L}}_{3D} = 
        \begin{bmatrix}
        0 & 1 \\
        1 & 0 \\
        1 & 1 \\
        \end{bmatrix}
        \text{ or }
        \overset{\asymp}{\mathfrak{L}}_{3D}^{sparse} = 
        \begin{bmatrix}
        0 & 1 \\
        1 & 0 \\
        0 & 0 \\
        \end{bmatrix}
    \end{equation*}

    respectively, in which the symbol $\overset{\asymp}{\mathfrak{(\cdot)}}$ indicates a numerical estimation and $\overset{\asymp}{\mathfrak{L}}_{3D}^{sparse}$ is an equivalent sparse representation of $\overset{\asymp}{\mathfrak{L}}_{3D}$. For the \gls{3d} data set, each columns of \gls{efm} represent a pivot axis or principal component of the data set. The element $\overset{\asymp}{\mathfrak{L}}_{3D}^{00}=0$ means the 0th eigenfeature of this data set doesn't belong to class '0' and $\overset{\asymp}{\mathfrak{L}}_{3D}^{10}=1$ indicates the 1st eigenfeature belongs to class '0'. The two 1s in the 2nd (0-indexed) row of $\overset{\asymp}{\mathfrak{L}}_{3D}$ represents that the 2nd eigenfeature could be noise or background shared among the two classes. For a new input, we only need to sum up the probabilities that the input projects on the 1st and 2nd eigenfeature to decide the probability whether this input belongs to class '0'. One should notice the decision of class label of a input is not mutual exclusive because the summation of each class could be equal and the sum of the two probabilities could surpass 1 as they are independent decision.

    We give a more concrete development for this informal discussion. For vector $\mathring{\mathsf{x}}$ in aforementioned 3D data set, 

    \begin{equation}
    (\wwidehat{\mathring{\mathsf{x}}}^T\overset{\asymp}{P})(\wwidehat{\mathring{\mathsf{x}}}^T\overset{\asymp}{P})^T=1
    \end{equation}

    in which $\hat{(\cdot)}$ is the nomalization operator. Hence we denote the elements of 

    \begin{equation}
    \psi = \overset{\asymp}{P}^T\wwidehat{\mathring{\mathsf{x}}}
    \end{equation}

    as $\psi_j$ and we have 

    \begin{equation}
    \sum_{j=0}^{m-1} \psi_j^{2} = 1.
    \end{equation}

    For the aforementioned \gls{3d} data set, instead of treating the class label of a vector as a single categorical distribution, the \gls{pmf} of each class label $c$ given $\mathsf{x}$ is assumed to follow a independent Bernoulli distribution, such that the probability of one vector belongs to class '0' is

    \begin{equation}
    p_0(\mathsf{y}_0=1|\mathsf{x}^{(i)}; \theta) = [(\mathsf{x}^{(i)T} \overset{\asymp}{P}_{3D}) \odot (\mathsf{x}^{(i)T} \overset{\asymp}{P}_{3D})] (\overset{\asymp}{\mathfrak{L}}_{3D})_{\bullet, 0}
    \end{equation}

    in which  $\theta$ is the all unknown parameters, $\odot$ is element-wise Hadamard product operator, $\bullet$ is a placeholder for taking the entire column (as 0th index) or row (as 1st index) and $(\overset{\asymp}{\mathfrak{L}}_{3D})_{\bullet, 0}$ means the 0-th column of $\overset{\asymp}{\mathfrak{L}}_{3D}$. Thus, the combined probabilities of these two Bernoulli random variables with observed event as $\mathbb{1}_{2 \times 1}$ could be defined as 

    \begin{equation}
    \mathbf{p}(\mathsf{y}=\mathbb{1}_{2 \times 1}|\mathsf{x}^{(i)}; \theta) 
    = 
    \begin{bmatrix}
    p_0 \\
    p_1
    \end{bmatrix}
    = \{ [(\mathsf{x}^{(i)T} \overset{\asymp}{P}_{3D}) \odot (\mathsf{x}^{(i)T} \overset{\asymp}{P}_{3D})] (\overset{\asymp}{\mathfrak{L}}_{3D}) \}^T
    \end{equation}

    and the complement probabilities could be

    \begin{equation}
    \begin{split}
    & \mathbf{p}(\mathsf{y}=\mathbb{0}_{2 \times 1}|\mathsf{x}^{(i)}; \theta) 
    = \mathbb{1}_{2 \times 1} - \mathbf{p}(\mathsf{y}_0^{(i)}|\mathsf{x}^{(i)}; \theta) 
    = 
    \begin{bmatrix}
    1 - p_0 \\
    1 - p_1
    \end{bmatrix} 
    = \{ [(\mathsf{x}^{(i)T} \overset{\asymp}{P}_{3D}) \odot (\mathsf{x}^{(i)T} \overset{\asymp}{P}_{3D})] (\mathbb{1}_{3 \times 2} - \overset{\asymp}{\mathfrak{L}}_{3D}) \}^T
    \end{split}
    \end{equation}

    in which the outline font $\mathbb{0, 1}$ are vector or matrix with corresponding digits. 

    For simplicity, we could write these two \gls{pmf} of Bernoulli random variable $\mathsf{y}_0$ and $\mathsf{y}_1$ together. With the feature matrix and the mapping matrix, these stacked or combined \gls{pmf} of combined Bernoulli random variable $\mathsf{y}_k (k=0,1)$  given $\mathsf{x}$ (i.e. $\mathsf{y}_k|\mathsf{x}^{(i)}$, $i=0,1,\cdots,n-1$) could be written in 

    \begin{equation}
    \begin{split}
    \mathbb{p}(\mathsf{y}|\mathsf{x}^{(i)}; \theta)
     = &
    \begin{bmatrix} 
    \mathbf{p} &
     \mathbb{1}_{2 \times 1} - \mathbf{p}
    \end{bmatrix}
    \end{split}
    \end{equation}

    where the \textit{rows} are the vectors of corresponding \gls{pmf}. 




    To obtain the mutual exclusive decision on the class label, the unambiguous probability of $\mathsf{x}$ belonging to class '0' could be calculated as 

    \begin{equation}
    p_0(\mathsf{x}\textit{ belongs to class '0' and not '1'}|\mathsf{x}^{(i)}; \theta) 
    =p(
    \begin{bmatrix}
    0 \\
    1
    \end{bmatrix}
    |\mathsf{x}^{(i)}; \theta)
    = \mathbb{p}_{00}\mathbb{p}_{11}
    = \mathbf{p}_0 (1 - \mathbf{p}_1)
    = p_0 (1 - p_1)
    \end{equation}


    
    \subsection{Experiment results}
    
    We compared our model with \gls{lor}, \gls{lda}, \gls{qda}, \gls{svm}, \gls{ksvm} with a \gls{rbf} kernel.

      \subsubsection{Counting the parameters}
    
      For a data set with $m$ features and $l$ classes, the number of total parameters could be calculated as:
      \begin{itemize}
        \item[LoR] {$m+1$}
        \item[LDA] {$lm + (l - 1)$}
        \item[QDA] {$lm + (l-1) + l\frac{m(m+1)}{2}$}
        \item[SVM] {$m+1$}
        \item[ECA] {$lm + {m(m+1)}$}.  
      \end{itemize}
    
      We don't count the parameters of \gls{ksvm} in these experiments.
    
      \subsubsection{Two artificial data sets (2D and 3D)}
    
      \begin{itemize}
        \begin{table}[!ht]
         \caption{Comparison with other classifiers of 2D data set}
          \centering
          \begin{tabular}{|c|c|c|c|}
            \toprule
            & \multicolumn{2}{c|}{Metrics} & \\
            \cmidrule(r){2-3}
            Name & Accuracy & Confustion Matrix & Parameters \\
            \midrule
            LoR  & 0.5242    & $\bigl( \begin{matrix}0.5817 & 0.5313\\0.4183 & 0.4687\end{matrix}\bigr)$ & 3\\
            \midrule
            LDA & 0.5239    & $\bigl( \begin{matrix}0.5811 & 0.5313\\0.4189 & 0.4687\end{matrix}\bigr)$ & 5\\
            \midrule
            QDA & 0.8124    & $\bigl( \begin{matrix}0.8229 & 0.1977\\0.1771 & 0.8023\end{matrix}\bigr)$ & 11\\
            \midrule
            SVM & 0.6048    & $\bigl( \begin{matrix}0.8168 & 0.5998\\0.1832 & 0.4002\end{matrix}\bigr)$ & 3\\
            \midrule
            KSVM & 0.8063    & $\bigl( \begin{matrix}0.802 & 0.1894\\0.198 & 0.8106\end{matrix}\bigr)$ & \diagbox[dir=NW,width=6em,height=2.5em]{\;}{\;}\\
            \midrule
            ECA & 0.8139 & $\bigl( \begin{matrix}0.8192 & 0.1912\\0.1808 & 0.8088\end{matrix}\bigr)$ & 8\\
            \bottomrule
          \end{tabular}
          \label{tab:2d}
        \end{table}
    
        \item[2D] One class of the \gls{2d} data set (\Cref{fig:2d_3d}) is random normally generated with mean of $\left(\begin{matrix}0 & 0\end{matrix}\right)^T$ and covariance matrix $\left(\begin{matrix}1 & 0.8 \\0.8 & 1\end{matrix}\right)$; and the another class is gengerated with  mean $\left(\begin{matrix}0 & 0\end{matrix}\right)^T$ and covariance matrix $\left(\begin{matrix}1 & -0.8 \\-0.8 & 1\end{matrix}\right)$. 
    
        The $P$ and $\dot{L}$ obtained by \gls{eca} is 
    
        $$
        P_{2D} = 
        \begin{bmatrix}
        -0.7199738 & -0.7047859 \\
        -0.6945969 & 0.70967793 \\
        \end{bmatrix}
        \text{ and }
        \dot{L}_{2D} = 
        \begin{bmatrix}
        1.0000000e+00 & 1.0823610e-20 \\
        1.6251015e-20 & 1.0000000e+00 \\
        \end{bmatrix}
        $$
    
        respectively, with which the model could obtain an accuracy of $81.39\%$ on the validation data which is on par with \gls{qda} and outperforms the rest (\Cref{tab:2d}).

        \item[3D] One class of \gls{3d} data set (\Cref{fig:2d_3d} (b)) is random normally generated with mean of $\left(\begin{matrix}0 & 0 & 0\end{matrix}\right)^T$ and covariance matrix $\left(\begin{matrix}0.10 & 0 & 0 \\0 & 10 & 0\\0 & 0 & 10\end{matrix}\right)$; and another one is random normally generated with mean $\left(\begin{matrix}0 & 0 & 0\end{matrix}\right)^T$ and covariance matrix $\left(\begin{matrix}10 & 0 & 0 \\0 & 0.1 & 0\\0 & 0 & 10\end{matrix}\right)$ and mean $\left(\begin{matrix}0 & 0 & 0\end{matrix}\right)^T$ and covariance matrix $\left(\begin{matrix}10 & 0 & 0 \\0 & 10 & 0\\0 & 0 & 0.1\end{matrix}\right)$. 
    
        The $P$ and $\dot{L}$ obtained by \gls{eca} is 
    
        \begin{equation*}
        \begin{split}
        P_{3D} &= 
        \begin{bmatrix}
        0.01377437 & -0.00443745 & 1.0064974 \\
        0.02682977 & -0.9951887 & -0.00494725 \\
        1.0032624 & 0.02540882 & -0.01346647
        \end{bmatrix} \\
        \dot{L}_{3D} &= 
        \begin{bmatrix}
        6.8667921e-22 & 7.7224560e-22 \\
        1.0000000e+00 & 3.4256167e-22 \\
        1.8692240e-21 & 1.0000000e+00 \\
        \end{bmatrix}
        \end{split}
        \end{equation*}
    
        with which the model could obtain an accuracy of $94.24\%$ on the validation data. It  outperforms any other linear models included in the table. Meantime. We would obtain the equivalent form of $\dot{L}$ if we used \gls{aeca} with \Cref{equ:modified_prob}. That is 
    
        $$
        \mathfrak{L}_{3D}^{dense} = \nint{\dot{L}_{3D}^{dense}} = 
        \begin{bmatrix}
        1 & 1 \\
        1 & 0 \\
        0 & 1 \\
        \end{bmatrix}
        $$
    
        to which the result is rounded. And we won't mention if we round the result in the rest of the paper. 
    
        \begin{table}[!ht]
         \caption{Comparison with other classifiers of 3D data set}
          \centering
          \begin{tabular}{|c|c|c|c|}
            \toprule
            & \multicolumn{2}{c|}{Metrics} &\\
            \cmidrule(r){2-3}
            Name & Accuracy & Confustion Matrix & Total Parameters\\
            \midrule
            LoR  & 0.6671    & $\bigl( \begin{matrix}0.1974 & 0.1\\0.8026 & 0.9\end{matrix}\bigr)$ & 4\\
            \midrule
            LDA & 0.6667    & $\bigl( \begin{matrix}0.2005 & 0.1021\\0.7995 & 0.8979\end{matrix}\bigr)$ & 7\\
            \midrule
            QDA & 0.9368   & $\bigl( \begin{matrix}0.9683 & 0.0789\\0.0317 & 0.9211\end{matrix}\bigr)$ & 19\\
            \midrule
            SVM & 0.6684    & $\bigl( \begin{matrix}0.0 & 0.0\\1.0 & 1.0\end{matrix}\bigr)$ & 4\\
            \midrule
            KSVM & 0.4682    & $\bigl( \begin{matrix}0.9921 & 0.7915\\0.0079 & 0.2085\end{matrix}\bigr)$ & \diagbox[dir=NW,width=10em,height=2.5em]{\;}{\;}\\
            \midrule
            ECA & 0.9424 & $\bigl( \begin{matrix}0.9439 & 0.0583\\0.0561 & 0.9417\end{matrix}\bigr)$ & 15\\
            \bottomrule
          \end{tabular}
          \label{tab:3d}
        \end{table}
    
      \end{itemize}
    
    
    
      \subsubsection{MNIST data set (using approximated eigen component analysis (AECA), vanilla eigen component analysis (VECA) and eigen component analysis network (ECAN))}
    
      This experiment exhibited the capability of dimension reduction of \gls{eca}. Meantime, it's a good illustration of the extensionality of \gls{eca}. 
      
      \begin{itemize}
        \item{\gls{aeca}}
    
    
          With no more than 12 epochs of training, we could obtain an accuracy of $91.82\%$ on the MNIST data set, which outperforms \gls{lda} ($87.30\%$). \gls{qda} collapsed on this data set. The corresponding confusion matrices of \gls{eca}, \gls{lda}, \gls{qda} are listed together with their accuracy (\Cref{fig:mnist_cm_eca}).

          Part of the learnt eigenfeatures are displayed in \Cref{fig:mnist_nonoverlap_eigenfeatures}. These overlapped eigenfeatures (mapped to two and more classes) could be separated by amplitude based separator or raising dimension. The crowdedness of eigenfeatures (see \Cref{fig:mnist_degeneracy_and_crowdedness}) showed that the digit '1' needs the least eigenfeatures to express itself and the digit '8' needs the most eigenfeatures to express. From the overlapping histogram of classes on eigenfeatures (see \Cref{fig:mnist_overlapping}), we could found that more than 300 eigenfeatures are mapped to a single class. 
    

          Part of our obtained $P$ and $\dot{L}$ is
          \begin{equation*}
            P_{MNIST} = 
            \begin{bmatrix}
            -0.00413941 & 0.00901582 &  0.04932139 & \hdots & 0.02962722 & 0.03481099 & -0.00057647 \\
            0.00875694 & 0.00608521 & 0.01776391 & \hdots & 0.06011203 &  -0.00239336 & 0.01214911 \\
            0.02980248 & -0.05551565 & -0.05156798 & \hdots & 0.06693095 & 0.04165073 & 0.03867088 \\
            \vdots & \vdots & \vdots & \ddots & \vdots & \vdots & \vdots \\
            -0.0261358 & 0.00822736 &  -0.07249103 & \hdots & -0.00446923 & 0.09919091 & 0.01148881 \\
            -0.00769167 & 0.02939839 & 0.04298031 & \hdots & -0.01120429 & -0.00835039 & 0.00816467 \\
            -0.02147239 & -0.03735997 & -0.03697227 & \hdots & -0.00956103 & 0.02511721 & -0.00121021 \\
            \end{bmatrix}
          \end{equation*}
    
          and
    
          \begin{equation*}
            \dot{L}_{MNIST} = 
            \begin{bmatrix}
            4.28501236e-16 & 3.17267414e-16 & \hdots & 4.07497572e-16 & 3.69587051e-16 \\
            2.94324218e-11 & 1.02724047e-12 & \hdots & 1.69453180e-08 & 1.83475474e-11 \\
            2.42868087e-11 & 6.55227218e-12 & \hdots & 9.99977827e-01 & 8.40801428e-10 \\
            \vdots & \vdots & \ddots & \vdots & \vdots \\
            8.78414095e-13 & 4.65978872e-13 & \hdots & 5.07113839e-12 & 9.25257024e-11 \\
            6.30129141e-13 & 3.91200852e-13 & \hdots & 6.63759282e-12 & 1.00000000e+00 \\
            7.75741233e-15 & 8.41619221e-15 & \hdots & 1.31394444e-14 & 1.69584344e-13 \\
            \end{bmatrix}
          \end{equation*}
    
          such that 
    
            \begin{equation*}
            \mathfrak{L}_{MNIST} = 
            \nint{\dot{L}_{MNIST}}=
            \begin{bmatrix}
            0 & 0 & \hdots & 0 & 0 \\
            0 & 0 & \hdots & 0 & 0 \\
            0 & 0 & \hdots & 1 & 0 \\
            \vdots & \vdots & \ddots & \vdots & \vdots \\
            0 & 0 & \hdots & 0 & 0 \\
            0 & 0 & \hdots & 0 & 1 \\
            0 & 0 & \hdots & 0 & 0 \\
            \end{bmatrix}.
          \end{equation*}

          \begin{figure}
            \centering
            \includegraphics[width=\linewidth]{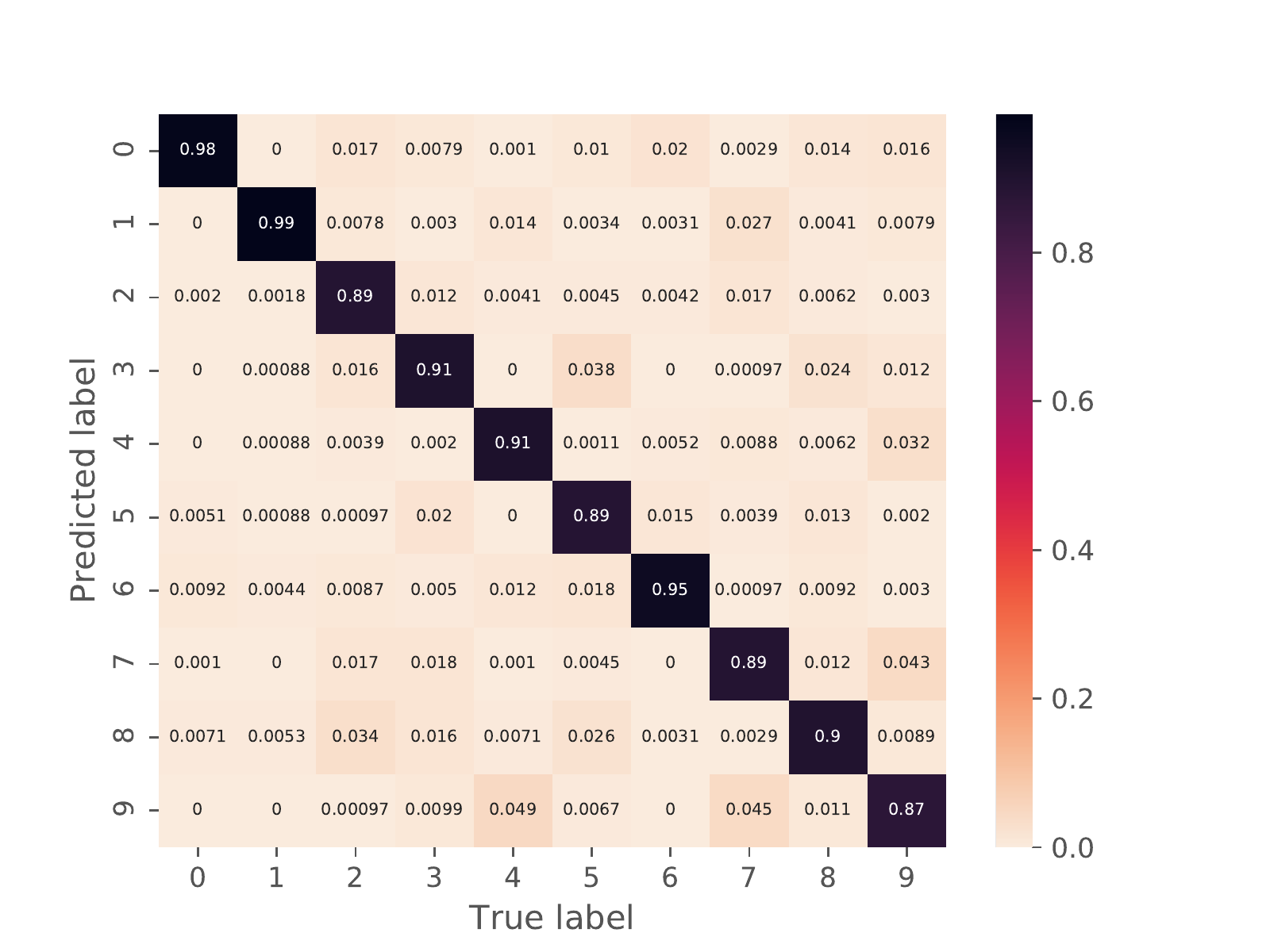}
            \caption{Confusion matrix of approximated eigen component analysis (AECA) model on MNIST data set (Accuracy of ECA, LDA, QDA are 0.918, 0.873, 0.144 respectively.)}
            \label{fig:mnist_cm_eca}
          \end{figure}

      \item{\gls{veca}}

        The $\mathfrak{L}$ learnt by \gls{veca} is extremely sparse, thus our eigenfeature is rather abstract. Using \gls{veca}, we achieved an validation accuracy of $90.48\%$. The reason that this accuracy is lower than that obtained by \gls{aeca} is that \gls{veca} is less tolerant to weak mapping between eigenfeatures and their class labels. \Gls{veca} intends to learn each element of an \gls{ecmm} as unambiguous as possible. Nevertheless, the most intriguing part is that we learnt only 110 \glspl{pe} (see \Cref{fig:mnist_ber_proj_freq_dist_ef,fig:mnist_ber_degeneracy_and_crowdedness,fig:mnist_ber_nonoverlap,fig:mnist_ber_overlapping}) attribute to the unambiguity. In \Cref{fig:mnist_ber_proj_freq_dist_ef} (i), we found the class '7' and the class '9' are both distant from distributions of other classes. However, this \gls{pe} is unambiguously assigned to the class '9'. This phenomenon indicates that \gls{veca} might be less tolerant to weak mappings between eigenfeature and their class labels, which is consistent to our objective in development of this two models. 
    
      \item{\gls{ecan}}

        We implemented several 2-fold \glspl{ecan} in this experiment. All these experiments are trained in 12 epochs. An indentity dimension operator is implemented if a \gls{rado} or \gls{redo} is not mentioned. 

        Since the major task in this demonstatrion is extensionability, the margin of accuracy for parameter tuning is possible for suited dimension operator. As limited by finding the orthogonal dictionary and linear operation, the performance in prediction accuracy is marginally underperformed than standard \glspl{dnn}. Also, in classical simulation, the training time is at least doubled than standard \glspl{dnn} because the extra linear operation which won't be a problem on a quantum computer. 


        The validation accuracy of the second fold achieved $94.58\%$ when we implemented a specially designed \gls{redo} (see \Cref{equ:redo}) in that reduced the dimension to $128$ in the first fold of \gls{ecan}. 


        Moreover, a non-quadratic \gls{rado} (see \Cref{equ:nq_rado}) and \gls{redo} (see \Cref{equ:nq_redo}) has been implemented in neural network with \gls{relu} activation function which is on par with these quadratic operators. The accuracies of each folds get to $96.86\%$ and $96.83\%$, $91.43\%$ and $94.66\%$, $96.6\%$ and $97.6\%$ for the three subexperiments with \gls{rado}, \gls{redo}, and both operators. 

        Instead of \gls{rado} or \gls{redo}, we implemented fully connected neural networks as a dimension operator. The accuracies of each folds get to $96.33\%$ and $96.37\%$, $91.29\%$ and $94.52\%$, $96.6\%$ and $96.87\%$ with a fully connected neural network with $128$ units implemented at the position of \gls{rado}, \gls{redo} and both place in the first fold. 

      \end{itemize}

      \subsubsection{Two breast cancer data sets}
      This experiment used two data sets which could be used to illustrate the high interpretability of \gls{eca}. We analyzed two eigenfeature of the first data set to explain the meaning of what we've obtained. 
    
      In this two experiments, we used two data sets downloaded from UCI machine learning repository, which was originally obtained from the University of Wisconsin Hospitals, Madison by Dr. William H. Wolberg) One data set was published in 1992 (abbreviated as Wis1992) and the other one was in 1995 (Wis1995). ECA achieved the validation accracies of $90.04\%$ and $94.14\%$, respectively, with all other afermentioned failed on these two data sets. 
    
      
      \begin{itemize}
        \item[Wis1992] {With \gls{veca}, we achieve an accuracy $90.04\%$ (\Cref{tab:wis1992}). The eigenvalue of $H$ and its corresponding degeneracy is listed below.
    
        \begin{center}
          \begin{tabular}{ |c|c|c|c| } 
            \hline
            Eigenvalue & Binary eigenvalue & Class label of \gls{pe}  & Degeneracy \\
            \hline
            0 & 00 & \diagbox[dir=NW,width=15em,height=0.8em]{\;}{\;} & 5 \\
            \hline
            1 & 01 & $0 = \log_2 (01)$ & 2 \\ 
            \hline
            2 & 10 & $1 = \log_2 (10)$ & 2 \\ 
            \hline
            3 & 11 & \diagbox[dir=NW,width=15em,height=0.8em]{\;}{\;} & 0 \\
            \hline
          \end{tabular} 
        \end{center}
    
        }
    
        \begin{table}[!ht]
         \caption{Comparison with other classifiers on the Wis1992 data set}
          \centering
          \begin{tabular}{|c|c|c|c|}
            \toprule
            & \multicolumn{2}{c|}{Metrics} &\\
            \cmidrule(r){2-3}
            Name & Accuracy & Confusion Matrix & Total Parameters \\
            \midrule
            LoR  & 0.3420 & $\bigl( \begin{matrix}0.0 & 0.0\\1.0 & 1.0\end{matrix}\bigr)$ & 10\\
            \midrule
            LDA & 0.3420 & $\bigl( \begin{matrix}0.0 & 0.0\\1.0 & 1.0\end{matrix}\bigr)$ & 19\\
            \midrule
            QDA & 0.3420 & $\bigl( \begin{matrix}0.0 & 0.0\\1.0 & 1.0\end{matrix}\bigr)$ & 109\\
            \midrule
            SVM & 0.3420 & $\bigl( \begin{matrix}0.0 & 0.0\\1.0 & 1.0\end{matrix}\bigr)$ & 10\\
            \midrule
            KSVM & 0.3420 & $\bigl( \begin{matrix}0.0 & 0.0\\1.0 & 1.0\end{matrix}\bigr)$ & \diagbox[dir=NW,width=10em,height=2.5em]{\;}{\;}\\
            \midrule
            ECA & 0.9004 & $\bigl( \begin{matrix}0.9539 & 0.2025\\0.0461 & 0.7975\end{matrix}\bigr)$ & 99\\
            \bottomrule
          \end{tabular}
          \label{tab:wis1992}
        \end{table}
    
        In Wis1992, the 9 original features are 'Clump Thickness', 'Uniformity of Cell Size', 'Uniformity of Cell Shape', 'Marginal Adhesion', 'Single Epithelial Cell Size', 'Bare Nuclei', 'Bland Chromatin', 'Normal Nucleoli' and 'Mitoses'. For this data set, the \gls{ecmm} $\dot{L}$ we obtained is 
    
        \begin{equation*}
        \begin{split}
        \dot{L}_{Wis1992} &= 
        \begin{bmatrix}
        1.0 & 1.2813353e-21 \\
        1.544713e-21 & 1.0 \\
        1.0064267e-20 & 9.3284e-21 \\
        1.0 & 1.7838357e-15 \\
        4.8638498e-20 & 4.5091693e-19 \\
        1.3446889e-13 & 4.2358635e-17 \\
        2.2195558e-13 & 1.1143089e-15 \\
        2.4169718e-14 & 1.0 \\
        0.0019637463 & 1.0590642e-09
        \end{bmatrix}
        \end{split}
        \end{equation*}
    
        First, we choose the 0th eigenfeature from \gls{efm} $P$ as obtained. With the \gls{ecmm}, we know this eigenfeature is a \gls{pe} mapping to the class '0' (i.e. 'benign' tumor). This eigenfeature and its squared value are 
    
        \begin{equation*}
        P_0 = 
        \begin{bmatrix}
        0.15118523\\0.2486596\\-0.5196432\\-0.020636577\\0.051793348\\-0.3954467\\0.6642099\\0.03258596\\-0.18750139
        \end{bmatrix}
        \text{ and }
        (P_0)^2 = 
        \begin{bmatrix}
        0.022856973\\0.061831594\\0.27002904\\0.00042586832\\0.0026825508\\0.15637809\\0.4411748\\0.0010618448\\0.03515677
        \end{bmatrix}.
        \end{equation*}
    
        What's the meaning of $P_0$ and $(P_0)^2$? In analysis of a new input vector, we could use \gls{efm}. To analyze the eigenfeatures in \gls{efm}, we should use a special \gls{efm}, the identity matrix. First of all, $P_0$ is a paradigm or a textbook solution of 'benign' tumor indicator. For $P_0$, the value of its elements represent the relative intensity on each original feature. In more detail, high 'Bland Chromatin', and relatively high 'Clump Thickness' and 'Uniformity of Cell Size' with low 'Uniformity of Cell Shape','Bare Nuclei', and 'Mitoses' tend to be symptoms of 'benign' tumor. This 'benign' \gls{pe} indicates how we take into each original feature into account when we make the decision of the tumor being 'benign'. Considering a special \gls{efm} $I_{9\times9}$, the 6th value in $(P_0)^2$ is $0.441748$, which means one should take $44.17\%$ into account of 'Bland Chromatin' together with 'Uniformity of Cell Shape' ($27.00\%$) and 'Bare Nuclei' ($15.64\%$). 
    
        Next, we take the 1st eigenfeature which is a 'malignant' \gls{pe} into inspection. This eigenfeature and its squared value are
    
        \begin{equation*}
        P_1 = 
        \begin{bmatrix}
        0.009413117\\-0.40779403\\0.09775494\\0.21722561\\0.26575032\\0.16900736\\0.103898466\\-0.048212416\\-0.81856734
        \end{bmatrix}
        \text{ and }
        (P_1)^2 = 
        \begin{bmatrix}
        8.8606765e-05\\0.16629598\\0.009556028\\0.047186967\\0.070623234\\0.028563488\\0.010794891\\0.002324437\\0.67005247
        \end{bmatrix}.
        \end{equation*}
    
        No doubt the 'Mitoses' is the factor that we should consider the most ($67.01\%$) to decide a tumor being 'malignant', together with 'Clump Thickness'($16.63\%$). If a patient with less 'Mitoses' and 'Uniformity of Cell Size' and relatively high 'Marginal Adhesion', 'Single Epithelial Cell Size', 'Bare Nuclei', and 'Bland Chromatin', a 'malignant' diagnosis might be on the way. 
    
        \item[Wis1995] {With \gls{veca}, we achieved an accuracy $94.14\%$ (\Cref{tab:wis1995}). The eigenvalue of $H$ and its corresponding degeneracy is listed below.
    
        \begin{center}
          \begin{tabular}{ |c|c|c|c| } 
            \hline
            Eigenvalue & Binary eigenvalue & Class label of \gls{pe}  & Degeneracy \\
            \hline
            0 & 00 & \diagbox[dir=NW,width=15em,height=0.8em]{\;}{\;} & 4 \\
            \hline
            1 & 01 & $0 = \log_2 (01)$ & 17 \\ 
            \hline
            2 & 10 & $1 = \log_2 (10)$ & 9 \\ 
            \hline
            3 & 11 & \diagbox[dir=NW,width=15em,height=0.8em]{\;}{\;} & 0 \\
            \hline
          \end{tabular} 
        \end{center}
    
        \begin{table}[!ht]
         \caption{Comparison with other classifiers on Wis1995 data set}
          \centering
          \begin{tabular}{|c|c|c|c|}
            \toprule
            & \multicolumn{2}{c|}{Metrics} &\\
            \cmidrule(r){2-3}
            Name & Accuracy & Confustion Matrix & Total Parameters \\
            \midrule
            LoR  & 0.4043 & $\bigl( \begin{matrix}0.0 & 0.0\\1.0 & 1.0\end{matrix}\bigr)$ & 31\\
            \midrule
            LDA & 0.4043& $\bigl( \begin{matrix}0.0 & 0.0\\1.0 & 1.0\end{matrix}\bigr)$ & 61\\
            \midrule
            QDA & 0.3032 & $\bigl( \begin{matrix}0.0 & 0.25\\1.0 & 0.75\end{matrix}\bigr)$ & 991\\
            \midrule
            SVM & 0.4043 & $\bigl( \begin{matrix}0.0 & 0.0\\1.0 & 1.0\end{matrix}\bigr)$ & 31\\
            \midrule
            KSVM & 0.5957 & $\bigl( \begin{matrix}1.0 & 1.0\\0.0 & 0.0\end{matrix}\bigr)$ & \diagbox[dir=NW,width=10em,height=2.5em]{\;}{\;}\\
            \midrule
            ECA & 0.9414 & $\bigl( \begin{matrix}0.9911 & 0.1316\\0.0089 & 0.8684\end{matrix}\bigr)$ & 960\\
            \bottomrule
          \end{tabular}
          \label{tab:wis1995}
        \end{table}
    
        }
      \end{itemize}


\section{Discussions}


  We proposed a new quantum machine learning algorithm. This algorithm could be simulated on a classical computer. We used \gls{veca} for data classification which outperforms \gls{lor}, \gls{lda}, \gls{qda}, \gls{svm} and \gls{ksvm} (with \gls{rbf} kernel). 
  One drawback of \gls{veca} is that it ignores the amplitude difference but focuses on the phase difference. The magnitude information was discarded in its classification performance, though, in practice, we found this magnitude information did not show substantial influence on the model. One solution to recover the lost information is to wrap the magnitude into our original vector. Other solutions include raising dimension before normalization or adopting a parallel \gls{fnn} on magnitude (see more about this in \Cref{app:ext_eca}). With extension, our algorithm could work with amplitude difference along each eigenfeature. Thus a combination method based upon these two components is expected to build a more robust linear classifier. \Gls{ecan} can further improve the performance by integrating with nonlinear models such as deep networks.
  This method could be used in text classification and sentiment analysis as text usually has intricate linearity. \ignore{Also, the quantitative trade could also exploit this method because the linearity of these financial data.  }

  The advantages of \gls{eca} can be found in several aspects. First, as a classifier, any hyperplane separating data can be separated by \gls{eca} with no more than one auxilliary dimension, i.e. an extra dimension with unit or constant length (details omitted due to space limitation). In addition, for classification, \gls{eca} could process more than two classes simultaneously. Unlike \gls{pca} or \gls{lda}, \gls{eca} does not need to specify the number of dimensions for a lower dimensional feature space. The concrete number could be calculated from \gls{ecmm} neither more nor less. Not only can \gls{eca} work as a good classifier, but it can also obtain a good dictionary. 

  Moreover, as this method is inspired by quantum mechanics, we introduce the concept of degeneracy in quantum mechanics as redundancy in machine learning problem. With degeneracy, we could also learn an undercomplete or overcomplete dictionary. With \gls{eca}, a complete dictionary could also be nontrivial. Indeed, the final dictionary (composed by \glspl{pe}) will be a subset of all eigenfeatures based upon the obtained \gls{ecmm}. Besides, this redundancy introduced could be not only used in our method, but also anywhere independently tackling a machine learning problem to avoid overfitting on linearity. In conclusion, \gls{eca} is an algorithm which deeply exploits the divide-and-conquer strategy.




\section{Methods}
We first develop a classical approximation of the algorithm which could be implemented on a classcial computer. Afterward, a quantum algorithm could be presented to implement on a quantum computer. We begin our development from the quantum intuition and fade out to the classical simulation followed by the full quantum algorithm. 

For an observable, we could develop a 'machine' or apparatus to measure it. In quantum mechanics, such an observable or measurable are represented by linear operators $H$. These eigenvalues of the operator are the possible to-be-observed results of a measurement. Corresponding eigenvectors of these eigenvalues represents unambiguously distinguishable states. When we measure on an observable $H$ at state $A$, the probability to observer value $\lambda_j$ is given by


  \begin{equation}
  p(\lambda_j) = \braket{A|\lambda_j}\braket{\lambda_j|A}.
  \label{equ:prob_eigen}
  \end{equation}


Now we define a new observable or measurable the \textit{class-label} \ignore{(as well as a hypothesis, see \Cref{app:hypothesis}) }$\mathcal{H}$ and its corresponding linear operator $H$. In addition to all the principles or assumptions  in quantum mechanics, there are two assumptions needed to be hold in the development of \gls{veca}:
\begin{itemize}
  \item {Assumption 1: Any system with a measurable \textit{class-label} is a quantum system; }
  \item {Assumption 2: 
    \begin{itemize}
      \item {On a quantum computer, for each class, $l$ commutative measurable could be built for each class label; }
      \item {On a classical computer, each class label of each eigenfeature follows independent Bernoulli distribution.}
    \end{itemize}
  }
\end{itemize}



If we view our vector representation $\ket{\mathsf{x}}$ of a to-be-classified object $\ket{\mathsf{o}}$ as a quantum state, the measurement could be expressed as 

\begin{equation}
\mathcal{H} \ket{\mathsf{o}}
\end{equation}

or 

\begin{equation}
\label{equ:measurement}
{H} \ket{\mathsf{x}} = H\sum_{\lambda}\psi_{\lambda} \ket{\lambda}=\sum _{\lambda}\psi_{\lambda} \lambda \ket{\lambda}.
\end{equation}

in which $\lambda$ is the eigenvalue and $\ket{\lambda}$ is the eigenvector of $H$. We conduct a series of measuremnts on states $\mathsf{x}$ to obtain its correponding observed value $\mathsf{y}$. Then we acquire a data set of measurement results on states $\{\ket{x^{(i)}}\} \text{ with observed values (class label) } \{y^{(i)}\} \; (\text{where } i=0,1,\cdots,n-1)$. One ideal situation for totally unambiguous states, the measurement on \textit{class-label} could observe the class label (which is an integer). Hence the class label is the corresponding eigenvalue of $H$. That is 

\begin{equation}
\label{equ:analytic}
H \ket{\mathsf{x}^{(i)}} = y^{(i)} \ket{\mathsf{x}^{(i)}} \quad (i=0,1,\cdots,n-1).
\end{equation}

For a data set with a full rank design matrix $X$ whose states are unambiguous, we could obtain an analytic solution for \Cref{equ:analytic}, such that

\begin{equation}
\begin{aligned}[]
& H X^{T} & = & Y^T \star X^T \\
\Rightarrow & H X^{T}X & = & (Y^T \star X^T) X \\
\Rightarrow & H & = & (Y^T \star X^T) X (X^{T}X)^{-1}
\end{aligned}
\end{equation}

in which $Y$ is the vector of all the class labels and $\star$ is element-wise scalaring along columns. Given a new state $\ket{\mathsf{x}}$, the prediction could be obtained by the rounded result of the expectation value of $H$ given state $\ket{\mathsf{x}}$. This analytic solution is implemented with performance over several classical algorithms mentioned in experiments section. However, one eigenstate (i.e. eigenfeatures) could be overlapped by several classes and states in one class could project on several eigenfeatures. What we could predict, in the measurement of $\mathcal{H}$ on a state $\ket{\mathsf{x}}$, would be the expectation of the observable. With \Cref{equ:measurement}, we have

\begin{equation}
\braket{H} = \braket{\mathsf{x}|H|\mathsf{x}}=\sum_{\lambda}\lambda p_{\lambda}.
\end{equation}

These eigenvalues crushed together, which cannot give us the information about the \gls{pmf} of how the input state would collapse on eigenstates. Hence it couldn't give us the information which the classes it belongs to. Thus for a $l$-class data set, we need to define $l$ commutative operators 

\begin{equation}
H_0, H_1, \cdots, H_{l-1}.
\end{equation}

To identify the class label, we need to know the \gls{pmf} $p_\lambda$ firstly. For matrix form of $H$ (we'll assume $H$ is a matrix here and non-matrix $H$ would be discussed in Appendix), any Hermitian matrix could be diagonalized, such that

\begin{equation}
H_k = P \Lambda_k P^{\dagger}, k=0, 1, \cdots, l-1
\end{equation}

in which $\Lambda$ is composed by eigenvalues of $H_k$ on its diagonal and unitary operator $P$ is composed by simultaneous eigenvectors of all $l$ operators $H$ as its column vector. Thus, for all these $l$ operators, we want to find a complete basis of simultaneous eigenvectors of all these $l$ operators and the eigenvalues of each measurable. 

Then, with (\Cref{equ:prob_eigen}), the \gls{pmf} of collapsing on each eigenstate $\ket{\lambda}$ given a state $\ket{\mathsf{x}}$ is

\begin{equation}
p(\ket{\lambda}|\ket{\mathsf{x}}) = (\bra{\mathsf{x}}P)^* \odot (\bra{\mathsf{x}}P)
\label{equ:collapsing_prob}
\end{equation}

To identify the unambiguous relationship between eigenfeature and class label, we assume the Bernoulli random variable $\mathsf{y}_k$ if one eigenfeature or vector is belonging to a class $k$ follows independent Bernoulli distributions (development based upon a categorical distribution assumption would be attached in Appendix). Thus, the \gls{pmf} of the classes to which an eigenfeature is belonging could be described as 

\begin{equation}
p_{k,\lambda}(\mathsf{y}_k=1|\ket{\lambda}).
\end{equation}

Hence , by the principle of superposition, the \gls{pmf} of $\mathsf{y}_k$ given $\ket{\mathsf{x}}$ which is the decision if one state belongs to one class $k$ would be

\begin{equation}
\label{equ:single_class_pmf_no_rounded}
p_{k}(\mathsf{y}_k=1|\ket{\mathsf{x}}) = \sum_{\lambda} p_{k,\lambda}(\mathsf{y}_k|\ket{\lambda}) p({\lambda}|\ket{\mathsf{x}}) 
\end{equation}


For all classes, the matrix composed by stacked or combined \gls{pmf} of $\mathsf{y}$ with combined Bernoulli random variables $\mathsf{y}_k$ given $\ket{\mathsf{x}}$ could be denoted as 

\begin{equation}
\begin{split}
\mathbb{p}(\mathsf{y}|\ket{\mathsf{x}}) 
 = 
 \begin{pmatrix}
\sum_{\lambda} p_{0,\lambda}^T(\mathsf{y}_0=1|\ket{\lambda}) p({\lambda}|\ket{\mathsf{x}}) & \sum_{\lambda} p_{0,\lambda}^T({\mathsf{y}_0}=0|\ket{\lambda}) p({\lambda}|\ket{\mathsf{x}}) \\[6pt]
\sum_{\lambda} p_{1,\lambda}^T(\mathsf{y}_1=1|\ket{\lambda}) p({\lambda}|\ket{\mathsf{x}}) & \sum_{\lambda} p_{1,\lambda}^T({\mathsf{y}_1}=0|\ket{\lambda}) p({\lambda}|\ket{\mathsf{x}}) \\[6pt]
\vdots & \vdots \\[6pt]
\sum_{\lambda} p_{l-1,\lambda}^T(\mathsf{y}_{l-1}=1|\ket{\lambda}) p({\lambda}|\ket{\mathsf{x}}) & \sum_{\lambda} p_{l-1,\lambda}^T({\mathsf{y}_{l-1}}=0|\ket{\lambda}) p({\lambda}|\ket{\mathsf{x}}) 
\end{pmatrix}_{l \times 2} 
 = \sum_{\lambda} {\mathbb{p}_{\lambda}(\mathsf{y}|\ket{\lambda})} p({\lambda}|\ket{\mathsf{x}})
\label{equ:combine_pmf}
\end{split}
\end{equation}

in which the bold font $\mathbb{p}(\cdot)$ indicates a matrix and the subscript indicates the size of the vector or matrix. 

Furthermore, the mapping between eigenfeature and class label follow the rule of winner-take-all, i.e the probability would be rounded to 0 or 1, such that

\begin{equation}
\label{equ:single_class_pmf_rounded}
p_{k}(\mathsf{y}_k=1|\ket{\mathsf{x}}) = \sum_{\lambda} \nint{p_{k,\lambda}(\mathsf{y}_k|\ket{\lambda})} p({\lambda}|\ket{\mathsf{x}}) 
\end{equation}

and

\begin{equation}
\label{equ:pmf_rounded}
\mathbb{p}(\mathsf{y}|\ket{\mathsf{x}}) = \sum_{\lambda} \nint{\mathbb{p}_{\lambda}(\mathsf{y}|\ket{\lambda})} p({\lambda}|\ket{\mathsf{x}}).
\end{equation}

in which $\nint{\cdot}$ is round operation.

Then we put the rounded distribution of $\mathsf{y}_k$ given $\ket{\lambda}$ together to form a $m\times l$ matrix $\mathfrak{L}$. By substituting \Cref{equ:collapsing_prob} into \cref{equ:combine_pmf}, the combined probabilities of $\mathbb{y}$ given $\ket{\mathsf{x}}$ could be written in

\begin{equation}
\mathbb{p}(\mathsf{y}|\ket{\mathsf{x}}; \theta) = 
\begin{Bmatrix}
\{ [(\bra{\mathsf{x}} P)^* \odot (\bra{\mathsf{x}} P)] \mathfrak{L}\}^T & \{ [(\bra{\mathsf{x}} P)^* \odot (\bra{\mathsf{x}} P)] ( \mathbb{1}_{m\times l} - \mathfrak{L} )\}^T
\end{Bmatrix}_{l \times 2}
\label{equ:eigen_prob}
\end{equation}

in which $\theta$ denotes all the unknown parameters and $\mathbb{1}_{m\times l}$ is a $m \times l$ all-ones matrix. When one eigenfeature only belongs to one class, the corresponding row of $\mathfrak{L}$ is a bitwise representation of the binary digit of the class label. With these eigenfeature overlapped by several calsses, we define its eigenvalue the binarized number of the corresponding reversed row of $\mathfrak{L}$. Hence, we define

\begin{equation}
\Lambda \equiv Diag(\mathfrak{L}_{\overset{\leftarrow}{2}}).
\end{equation}

in which the subscript ${\overset{\leftarrow}{2}}$ means reversely binarizing the row vector of $\mathfrak{L}$ and $Diag(\cdot)$ denotes the operation diagnalizing a vector into a matrix with elements on diagonal and otherwise 0. With $\Lambda$ or $\mathfrak{L}$, in classical simulation, the $l$ commutative operators ($H_0, H_1, \cdots, H_{l-1}$) could be combined into a single operator $H$ such that

\begin{equation}
H = P \Lambda P^{\dagger}
\end{equation}

In the example of the \gls{2d} and \gls{3d} data set (see \Cref{fig:2d_3d,example}), the corresponding separator has eigenvalues $\bigl\{ (01)_2, (10)_2 \bigr\}$ and $\bigl\{ (10)_2,(01)_2,(11)_2 \bigr\}$ respectively, in which $(\cdot)_2$ indicates that the number or the number in a set is binary. Correspondingly, we also convert the class labels as reversely bitwise view of its one-hot vector expression. The conversion is depicted as 

\begin{equation}
\begin{split}
\mathcal{C} = & \{0,1,\cdots,l-1\} 
\Rightarrow 
\begin{Bmatrix} 
\bigl[ 1&0&\hdots&0 \bigr]^T, \\[6pt]
\bigl[ 0&1&\hdots&0 \bigr]^T, \\[6pt]
\multicolumn{4}{c|}{\vdots,} \\
\bigl[ 0&0&\hdots&1 \bigr]^T \\[6pt]
\end{Bmatrix} 
\Rightarrow \\
\mathcal{C}_2=  & \bigl\{ (000\cdots001)_2,(000\cdots010)_2,\cdots, (100\cdots000)_2 \bigr\}. \\
\end{split}
\end{equation}

with which we could easily to determine which classes an eigenfeature belongs to. These eigenfeatures that only belongs to one class are called \gls{pe}. In our terminology, \gls{pe} are $1^o$ eigenfeature in which the superscript $k^o$ denotes $k$ degree of overlapping, which means one eigenfeature has been shared by $k$ classes. For a \gls{pe} $\mathfrak{f}$ corresponding to eigenvalue $\lambda$, the class label $c$ of it could be calculated trivially such that

\begin{equation}
c_{\mathfrak{f}} = \log_2(\lambda).
\end{equation}

Therefore, to predict the measurements on measurable \textit{class-label} $\mathcal{H}$, instead of learning $H$ directly (to build neural networks simulating a function $H$ to represent $\mathcal{H}$), we could learn $P$ and $\Lambda$ (i.e. $\mathfrak{L}$) to construct our $H$ in a classical simulation. In the quantum algorithm, a subtle difference is that $H$ could be learned directly with a quantum computer.  

Given a data set $\{(\mathsf{x}^{(i)}, \mathsf{y}^{(i)}), i=0,1,\cdots,n-1\}$, with \Cref{equ:single_class_pmf_rounded}, we could denote the combined probabilities of observing $\mathbb{1}_{l \times 1}$ given $\mathsf{x}^{(i)}$ as

\begin{equation}
\mathbf{p}(\mathsf{y}=\mathbb{1}_{l \times 1}|\mathsf{x}^{(i)}; \theta) 
= 
\begin{bmatrix}
p_0 \\
p_1 \\
\vdots \\
p_{l-1}
\end{bmatrix}
\end{equation}

and then with \Cref{equ:combine_pmf} we have

\begin{equation}
\mathbb{p} 
=
\begin{bmatrix}
\mathbf{p} & \mathbb{1}_{l \times 1} - \mathbf{p}
\end{bmatrix}.
\label{equ:bernoulli_prob_expansion}
\end{equation}

As we assume the decision on each class label of each eigenfeature are independent, the probability of a measurement of $\mathsf{y}^{(i)}$ given $\mathsf{x}^{(i)}$ is 

\begin{equation}
p(\mathsf{y}^{(i)}|\mathsf{x}^{(i)}; \theta) 
= p(\mathsf{y}_0^{(i)}, \mathsf{y}_1^{(i)}, \cdots, \mathsf{y}_{l-1}^{(i)}|\mathsf{x}^{(i)}; \theta)
= \prod_{k=0}^{l-1}p_k(\mathsf{y}_k^{(i)}|\mathsf{x}^{(i)}; \theta_k).
\label{equ:prod_of_joint_dist}
\end{equation}



Then the log-likelihood function is
\begin{equation}
\begin{split}
\mathcal{L}(\theta;\mathsf{x}, \mathsf{y}) = & \log(\prod_{i=0}^{n-1}p(\mathsf{y}^{(i)}|\ket{\mathsf{x}^{(i)}}; \theta)). \\
\end{split}
\end{equation}


To learn $P$ and $\mathfrak{L}$, our objective could be

\begin{equation}
\begin{split}
\argmax_{P, \mathfrak{L}} & \log(\prod_{i=0}^{n-1}p(\mathsf{y}^{(i)}|\ket{\mathsf{x}^{(i)}}; \theta)) \\
\text{subject to } & H^{\dagger} = H
\end{split}
\label{equ:likelihood_obj}
\end{equation}

in which the constraint on $H$ is a shorthand for the constraints on $P$ and $\mathfrak{L}$. Actually, the optimization of this objective is NP-hard.

By substituting \Cref{equ:prod_of_joint_dist} into \Cref{equ:likelihood_obj},  then with expansion and regrouping the objective could be simplified as

\begin{equation}
\begin{split}
\mathcal{L}(\theta;\mathsf{x}, \mathsf{y}) = & \log(\prod_{i=0}^{n-1}p(\mathsf{y}^{(i)}|\ket{\mathsf{x}^{(i)}}; \theta)) \\
= & \log  \left( \prod_{i=0}^{n-1} \prod_{k=0}^{l-1} p_k(\mathsf{y}_k^{(i)}|\ket{\mathsf{x}^{(i)}}; \theta_k) \right) \\
= & \log \left( \prod_{i=0}^{n-1} \prod_{k=0}^{l-1} \prod_{\kappa=0}^{1} p_k(\kappa|\ket{\mathsf{x}^{(i)}}; \theta_k)^{1\{\kappa=\mathsf{y}_k^{(i)}\}} \right) \\
= & \sum_{i=0}^{n-1} \sum_{k=0}^{l-1} \sum_{\kappa=0}^{1} {1\{\kappa=\mathsf{y}_k^{(i)}\}} \log p_k(\kappa|\ket{\mathsf{x}^{(i)}}; \theta_k)  \\
= & \sum_{i=0}^{n-1} \sum_{k=0}^{l-1}  {\mathbb{1}^T\{\mathsf{y}_k^{(i)}=1\}} \log p_k(\mathsf{y}_{k}|\ket{\mathsf{x}^{(i)}}; \theta_k)  \\
= & \sum_{i=0}^{n-1} \mathsf{y}^{(i)T} \log \mathbf{p}^{(i)} + (\mathbb{1}_{1 \times l} - \mathsf{y}^{(i)T}) \log (\mathbb{1}_{l \times 1} - \mathbf{p}^{(i)}) \\
= & \sum_{i=0}^{n-1} \text{Tr} (\mathbb{y}^{(i)T} \log \mathbb{p}^{(i)})
\end{split}
\end{equation}

in which the $\widetilde{\mathsf{y}}$ is the one's complement of each element of $\mathsf{y}$, $\text{Tr}(\cdot)$ is the trace of corresponding matrix and $\mathbf{p}^{(i)}$ and $\mathbb{p}^{(i)}$ are $\mathbf{p}(\mathsf{y}^{(i)}=\mathbb{1}_{l \times 1}|\ket{\mathsf{x}^{(i)}}; \theta)$ and $\mathbb{p}(\mathsf{y}|\ket{\mathsf{x}^{(i)}}; \theta)$ in short respectively.

Hence our objective becomes

\begin{equation}
\begin{split}
\argmax_{P, \mathfrak{L}} & \sum_{i=0}^{n-1} \text{Tr} (\mathbb{y}^{(i)T} \log \mathbb{p}^{(i)}) \\
\text{subject to } & H^{\dagger} = H.
\end{split}
\end{equation}

With \Cref{equ:prob_eigen}, we have

\begin{equation}
\begin{split}
\argmin_{P, \mathfrak{L}} & - \sum_{i=0}^{n-1} \big\{ \mathsf{y}^{(i)T} \log \{[(\bra{\mathsf{x}^{(i)}} P)^* \odot (\bra{\mathsf{x^{(i)T}}} P)] \mathfrak{L} \}^T \\
& + (\mathbb{1}_{1 \times l} - \mathsf{y}^{(i)T}) \log \{[(\bra{\mathsf{x}^{(i)}} P)^* \odot (\bra{\mathsf{x}^{(i)}} P)] (\mathbb{1}_{m \times l} - \mathfrak{L}) \}^T \big\} \\
\text{subject to } & H^{\dagger} = H.
\end{split}
\end{equation}

\begin{figure}
  \centering
  \includegraphics[width=\linewidth]{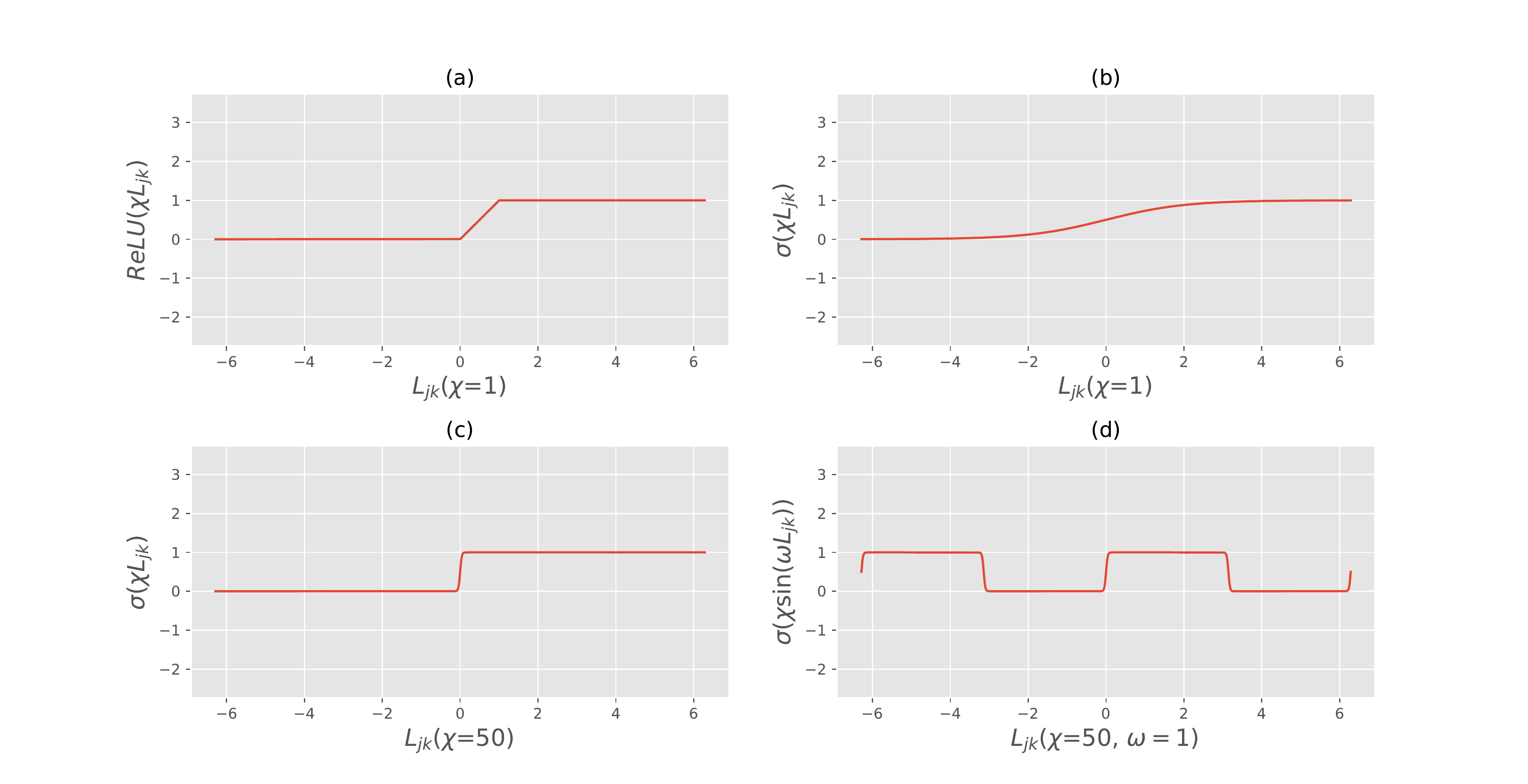}
  \caption{Several hypothesis of eigenfeature-class mapping matrix (ECMM). (a)ReLU function; (b)Sigmoid function; (c)Sigmoid function which is more concentrated on 0 or 1; (d)Sigmoid function on sinusoid which periodicly concentrates on 0 or 1.}
  \label{fig:imambiguity}
\end{figure}

The round operation is not differentiable. Thus, we replace $\mathfrak{L}$ with $\dot{L}$ which is sigmoid function of parameters $L$ (\Cref{fig:imambiguity} (a)):

\begin{equation}
\dot{L}_{jk} = \frac{1}{1+e^{-\chi L_{jk}}}
\end{equation}

in which $\chi$ is the imambiguity factor to make the probability more concentrate on $0$ or $1$ and $\dot{()}$ is the sigmoid function operator. The higher $\chi$ then the probability would be more concentrated on 0 or 1. Then the round operation on $\dot{L}$ would be less risky.  As we want $\mathfrak{L}$ be a binary matrix, to make this constraint more neat, we add an auxiliary sinusoid function on $L$ (\Cref{fig:imambiguity} (d)) such that

\begin{equation}
\dot{L}_{jk} = \frac{1}{1+e^{-\chi \sin(\omega L_{jk})}}.
\end{equation}

The sigmoid function could output 0 or 1 when the input approches $-\infty \text{ or } + \infty$ respectively. Nevertheless, the elements of $\dot{L}$ could never approch 0 or 1 theoretically. In practice, with a relatively large $\chi$, $\dot{L}$ works as a good approximation of sigmoid function.



Thus, we replace the constraint on $H$ with $P$ and $L$. Our objective becomes

\begin{equation}
\begin{split}
\argmin_{P, L} & - \sum_{i=1}^{n} \big\{ \mathsf{y}^{(i)T} \log \{[(\bra{\mathsf{x}^{(i)}} P)^* \odot (\bra{\mathsf{x^{(i)T}}} P)] \dot{L} \}^T \\
& + (\mathbb{1}_{1 \times l} - \mathsf{y}^{(i)T}) \log \{[(\bra{\mathsf{x}^{(i)}} P)^* \odot (\bra{\mathsf{x}^{(i)}} P)] (\mathbb{1}_{m \times l} - \dot{L}) \}^T \big\} \\
\text{subject to } & (P^{\dagger}P)^* \odot (P^{\dagger}P) = I \\
& \sin^2(L) = \mathbb{1}_{m \times l} \; (\text{i.e. } \cos{(2L)}=-\mathbb{1}_{m \times l}).
\end{split}
\end{equation}


Then we obtain our objective function using Frobenius norm which is

\begin{equation}
\begin{split}
\argmin_{P, L} & - \sum_{i=1}^{n} \big\{ \mathsf{y}^{(i)T} \log \{[(\bra{\mathsf{x}^{(i)}} P)^* \odot (\bra{\mathsf{x^{(i)T}}} P)] \dot{L} \}^T \\
& + (\mathbb{1}_{1 \times l} - \mathsf{y}^{(i)T}) \log \{[(\bra{\mathsf{x}^{(i)}} P)^* \odot (\bra{\mathsf{x}^{(i)}} P)] (\mathbb{1}_{m \times l} - \dot{L}) \}^T \big\} \\
& + {\xi \|I - (P^{\dagger}P)^* \odot (P^{\dagger}P)\|^2_F} \\ 
& + \gamma \|\mathbb{1}_{m \times l}+\cos(2L)\|^2_F.
\end{split}
\end{equation}

Since most samples we meet are in real coordinate space, for simplicity and without loss of generality, we assume all the vectors would always project only on real space (i.e. the imaginary part always equals 0). The real version of our objective could be

\begin{equation}
\begin{split}
\argmin_{P, {L}} & - \sum_{i=1}^{n} \big\{ \mathsf{y}^{(i)T} \log \{[({\mathsf{x}^{(i)T}} P) \odot ({\mathsf{x^{(i)T}}} P)] \dot{L} \}^T \\
& + (\mathbb{1}_{1 \times l} - \mathsf{y}^{(i)T}) \log \{[({\mathsf{x}^{(i)T}} P) \odot ({\mathsf{x}^{(i)T}} P)] (\mathbb{1}_{m \times l} - \dot{L}) \}^T \big\} \\
& + {\xi \|I - P^{T}P\|^2_F} \\ 
& + \gamma \|\mathbb{1}_{m \times l}+\cos(2L)\|^2_F.
\end{split}
\end{equation}

  \subsection{Approximation of eigen component analysis (ECA)}

  Furthermore, for a relative large data set, the combined probabilities of a combined random vector $\mathbb{y}$ of Bernoulli random variables $\mathsf{y}_k$ could be an approximation of the \gls{pmf} of $y$, such that these combined probabilities

  \begin{equation}
  \mathbf{p}(\mathsf{y}^{(i)}|\mathsf{x}^{(i)};\theta)
  \end{equation}
  could be used to estimate \gls{pmf} of $y$ given $\mathsf{x}$

  \begin{equation}
  p(y|\mathsf{x}^{(i)};\theta).
  \end{equation}

  The log-likelihood function becomes 

  \begin{equation}
  \begin{split}
  \mathcal{L}(\theta; y, \mathsf{x}) 
  = & \log \prod_{i=0}^{n-1} \prod_{i=k}^{l-1} p(k|\mathsf{x}^{(i)};\theta)^{1\{k=y^{(i)}\}} \\
  = & \sum_{i=0}^{n-1} \sum_{i=k}^{l-1} {1\{k=y^{(i)}\}} \log p(k|\mathsf{x}^{(i)};\theta) \\
  = & \sum_{i=0}^{n-1} {\mathbb{1}\{y=y^{(i)}\}} \log p(y|\mathsf{x}^{(i)};\theta). \\
  = & \sum_{i=0}^{n-1} \mathsf{y}^{(i)}\log p(y|\mathsf{x}^{(i)};\theta).
  \end{split}
  \end{equation}

  Then the objective of \gls{aeca} could be written in 

  \begin{equation}
  \begin{split}
  \argmin_{P, {L}} & - \sum_{i=1}^{n} \mathsf{y}^{(i)T} \log \{[({\mathsf{x}^{(i)T}} P) \odot ({\mathsf{x^{(i)T}}} P)] \dot{L} \}^T \\
  & + {\xi \|I - P^{T}P\|^2_F} \\ 
  & + \gamma \|\mathbb{1}_{m \times l}+\cos(2L)\|^2_F.
  \end{split}
  \end{equation}

  In default, the combined probabilities of $\mathsf{y}$ given $\mathsf{x}$ is 
  \begin{equation}
  \mathbf{p}(\mathsf{y}^{(i)}|\ket{\mathsf{x}^{(i)}}; \theta) = [(\bra{\mathsf{x}^{(i)}} P)^* \odot (\bra{\mathsf{x}^{(i)}} P)] \dot{L}.
  \label{equ:unmodified_prob}
  \end{equation}
  As for a specific eigenvalue might be overlapped by more than one class, the probability on that eigenvector should be weighted. The modified probabilities or approximated \gls{pmf} is

  \begin{equation}
  \label{equ:modified_prob}
  \mathbf{p}(\mathsf{y}^{(i)}|\ket{\mathsf{x}^{(i)}}; \theta) = [(\bra{\mathsf{x}^{(i)}} P)^* \odot (\bra{\mathsf{x}^{(i)}} P)] \dot{L} / o(\dot{L})
  \end{equation}


  in which $o(\dot{L})$ is the degree of overlapping of each eigenfeature. With the modification, then the sum of probabilities in $\mathbf{p}$ could sum up to $1$. 


  \subsection{Raising and reducing dimension and eigen component analysis networks (ECANs)}

  In dealing with image or other interpolation tolerant data, the simplest method for rasing dimension might be resizing.For more advanced raising dimension strategy, we introduce a \gls{rado} and a \gls{redo}. In the development of this section, the vector $\mathsf{x}$ are all originally non-normalized vector. A layer of nerual network with $m$-dimensional input and $m'$ units of neurons is defined as 

  \begin{equation}
  W_{m' \times {(m+1)}}^T\begin{bmatrix}1\\\mathsf{x}\end{bmatrix}_{(m+1) \times 1}
  \end{equation}

  in which the extra 1 is viewed as an auxilliary dimension (we could call it the God-dimension because it provides the perspective over all other dimensions) with unit lenghth in \gls{eca}. Instead of neurons, all the units are viewed as dimensions in \gls{eca}. To deal with the dimension, in \gls{eca}, raising dimension is a process of unfold these wrapped dimension. Hence, this process is totally reversible as we project these unfolded dimension onto original dominated dimension. This is the same as to dimension reduction, with simple restriction, we combine several dimension whose corresponding eigenfeatures are degenerated. However, the restoration of reduced dimension could not guarantee intactness as the original one. 

  We could define our \gls{rado} $A$ that keep the original vector intact and unfold these wrapped dimensions, such that

  \begin{equation}
  A\mathsf{x}=
  \left[
  \begin{matrix}
  \mathsf{x}\\
  f_1(\mathsf{x})\\
  f_2(\mathsf{x})\\
  f_3(\mathsf{x})\\
  \vdots\\
  f_{m-\mathfrak{p}}(\mathsf{x})\\
  \end{matrix}
  \right].
  \label{equ:rado}
  \end{equation}

  We could also learn these parameters for rasing dimension. For a \gls{rado} $A$, it could be defined as

  \begin{equation}
  \mathsf{x}_{m\times1} = A_{m\times \mathfrak{p}}
  \mathsf{x}_{\mathfrak{p}\times1}\\
  =
  \left[
  \begin{matrix}
  \mathsf{x}_{\mathfrak{p}\times 1} \\[6pt]
  \sqrt{
  {A_{(m-\mathfrak{p})\times (\mathfrak{p}+1)}^{'}
  \left(
  \begin{matrix}
  1\\[3pt]
  \mathsf{x}_{\mathfrak{p}\times1}^2\\
  \end{matrix}
  \right)} 
  }
  \end{matrix}
  \right], \quad m>\mathfrak{p} \text{ and } A^{'}>0 \text{ for real \gls{eca}}.
  \end{equation}

  For dimension reduction operator $D$, that could be defined as

  \begin{equation}
  \mathsf{x}_{m\times1} = D_{m\times \mathfrak{p}}\mathsf{x}_{\mathfrak{p}\times 1}
  =\sqrt{D_{m\times (\mathfrak{p}+1)}^{'}
  \left[
  \begin{matrix}
  1\\
  \mathsf{x}_{\mathfrak{p}\times1}^2\\
  \end{matrix}
  \right]}
  ,\quad m<\mathfrak{p} \text{ and } D' > 0 \text{ for real \gls{eca}}.
  \label{equ:redo}
  \end{equation}

  Additionally, a non-quadratic \gls{rado} could be defined as 

  \begin{equation}
  \mathsf{x}_{m\times1} = A_{m\times \mathfrak{p}}
  \mathsf{x}_{\mathfrak{p}\times1}\\
  =
  \left[
  \begin{matrix}
  \mathsf{x}_{\mathfrak{p}\times 1} \\[6pt]
  \text{ReLU} \left( {A_{(m-\mathfrak{p})\times (2\mathfrak{p}+1)}^{'}
  \left(
  \begin{matrix}
  1\\[3pt]
  \mathsf{x}_{\mathfrak{p}\times 1}\\
  \end{matrix}
  \right)} \right)
  \end{matrix}
  \right], \quad m>\mathfrak{p} 
  \label{equ:nq_rado}
  \end{equation}

  and a non-quadratic \gls{redo} as well as a fully connected neural network could be 

  \begin{equation}
  \mathsf{x}_{m\times1} = D_{m\times \mathfrak{p}}\mathsf{x}_{\mathfrak{p}\times 1}
  ={\text{ReLU} \left( D_{m\times (\mathfrak{p}+1)}^{'}
  \left[
  \begin{matrix}
  1\\
  \mathsf{x}_{\mathfrak{p}\times1}\\
  \end{matrix}
  \right] \right)}
  ,\quad m<\mathfrak{p}
  \label{equ:nq_redo}
  \end{equation}

  in which $\text{ReLU} (\cdot)$ is the activation function of \gls{relu}. This operator becomes a general dimension operator with fully connected neural networks when we get rid of the constraint of size on $m$ and $\mathfrak{p}$.

  Moreover, we define two special dimension operator $A_I$ and $D_I$ with  which 

  \begin{equation}
  A_I\mathsf{x} = \mathsf{x} \text{ and } D_I\mathsf{x} = \mathsf{x}.
  \end{equation}

  The reduced dimension could be dynmaicly adapted to the overlapping and crowdedness of \gls{ecmm}. By raising dimension, these amplitude variances could be transformed into phase variances. 



    \subsubsection{Eigen component analysis networks (ECANs)}

    \begin{figure}
      \centering
      \includegraphics[width=\linewidth]{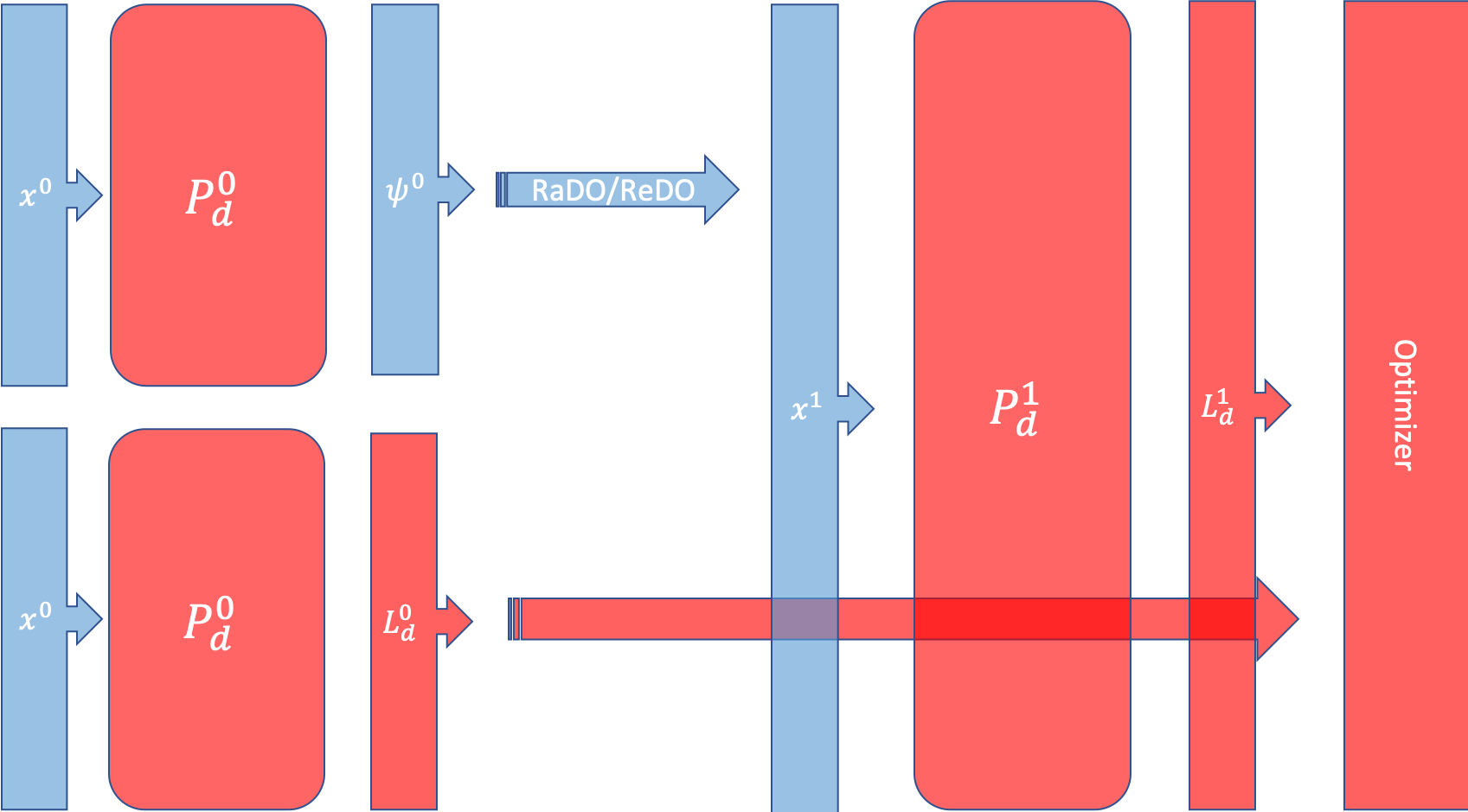}
      \caption{A 2-fold eigen component analysis network (ECAN) with a rasing dimension operator (RaDO) or reducing dimension operator (ReDO) in between.}
      \label{fig:ecan}
    \end{figure}

    In the development in this section, we also use the combined probabilities

    \begin{equation}
    \mathbf{p}(\mathsf{y}^{(i)}|\mathsf{x}^{(i)};\theta)
    \end{equation}

    as an approximation of \gls{pmf} of $y$ given $\mathsf{x}$

    \begin{equation}
    p(y|\mathsf{x}^{(i)};\theta).
    \end{equation}

    For each $P$, the corresponding state should be normailized. That a $\tau$-fold ($\tau=0,1,\cdots$) \gls{ecan} (see \Cref{fig:ecan}) is defined as 

    \begin{equation}
    \begin{split}
    f_{\tau}(\ket{{\mathsf{x}}}) &= \prod_{t=1}^{\tau}{D_t{P_{t}^{\dagger}\wwidehat{A_t\ket{{\mathsf{x}}}}}}\\
    \text{and } g_{t}(\ket{\mathsf{x}}) &= {{P_{t}^{\dagger}\wwidehat{A_t\ket{f_{t-1}(\ket{\mathsf{x}})}}}}, \text{ where } f_0(\ket{\mathsf{x}})=\ket{\mathsf{x}}\\
    \end{split}
    \end{equation}



    in which $\wwidehat{(\cdot)}$ is the normalization operator. One subtle point that is worth noticing, in implementation this algorithm, the \gls{redo} or \gls{rado} should use the identity operator alternatively because two consequent dimension operators are trival. 

    Specifically, a $0$-fold \gls{ecan} is 

    \begin{equation}
        p(y|\ket{\mathsf{x}};\theta) =[(\wwidehat{\bra{\mathsf{x}}}P_I)^* \odot (\wwidehat{\bra{\mathsf{x}}}P_I)]^T\mathfrak{L}_{\tau}\\
    \end{equation}

    in which $P_I$ is an \gls{efm} of the identity matrix . 

    Thus, the \gls{pmf} of $y$ given $\mathsf{x}$ is 

    \begin{equation}
    \begin{split}
    p(y|\ket{\mathsf{x}};\theta) & =[g_{\tau}^{*}(\ket{\mathsf{x}}) \odot g_{\tau}(\ket{\mathsf{x}})]^T\mathfrak{L}_{\tau}.\\
    \end{split}
    \end{equation}

    Then the objective becomes
    \begin{equation}
    \begin{split}
    \argmin_{P, \mathfrak{L}} & -\sum_{t =1}^{\tau}\pi_t\sum_{i=1}^{n} \mathsf{y^{(i)T}} \log \{[g_{t}^{*}(\ket{\mathsf{x}}) \odot g_{t}(\ket{\mathsf{x}})]^T\mathfrak{L}_{t} \}^T \\
    \text{subject to } & H_{t}^{\dagger} = H_{t} \text{ for } t = 1,2,\cdots,\tau-1.
    \end{split}
    \end{equation}

  \subsection{Generative model of eigen component analysis (ECA)}
  \label{app:gen_eca}


  Not only \gls{eca} could be used to develop a model with generative learning interpretation, but also we could also build \gls{gecan}.Empirically, We found the projections on the eigenfeature are drawn from multimodal normal distribution, such that

  \begin{equation}
  \braket{\mathsf{x}|\lambda}|y \sim \mathcal{M}(\mu_q, \sigma_q), q=0,1,\cdots
  \end{equation}

  in which $\mathcal{M}$ denotes the multimodal distribution of several complex normal distributions with mean $\mu_q$ and variance $\sigma_q$ for each. Indeed, for simplicity ,we could still assume independent complex normal distribution of projection on each eigenfeature for each class, such that

  \begin{equation}
  \braket{\mathsf{x}|\lambda_i}|y \sim \mathcal{N}_\mathbb{C}(\mu_{\lambda_i}, \sigma_{\lambda_i})
  \end{equation}

  or a real normal distribution 

  \begin{equation}
  \braket{\mathsf{x}|\lambda_i}|y \sim \mathcal{N}_\mathbb{R}(\mu_{\lambda_i}, \sigma_{\lambda_i}). 
  \end{equation}

  We assume class label $y$ follows a  categorical distribution, such that

  \begin{equation}
  y \sim \text{Categorical} (\phi_k), \text{ for } k = 0, 1, \cdots, l - 1 .
  \end{equation}

  For a classical generative model, the objective is

  \begin{equation}
  \argmax_y p(\bra{\mathsf{x}}|\mathsf{y})p(y).
  \label{equ:gen_objective}
  \end{equation}

  For the \gls{efm} $P$ and \gls{ecmm} $\mathfrak{L}$ to be determined, we denote the column vector of $P$ and $\mathfrak{L}$ as $P_j\ (j=0,1,\cdots,m-1)$ and $\mathfrak{L}_k\ (k=0,1,\cdots,l-1)$ respectively. 

  Given $y$, since we assume eigenfeatures are independent, the probabilty of $\mathsf{x}$ gcould be en y is a joint distribution which could be  

  \begin{equation}
  \begin{split}
  p(\bra{\mathsf{x}}|y) = & p(\braket{\mathsf{x}|P_0}, \braket{\mathsf{x}|P_1}, \cdots, \braket{\mathsf{x}|P_j}, \cdots, \braket{\mathsf{x}|P_{m-1}}|y) ,\text{ for } j \in \{j| \mathfrak{L}_{jy} = 1, j=0, 1, \cdots, m-1 \} \\
  = & \prod_{j=0}^{m-1} p(\braket{\mathsf{x}|P_j}|y)^{\mathfrak{L}_{jy}}
  \label{equ:gen_x_given_y}
  \end{split}
  \end{equation}



  Given the data set $\{(\mathsf{x}^{(i)}, y^{(i)})\}$, by substituting \Cref{equ:gen_x_given_y} into \Cref{equ:gen_objective}, the log-likelihood function is given by


  \begin{subequations}
  \begin{align}
  \mathcal{L}(\mu, \sigma, P, L, \phi) = & \log \prod_{i=0}^{n-1} \prod_{j=0}^{m-1} p(\braket{\mathsf{x}|P_j}|y^{(i)})^{\mathfrak{L}_{jy^{(i)}}} \mathfrak{L}_{y^{(i)}} \phi_{y^{(i)}} \\
  = & \sum_{i=0}^{n-1} \sum_{i=0}^{m-1} {\mathfrak{L}_{jy^{(i)}}} \log p(\braket{\mathsf{x}|P_j}|y^{(i)}) + \sum_{i=0}^{n-1} \log \phi_{y^{(i)}} \\
  = & \sum_{i=0}^{n-1} \sum_{i=0}^{m-1} {\mathfrak{L}_{jy^{(i)}}} \log \prod_{k=0}^{l-1} p(\braket{\mathsf{x}|P_j}|k)^{1\{k=y^{(i)}\}} + \sum_{i=0}^{n-1} \log \prod_{k=0}^{l-1} \phi_k^{1\{k=y^{(i)}\}} \\
  = & \sum_{i=0}^{n-1} \sum_{i=0}^{m-1} {\mathfrak{L}_{jy^{(i)}}} \sum_{k=0}^{l-1} {1\{k=y^{(i)}\}} \log p(\braket{\mathsf{x}|P_j}|k) + \sum_{i=0}^{n-1} \sum_{k=0}^{l-1} \log \phi_k^{1\{k=y^{(i)}\}} \\
  = & \sum_{i=0}^{n-1} \sum_{i=0}^{m-1} {\mathfrak{L}_{jy^{(i)}}} {\mathbb{1}^T\{y=y^{(i)}\}} \log p(\braket{\mathsf{x}|P_j}|y) + \sum_{i=0}^{n-1} \mathbb{1}^T\{y=y^{(i)}\} \log \phi \\
  = & \sum_{i=0}^{n-1} \sum_{i=0}^{m-1} [{\mathfrak{L}_{j}} \odot {\mathbb{1}^T\{y=y^{(i)}\}}] \log p(\braket{\mathsf{x}|P_j}|y) + \sum_{i=0}^{n-1} \mathbb{1}^T\{y=y^{(i)}\} \log \phi \\
  = & \sum_{i=0}^{n-1}  {\mathbb{1}^T\{y=y^{(i)}\}} {\mathfrak{L}^T}_{l \times m} \log p^T(\bra{\mathsf{x}}P|y)_{m \times l} {\mathbb{1}\{y=y^{(i)}\}} + \sum_{i=0}^{n-1} \mathbb{1}^T\{y=y^{(i)}\} \log \phi \\
  = & \sum_{i=0}^{n-1}  \mathsf{y}^{(i)T} {\mathfrak{L}^T}_{l \times m} \log p^T(\bra{\mathsf{x}}P|y)_{m \times l} \mathsf{y}^{(i)} + \sum_{i=0}^{n-1} \mathsf{y}^{(i)T} \log \phi \label{subequ:gen_eca_wasted_comp}
  \end{align}
  \end{subequations}

  in which the matrix operation in \Cref{subequ:gen_eca_wasted_comp} increased the computation  $l^2$ times futilely. 

  Then our objective could be

  \begin{equation}
  \begin{split}
  \argmax_{\mu, \sigma, P, L, \phi} \quad & \mathcal{L}(\mu, \sigma, P, L, \phi) \\
  \text{subject to } & H^{\dagger} = H.
  \end{split}
  \end{equation}

  In practice, these projections usually happen in real coordinate space. Thus, we substitute the real normal distribution

  \begin{equation}
  p({\mathsf{x}^T}P_j|y=k; \mu_{jk}, \sigma_{jk}) = \frac{1}{\sigma_{jk}\sqrt{2\pi}} e^{-\frac{1}{2}\bigl(\frac{x^T P_j - \mu_{jk}}{\sigma_{jk}}\bigr)^2}
  \end{equation}

  into \Cref{subequ:gen_eca_wasted_comp}. Thus, the log-likelihood function becomes 

  \begin{equation}
  \begin{split}
  \mathcal{L}_{\mathbb{R}}(\mu, \sigma, P, L, \phi) = & \sum_{i=0}^{n-1}  \mathsf{y}^{(i)T} {\mathfrak{L}^T}_{l \times m} \log \left\{\frac{1}{(\sigma\mathsf{y}^{(i)})^T \sqrt{2\pi}} \odot e^{-\frac{1}{2} \Bigl({(x^T P \ominus (\mu\mathsf{y}^{(i)})^T} \odot \frac{1}{(\sigma\mathsf{y}^{(i)})^T} \Bigr)^2} \right\}^T + \sum_{i=0}^{n-1} \mathsf{y}^{(i)T} \log \phi \\
  \end{split}
  \end{equation}

  in which $\mu$ and $\sigma$ are $m \times l$ matrix.

  Then our objective could be

  \begin{equation}
  \begin{split}
  \argmin_{\mu, \sigma, P, L, \phi} \quad & -\mathcal{L}_{\mathbb{R}}(\mu, \sigma, P, L, \phi)\\
  & + {\xi \|I - P^{T}P\|^2_F} \\ 
  &  + \gamma \|\mathbb{1}_{m \times l}+\cos(2L)\|^2_F.
  \end{split}
  \end{equation}


  By taking the gradient on $\phi$ and set that gradient to zero, we could obtain the solution of $\phi$. 


  \subsection{Eigen component analysis (ECA) and generative adversarial networks (GANs)}
  \label{app:gan_eca}
  
  \begin{figure}
    \centering
    \includegraphics[width=\linewidth]{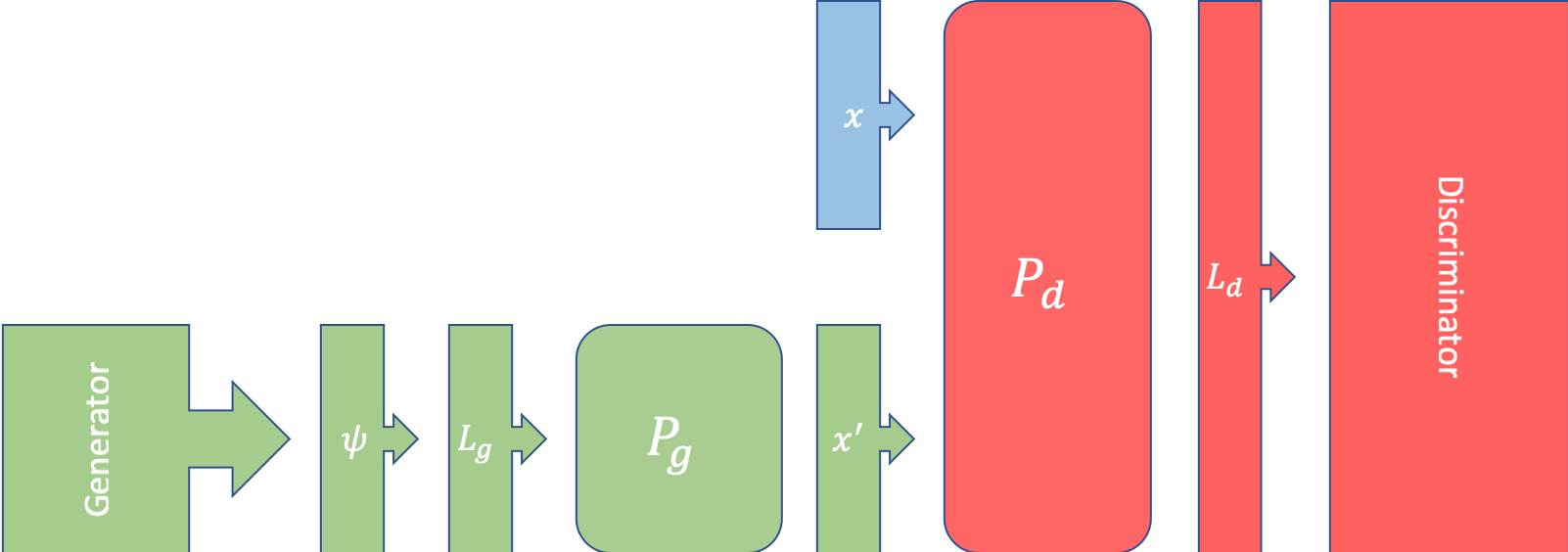}
    \caption{The architecture of an eigen component analysis-based generative adversarial network (ECAbGAN). Blocks in green belongs to generator. These blue blocks is related to ground truth and its evolution. The red blocks consists of the discriminator. }
    \label{fig:ecabgan}
  \end{figure}

  Learning cuisine in the field is way harder than in the kitchen of a top chef. In the field, one need to care about those ingridents as well as recipe. In the kitchen of a top chef, one is less risky to preparable an extremly terrible meal. As all the ingridents are well prepared, one can still 'mix' a not-that-bad meal even if they know nothing about cook. Hence, to learn cuisine, the only thing that one need to learn narrows to the recipe. Therefore, a cuisine learning problem could be devided into a ingredient learning and a recipe learning problem. 

\begin{figure}
    \centering
    \includegraphics[width=\linewidth]{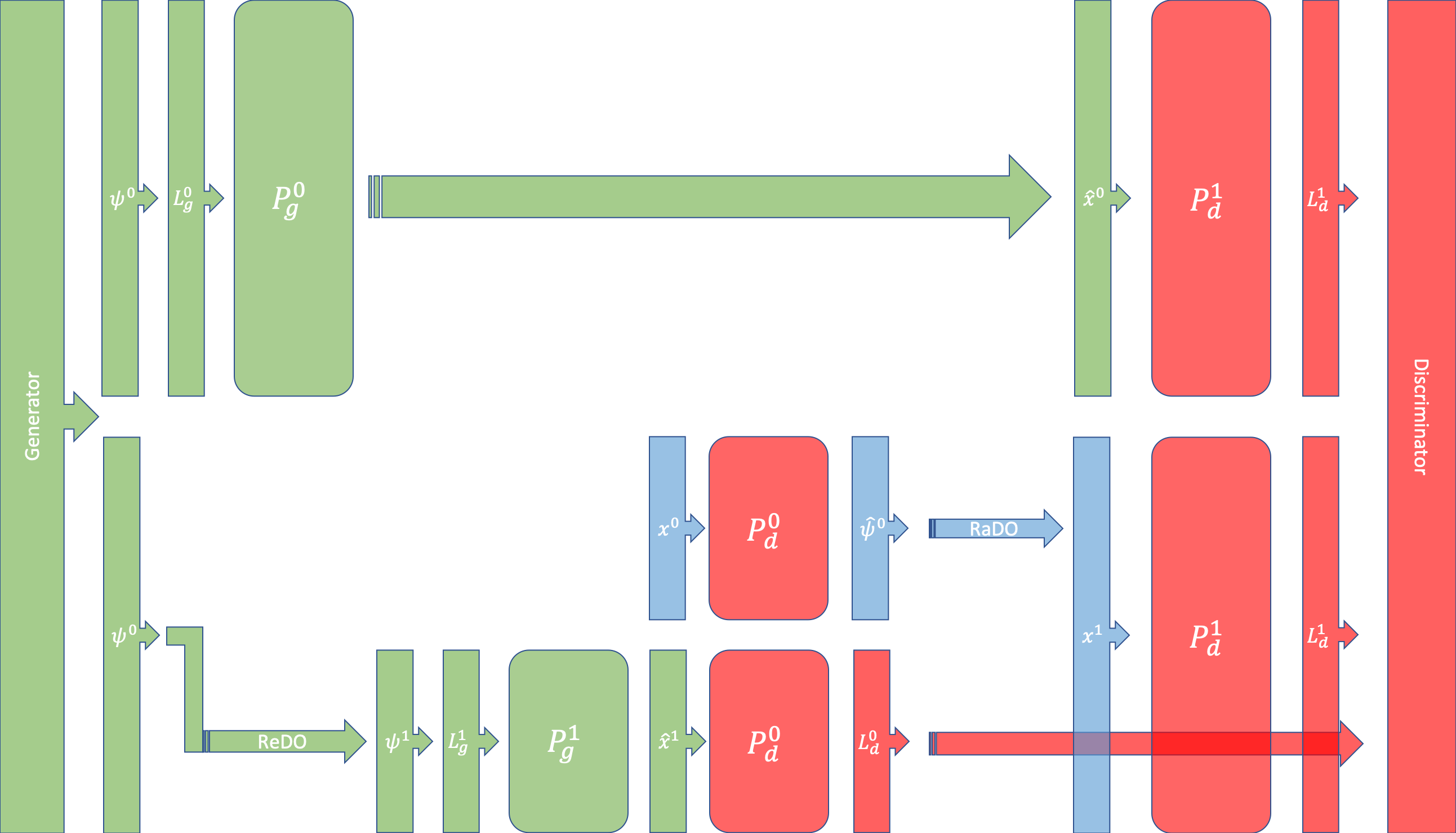}
    \caption{The architecture of a $2 \times 2$ eigen component analysis network-based generative adversarial network (ECANbGAN) was illustrated. In the first fold of eigen component analysis network (ECAN) in generator and discriminator, the implicit rasing dimension and dimension reduction operator are all identity operator.}
    \label{fig:ecabgan_2x2}
  \end{figure}

  We could build up an \gls{ecabgan} or \gls{ecanbgan} which learns the ingredients firstly then the recipe (\Cref{fig:ecabgan,fig:ecabgan_2x2}). 

  \subsection{Unsupervised eigen component analysis (UECA)}
  \label{app:ueca}

  For unlabeled data set, we still want to find some userful structure. For a data set $\{\mathsf{x}^{(i)}, i=0,1,\cdots,n-1\}$, we want to fit a model of $p(x, z)$ to the data, in which $z$ is the latent variable. Thus the log-likelihood function is 

  \begin{equation}
  \begin{split}
  \mathcal{L} (\theta)= & \sum_{i=0}^{n-1} \log p(\mathsf{x}) \\
  = & \sum_{i=0}^{n-1} \log \sum_z p(\mathsf{x, z; \theta}) \\
  \end{split}
  \end{equation}

  To optimize this function, we could use \gls{em} algorithm. We assume $z$ follows categorical distribution such that

  \begin{equation}
  z \sim \text{Categorical} (\phi_k), \text{ for } k = 0, 1, \cdots, \tilde{l} - 1, 
  \end{equation}

  in which $\hat{l}$ is a hyperparamter of to-be-classified classes in the data set. 

  Then  the algorithm could be described as:

  \begin{adjustwidth}{.5in}{0in}
  Repeat untill convergence \{
  \begin{itemize}
    \item[] {}

    \begin{adjustwidth}{.5in}{0in}
    \item[(E-step)] { For each i, set
    \begin{equation}
    \begin{split}
    Q_i(z^{(i)}) := & p(z^{(i)}|\mathsf{x}^{(i)};P, L) \\
    = & [\bra{\mathsf{x}^{(i)}} P \odot \bra{\mathsf{x}^{(i)}} P]^T \mathfrak{L}_{z^{(i)}} 
    \end{split}
    \end{equation}

    }
    \end{adjustwidth}

    \begin{adjustwidth}{.5in}{0in}
    \item[(M-step)] {

    \begin{equation}
    \begin{aligned}
    Q_i(z^{(i)}) :=  \argmax_{\mu, \sigma, P, L, \phi} & \sum_i \sum_{z^{(i)}} Q_i(z^{(i)}) \log \frac{p(\mathsf{x}^{(i)}, z^{(i)};\mu, \sigma, P, L, \phi)}{Q_i(z^{(i)})} \\
     \text{subject to } & H^{\dagger}=H \\
    =  \argmax_{\mu, \sigma, P, L, \phi} & \sum_i \sum_{z^{(i)}} Q_i(z^{(i)}) \log \frac{p(\mathsf{x}^{(i)}| z^{(i)};\mu, \sigma, P, L) p(z^{(i)}; \phi)}{Q_i(z^{(i)})} \\
     \text{subject to } & H^{\dagger}=H \\
    \end{aligned}
    \end{equation}

    }
    \end{adjustwidth}

  \end{itemize}
  \}
  \end{adjustwidth}



  \subsection{Eigen component analysis (ECA) on a quantum computer}
  \begin{figure}
    \centering
    \includegraphics[width=\linewidth]{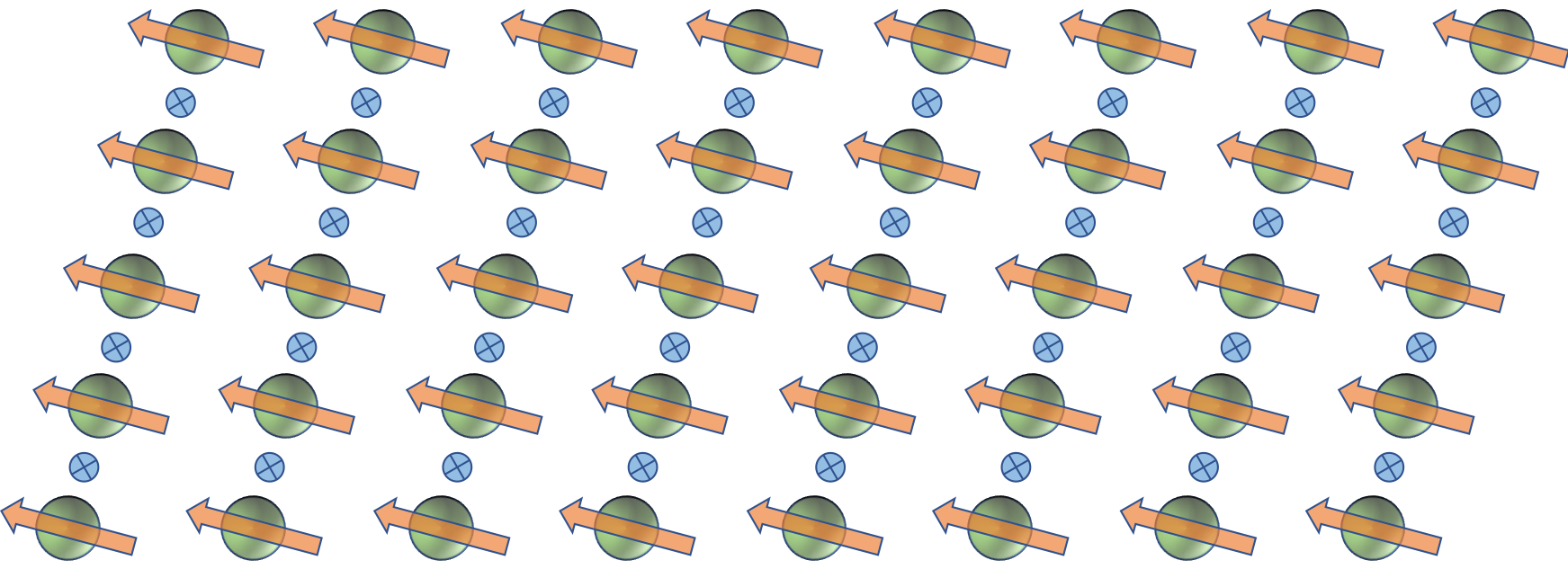}
    \caption{A prototype of a quantum computer for eigen component analysis (ECA) with $5 \times 8$ qubits. With this computer, a data set with $32$-dimenional state vector and $8$ classes could be parallelly processed. On the other hand, this computer could be used to process data with at most $2^{40}$-dimensional state vector. This ajustment depends on the task and the blance between the cost for preparing a state and adding qubits. }
    \label{fig:quantum_computer}
  \end{figure}

  The quantum version of this algorithm is litte bit different from the classical version in the state representation and measurements. However, all the underneath motivation is the same. We need to build a 'machine' or appratus to finish the measurement. For a $m$-dimensional data, all of its states could be represented by $\left \lceil{\log(m)}\right \rceil$ qubits (for simplicity, we always assume $\log(m)$ is an integer). \ignore{Also, for simplicity, we assume that no entanglement involved in the measurement of $H$, i.e. all the states are product states. }For a single qubit, the operator of its corresponding measurable $H_b$ could be written in spectral decomposition with unitary matrix $U_b$ and diagonal matrix $\Lambda_b$ such that

  \begin{equation}
  H_b = U_b \Lambda_b U_b^{\dagger}.
  \end{equation} 

  To decide the measurable, we need to determine its corresponding unitary operator as well as the eigenvalues in $\Lambda_b$. For each qubit, the eigenvalues could be $-1$ or $+1$. The unitary operator of a qubit could be written in

  \begin{equation}
  U_b = e^{i\varphi/2}
  \begin{bmatrix}
  e^{i\varphi_0} \cos \theta & e^{i\varphi_1} \sin \theta \\
  -e^{i\varphi_1} \sin \theta & e^{i\varphi_0} \cos \theta \\
  \end{bmatrix}.
  \end{equation}

  Together with the parameters for two eigenvalues, for a single qubit, there are 6 parameters needed to be determined for a $H_b$ in tatal (complex number counted as a single parameter). 

  For the composite system $\mathcal{H}$, its corresponding measurable $H$ could be the tensor product

  \begin{equation}
  H = H_{b_0} \otimes H_{b_1} \otimes \cdots \otimes H_{b_{\log(m)-1}}.
  \end{equation} 

  On the other hand, we need $l$ operators 

  \begin{equation*}
  H_0, H_1, \cdots, H_{l-1}
  \end{equation*}

  to measure on $l$ classes. Actually, as we want to find a complete set of simultaneous vectors, these $l$ operators commute because for $c, d = 0, 1, \cdots, l-1$ 

  \begin{equation*}
  [H_c, H_d] = H_cH_d - H_dH_c = U \Lambda_c U^{\dagger} U \Lambda_d U^{\dagger} - U \Lambda_d U^{\dagger} U \Lambda_c U^{\dagger} = 0
  \end{equation*}

  for which we could measure these $l$ measurables in arbitary order. 


  The measurement of $H$ could be the same as meansuring each subsystem $H_b$ (i.e qubit). As each measurement destroys a measuring state, if not for no coloning theorem, we could take $Q$ mesurements on each qubit and the observed value of the whole system could be caculated as product of the measurments of each qubit. Therefore, for '1' observed $r$ times, the probability $\frac{r}{Q}$ could be used for optimizing the design of our measurable. 

  For a data set with $l$ classes, measurement should be taked on each class with the same unitary operator but different eigenvalues. If we measure $l$ classes parallely, all the qubits needed is $l \log (m)$. The concurrency could be further improved with another $Q$ times qubits. The total number of parameters is $(4+2l) \log(m)$. One thing shoud be noticed, as theese $l$ operators commute, that a prepared state could be reused for $l$ times (if the number of qubits is limited and preparation is relatively expensive). A prototype of the quantum computer is depicted in \Cref{fig:quantum_computer}. Such a quantum computer could be used for processing a data set with $2^5$ eigenstates and $8$ classes parallelly.  

  Once the measurements finished, a classical computer is needed for optimizing these operators. Together with the data set, the corresponding probability $\frac{r}{Q}$ could be used for optimizing these parameters on a classical computer using any off-the-shelf optimizing algorithm. 

  To sum up, the algorithm could be described as 

  \begin{adjustwidth}{.5in}{0in}
  Repeat untill convergence \{
  \begin{itemize}
    \item[] {}

    \begin{adjustwidth}{.5in}{0in}
    \item[(1)] { 

    Initialize or update parameters of measurable $\mathcal{H}_b$ of each qubit for $l\log(m)$ qubits of all operators $H$ for $l$ measurables $\mathcal{H}$ on a classical computer. These parameters then could be used for buidling the 'machine' or apparatus for measurement on a quantum computer. 

    }
    \end{adjustwidth}

    \begin{adjustwidth}{.5in}{0in}
    \item[(2)] { 

    Prepare the input vector as $l$ identical copies of states on $l\log(m)$ qubits on a quantum computer.

    }
    \end{adjustwidth}

    \begin{adjustwidth}{.5in}{0in}
    \item[(3)] {

    Take measurments on all qubits for $Q$ times. For each measurable $\mathcal{H}$, the observed value is the product of all the $\log(m)$ qubits of that measurable. In the case of observing '1' for $r$ times, then the $l$ probabilities $\frac{r_k}{Q_k} (k=0, 1, \cdots, l-1)$ are recorded. 

    }
    \end{adjustwidth}

    \begin{adjustwidth}{.5in}{0in}
    \item[(4)] {

    Together with the ground truth of the data set,  these $l$ probabilities $\frac{r_k}{Q_k} (k=0, 1, \cdots, l-1)$ could be used for optimization of the parameters of operators for measurable $\mathcal{H}_b$ of each qubit on a classical computer with a classical optimization. 

    }
    \end{adjustwidth}

  \end{itemize}
  \}
  \end{adjustwidth}


\section{Acknowledgements}
This breast cancer databases was obtained from the University of Wisconsin Hospitals, Madison from Dr. William H. Wolberg. We also used Keras and the data set it provides. 


\nocite{*}

\bibliographystyle{unsrt}  
\bibliography{mlqm}  


\newpage


\begin{center}
\textbf{\large Supplemental Materials: Eigen Component Analysis}
\end{center}

\setcounter{section}{0} 
\setcounter{equation}{0}
\setcounter{figure}{0}
\setcounter{table}{0}
\setcounter{page}{1}

\makeatletter

\renewcommand{\thesection}{S\arabic{section}}
\renewcommand{\theequation}{S\arabic{equation}}
\renewcommand{\thefigure}{S\arabic{figure}}
\renewcommand{\thetable}{S\arabic{table}}

\begin{figure}[!htbp]
\centering
\includegraphics[width=\linewidth]{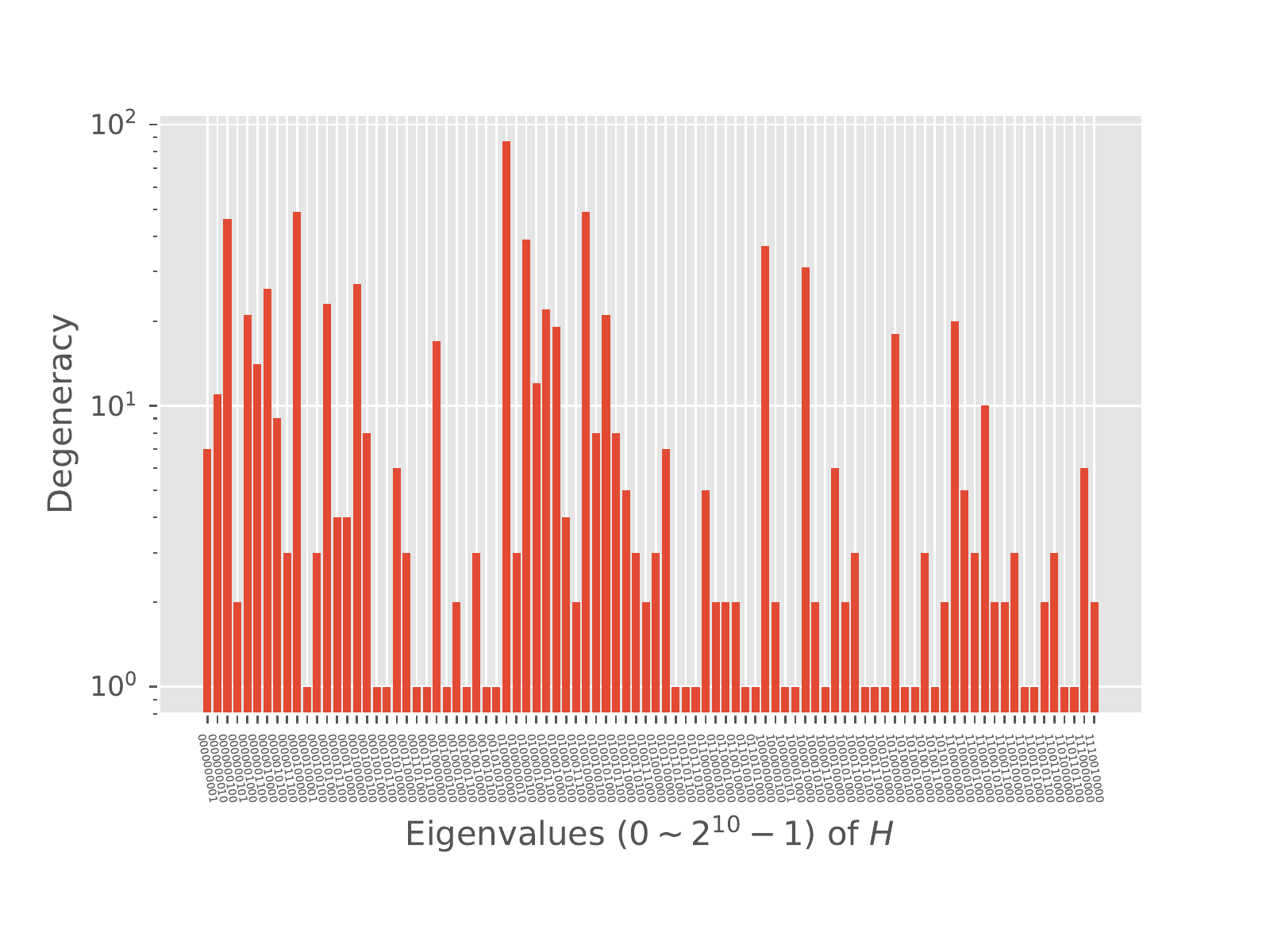}
\caption{Degeneracy of all $98$ distinctive eigenvalues of $H$ on the MNIST data set using approximated eigen component analysis (AECA). The degeneracy of eigenvalue $(0000000000)_2$ is $0$ which means there aren't eigenfeatures assume the similarities of the data set. The eigenvalue with the largest degeneracy ($171$) is $(0100000000)_2$, all corresponding to pure eigenfeatures (PEs) mapped to class label '8' ($\equiv \log_2(0100000000)$). The largest eigenvalue is $(1110010000)_2$ ($\equiv 912$) with degeneracy $2$. }
\label{fig:mnist_degeneracy_eigenvalues}
\end{figure}

\begin{figure}
\centering
\includegraphics[width=\linewidth]{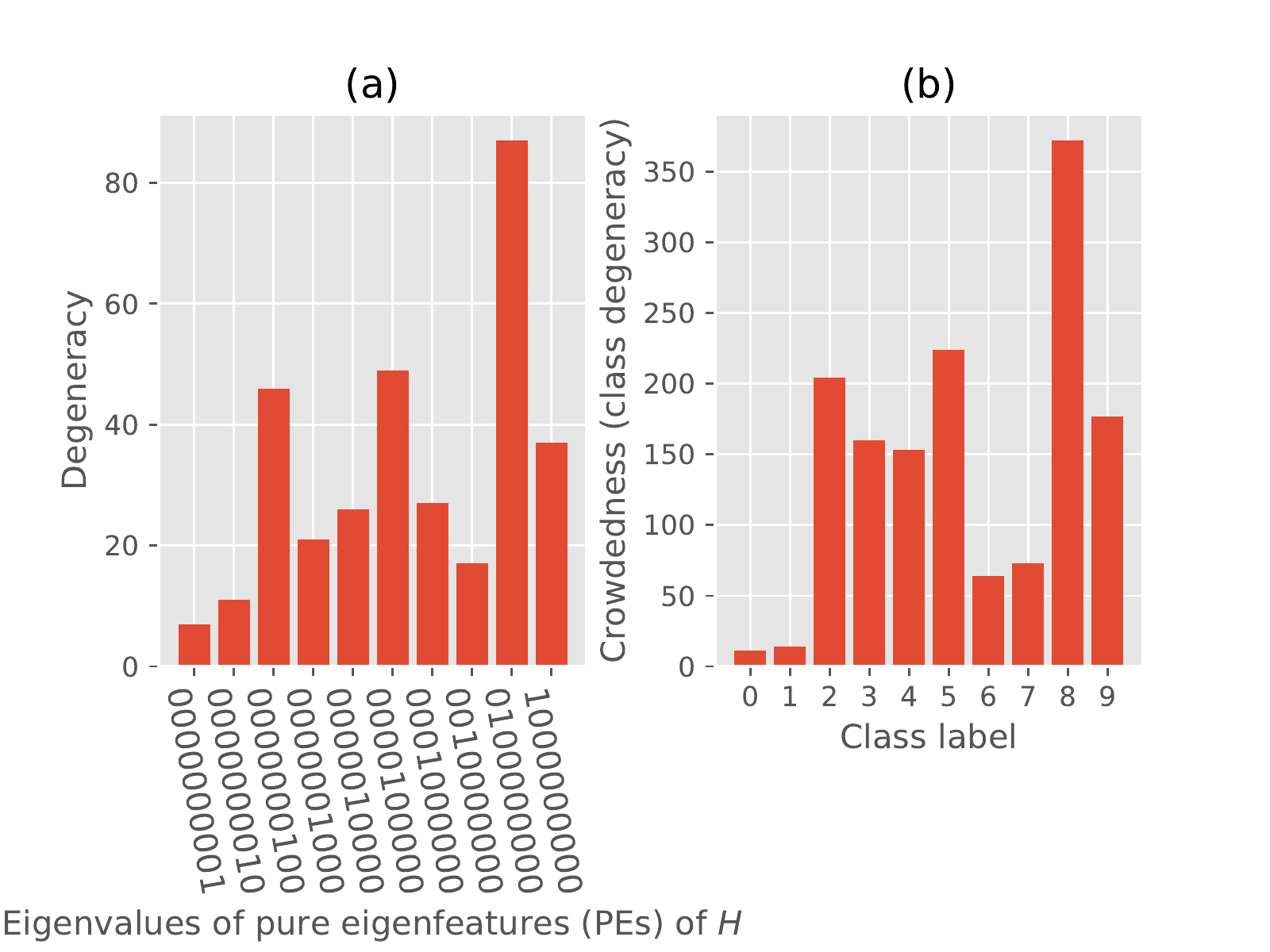}
\caption{(a) Degeneracy of pure eigenfeatures (PEs) of MNIST data set with approximated eigen component analysis (AECA); (b) Crowdedness of classes on MNIST data set with AECA. \ignore{From the figure we could find that the digit '8' needs more eigenfeatures than '1', as '8' is a more complicated digit.}}
\label{fig:mnist_degeneracy_and_crowdedness}
\end{figure}

\begin{figure}
\centering
\includegraphics[width=\linewidth]{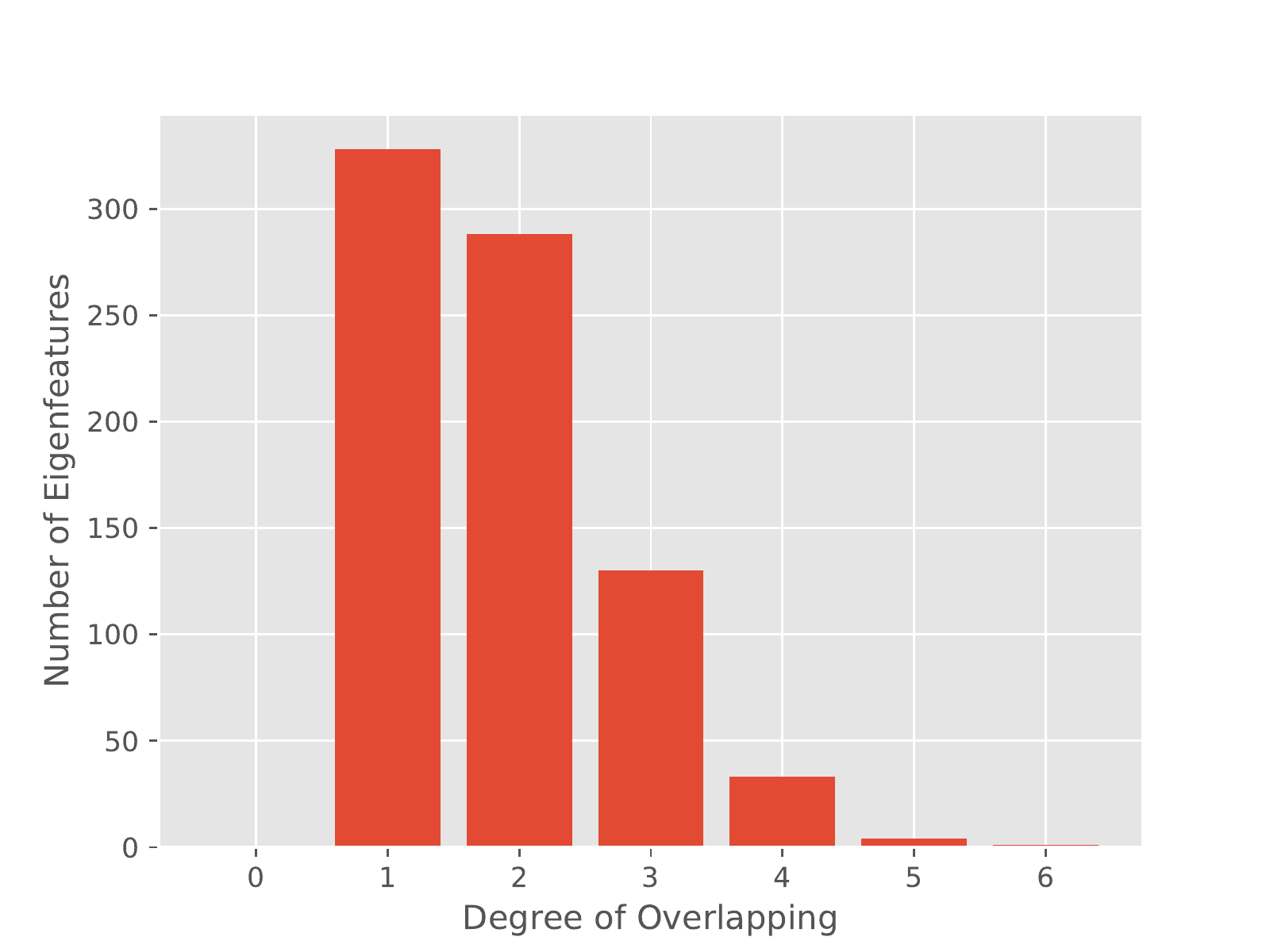}
\caption{Overlapping of classes on eigenfeatures of MNIST data set with approximated eigen component analysis (AECA).}
\label{fig:mnist_overlapping}
\end{figure}

\begin{figure}
  \centering
  \includegraphics[width=\linewidth]{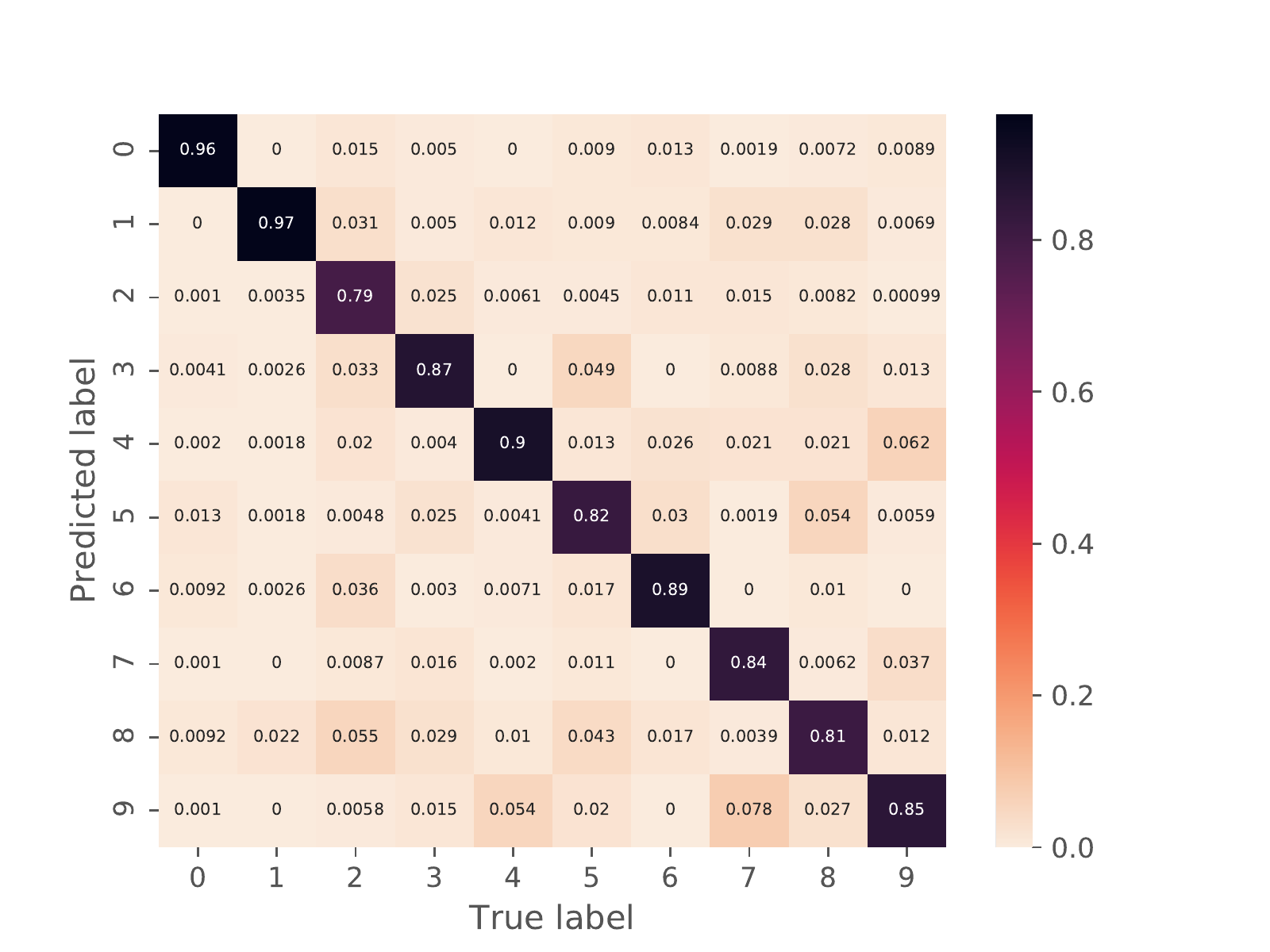}
  \caption{Confusion matrix of linear discriminant analysis (LDA) model on MNIST data set (Accuracy: 0.873)}
  \label{fig:mnist_cm_lda}
\end{figure}

\begin{figure}
  \centering
  \includegraphics[width=\linewidth]{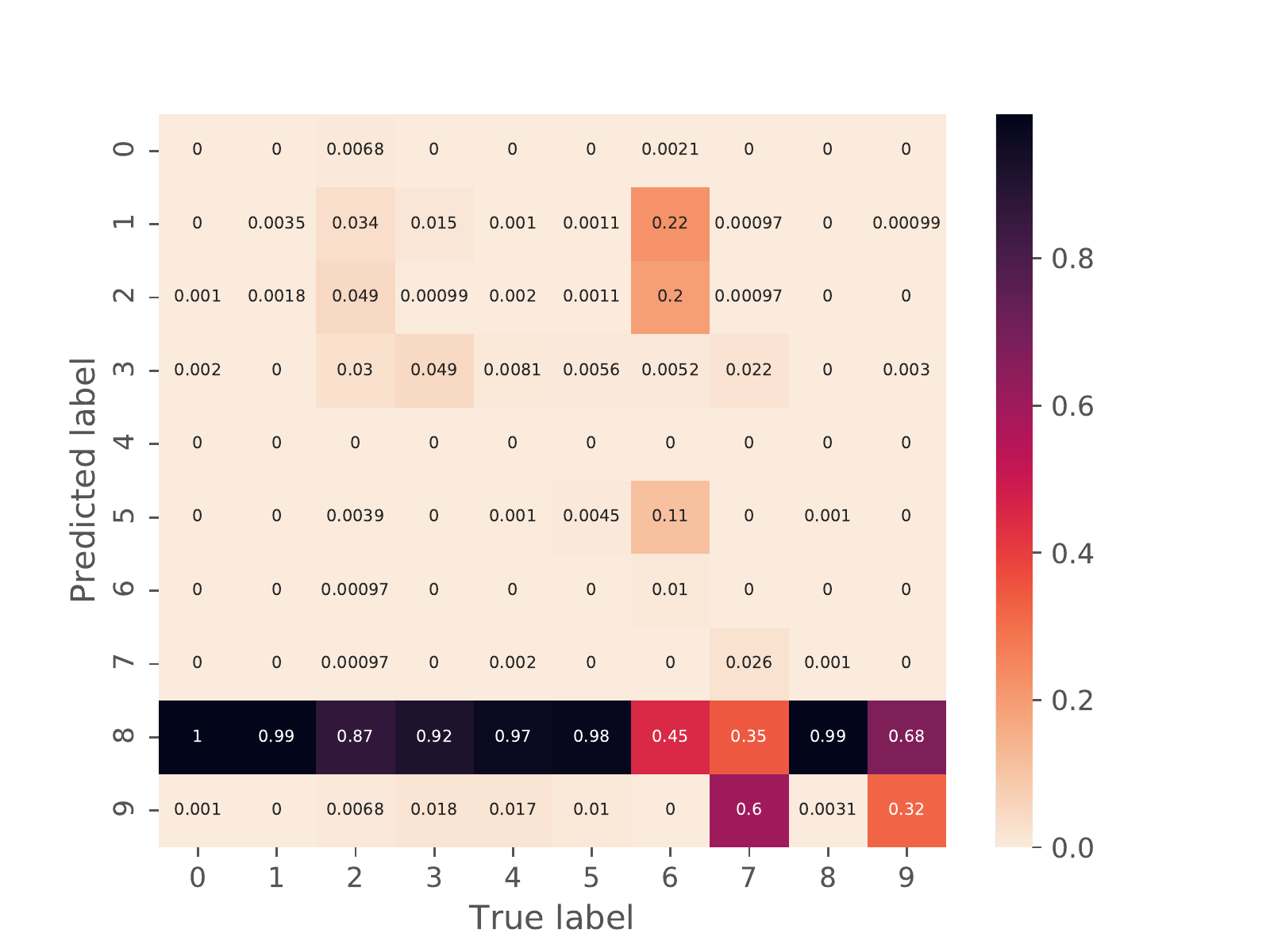}
  \caption{Confusion matrix of quadratic discriminant analysis (QDA) model on MNIST data set (Accuracy: 0.144)}
  \label{fig:mnist_cm_qda}
\end{figure}

\begin{figure}
  \centering
  \includegraphics[width=\linewidth]{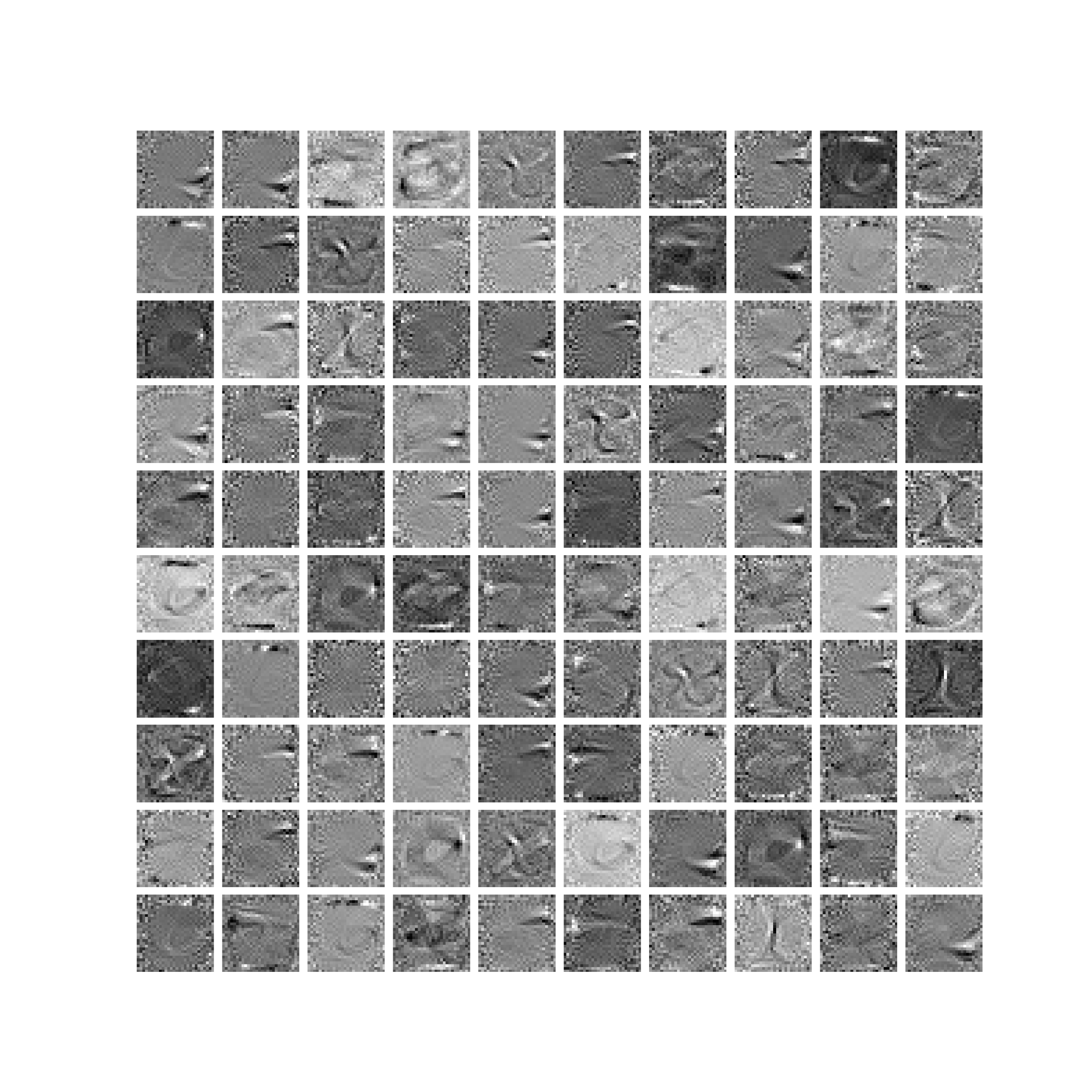}
  \caption{Some 100 pure eigenfeatures (PEs) randomly chosen from all 110 PEs learnt by vanilla eigen component analysis (VECA).}
  \label{fig:mnist_ber_nonoverlap}
\end{figure}

\begin{figure}
  \centering
  \includegraphics[width=\linewidth]{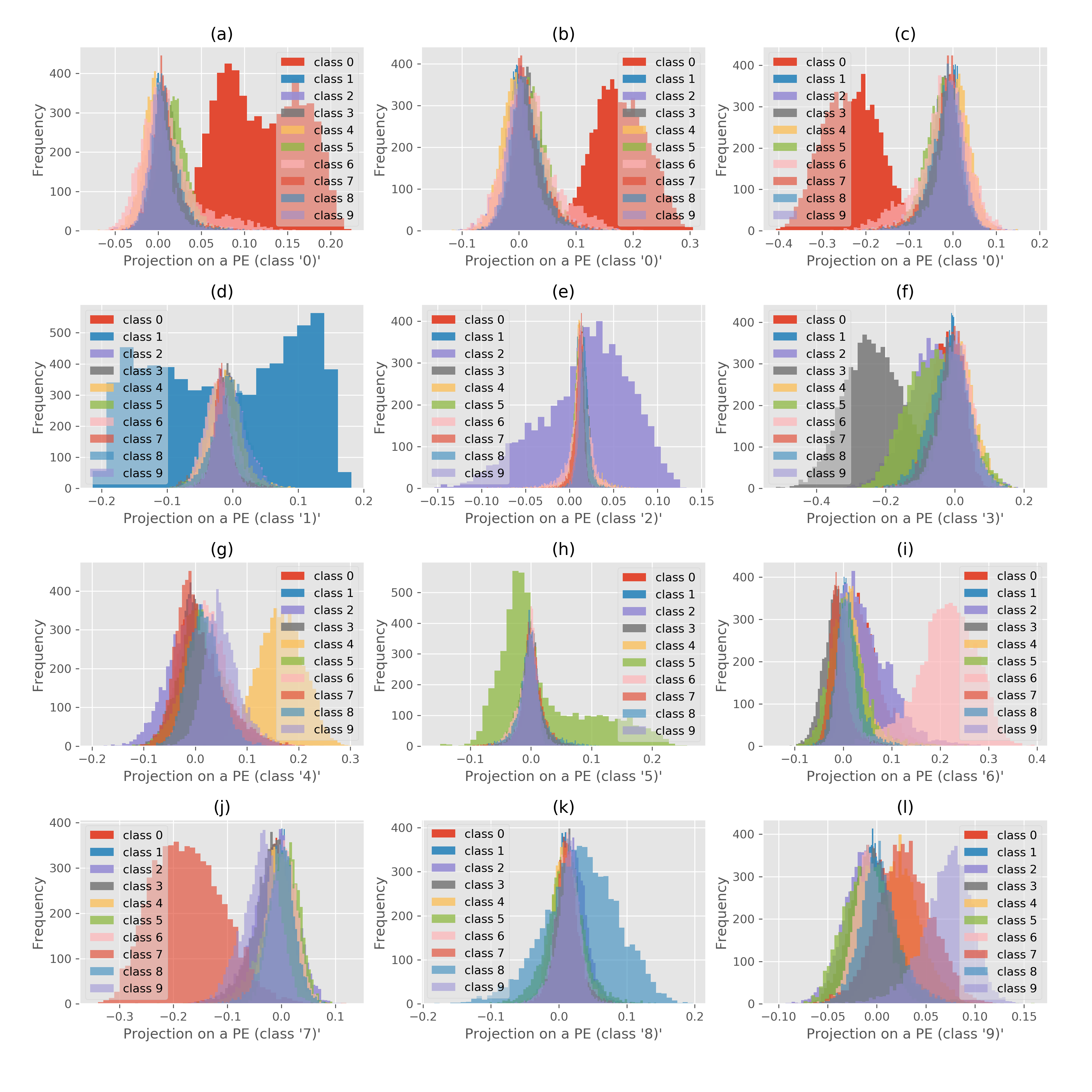}
  \caption{Distribution of projection on 12 pure eigenfeatures (PEs) belonging to all 10 classes with vanilla eigen component analysis (VECA) on MNIST dat set}
  \label{fig:mnist_ber_proj_freq_dist_ef}
\end{figure}

\begin{figure}
  \centering
  \includegraphics[width=\linewidth]{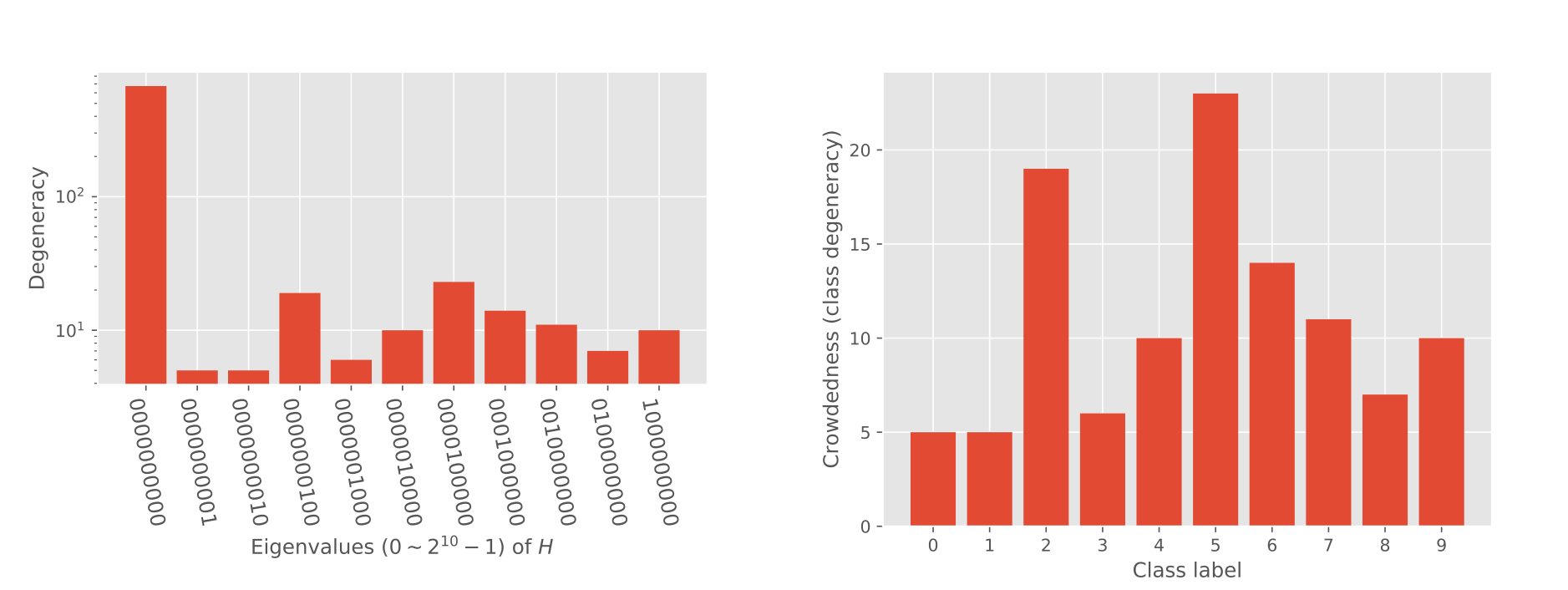}
  \caption{(a) Degeneracys of all eigenvalues of MNIST data set with vanilla eigen component analysis (VECA); (b) Crowdedness of classes of MNIST data set with VECA. }
  \label{fig:mnist_ber_degeneracy_and_crowdedness}
\end{figure}

\begin{figure}
  \centering
  \includegraphics[width=\linewidth]{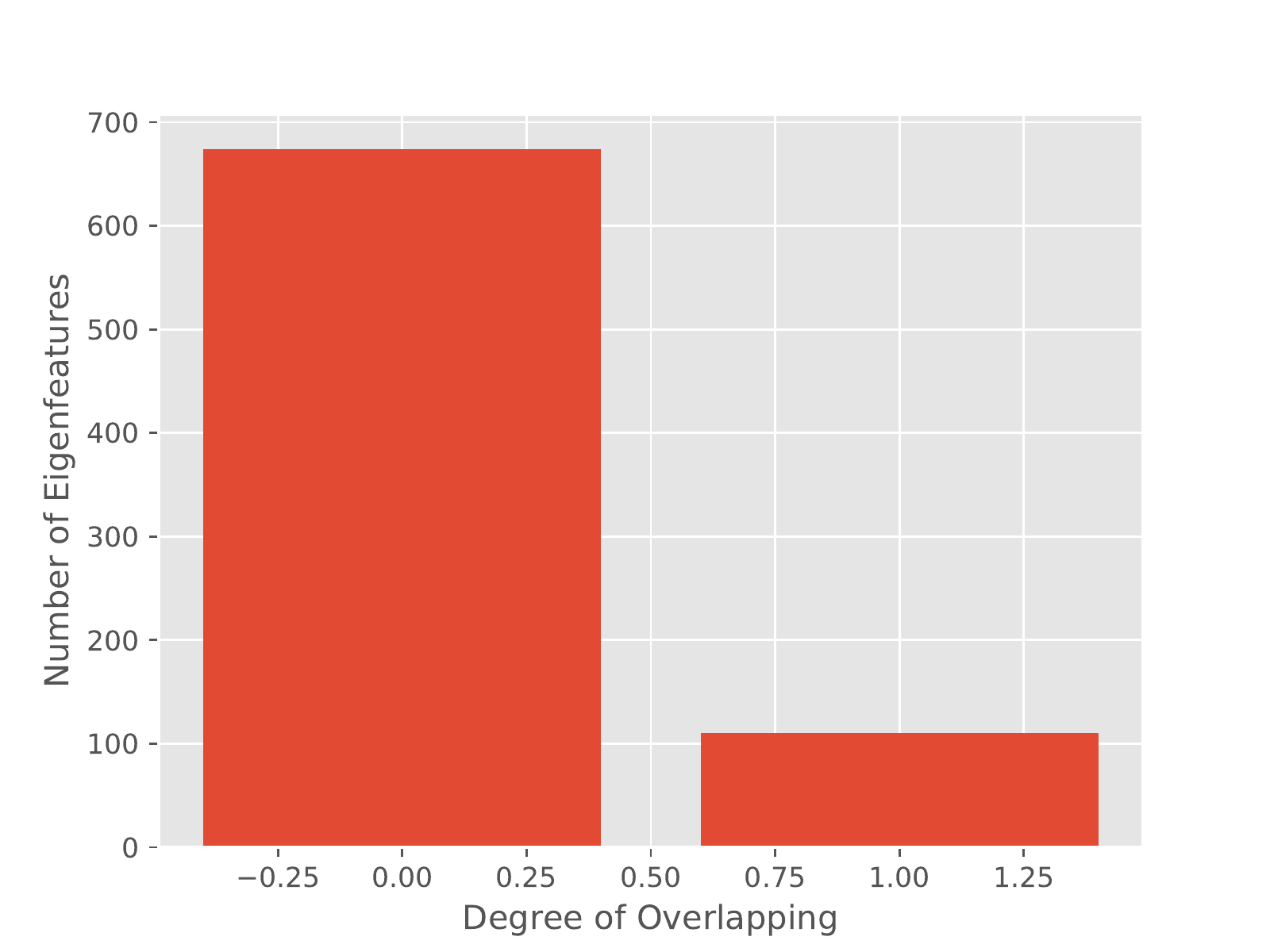}
  \caption{Overlapping of eigenfeature of MNIST data set with vanilla eigen component analysis (VECA). Only $0^o$ and $1^o$ eigenfeatures are learnt.}
  \label{fig:mnist_ber_overlapping}
\end{figure}


\section{Categorical distribution assumption of class label given eigenfeature}

The \gls{pmf} of class label given eigenfeature could be assumed as a categorical distribution. We also drop the winner-take-all rule. With \Cref{equ:single_class_pmf_no_rounded}, the new \gls{pmf} could be rewritten as 

\begin{equation}
p(y|\ket{\mathsf{x}}) = \sum_{\lambda} {p_{\lambda}(y|\ket{\lambda})} p({\lambda}|\ket{\mathsf{x}})
\end{equation}

Therefore, our objective becomes

\begin{equation}
\begin{split}
\argmin_{P, L} \quad & -\sum_{i=1}^{n} \mathsf{y^{(i)T}} \log \{ [({\bra{\mathsf{x}^{(i)}}} P)^* \odot ({\bra{\mathsf{x}^{(i)}}} P)] \Xi(L) \}^T \\
& + {\xi \|I - (P^{\dagger}P)^* \odot (P^{\dagger}P)\|^2_F} \\ 
\end{split}
\end{equation}

in which $\Xi(\cdot)$ is row-wise softmax function on each row of $L_{m \times l}$.


Furthermore, we could train the amplitude and phase together such that

\begin{equation}
\begin{split}
\argmin_{P, L} \quad & -\sum_{i=1}^{n} \mathsf{y^{(i)T}} \log \{ [({\bra{\mathsf{x}^{(i)}}} P)^* \odot ({\bra{\mathsf{x}^{(i)}}} P)] [\Xi(\Gamma) \odot \Xi(L)] \}^T \\
& + {\xi \|I - (P^{\dagger}P)^* \odot (P^{\dagger}P)\|^2_F} \\ 
\end{split}
\end{equation}

in which $\Gamma$ is the \gls{fnn} output of each projection on eigenfeatures.  $L$ is the global information of how an eigenfeature is mapped to each class label. However $\Gamma$ is the local information of how the projection of an input vector on eigenfeatures is mapped to each class label. If we ignore the global information and make it embed into the local information, our objective could be simplified as 

\begin{equation}
\begin{split}
\argmin_{P, L} \quad & -\sum_{i=1}^{n} \mathsf{y^{(i)T}} \log \{ [({\bra{\mathsf{x}^{(i)}}} P)^* \odot ({\bra{\mathsf{x}^{(i)}}} P)] \Xi(\Gamma) \}^T \\
& + {\xi \|I - (P^{\dagger}P)^* \odot (P^{\dagger}P)\|^2_F} \\ 
\end{split}
\end{equation}



\section{Mean squared error (MSE) loss}
If we define our objective as minimizing \gls{mse} loss, then our objective of \gls{aeca} could be

\begin{equation}
\begin{split}
\argmin_{P, L} \quad & \sum_{i=1}^{n} \| \mathsf{y^{(i)T}} - [(\bra{\mathsf{x}^{(i)}} P)^* \odot (\bra{\mathsf{x}^{(i)}} P)] \dot{L} \|^2_2 + \\
& {\xi \|I - (P^{\dagger}P)^* \odot (P^{\dagger}P)\|^2_F} + \\ 
& \gamma \|\mathbb{1}_{m \times l}+\cos(2L)\|^2_F.
\end{split}
\end{equation}

The problem of low converge rate of \gls{mse} loss also exists in our model. 


\section{Gradient rule of eigen component analysis (ECA)}
With the objective of maximizing likelihood, let 
\begin{equation}
\begin{split}
\mathcal{L}_{\mathbb{C}}(P, L) = 
& - \sum_{i=1}^{n} \big\{ \mathsf{y^{(i)T}} \log \{ [(\bra{\mathsf{x}^{(i)}} P)^* \odot (\bra{\mathsf{x}^{(i)}} P)] \dot{L} \}^T \\
& + ( \mathbb{1}_{l \times 1} - \mathsf{y^{(i)T}}) \log \{ [(\bra{\mathsf{x}^{(i)}} P)^* \odot (\bra{\mathsf{x}^{(i)}} P)] (\mathbb{1}_{m \times l} - \dot{L}) \}^T \big\} \\
& + {\xi \|I - (P^{\dagger}P)^* \odot (P^{\dagger}P)\|^2_F} \\ 
& + \gamma \|\mathbb{1}_{m \times l}+\cos(2L)\|^2_F
\end{split}
\end{equation}

and

\begin{equation}
\begin{split}
\mathcal{L}_{\mathbb{R}}(P, L) = 
& -\sum_{i=1}^{n} \big\{ \mathsf{y^{(i)T}} \log \{ [({\mathsf{x}^{(i)}}^T P) \odot ({\mathsf{x}^{(i)}}^T P)] \dot{L} \}^T \\
& + ( \mathbb{1}_{l \times 1} - \mathsf{y^{(i)T}}) \log \{ [({\mathsf{x}^{(i)}}^T P) \odot ({\mathsf{x}^{(i)}}^T P)] ( \mathbb{1}_{m \times l} - \dot{L} ) \}^T \big\} \\
& + {\xi \|I - P^{T}P\|^2_F} \\ 
& + \gamma \|\mathbb{1}_{m \times l}+\cos(2L)\|^2_F.
\end{split}
\end{equation}

\subsection{Gradient rule on real ECA}


The gradient of $\mathcal{L}_{\mathbb{R}}$ on $L$ is 

\begin{equation}
\begin{split}
\nabla_{L} {\mathcal{L}_{\mathbb{R}}} 
= & -\sum_{i=1}^{n} \big\{ 2 [({\mathsf{x}^{(i)T}}P) \odot ({\mathsf{x}^{(i)T}}P)]^T \{\mathsf{y}^{(i)T} \odot \frac{1}{[({\mathsf{x}^{(i)T}}P) \odot ({\mathsf{x}^{(i)T}}P)]{L}}\} \frac{\partial{\dot{L}}}{\partial{{L}}} \\ 
& - [({\mathsf{x}^{(i)T}}P) \odot ({\mathsf{x}^{(i)T}}P)]^T \{\mathbb{1}_{1 \times l} \odot \frac{1}{[({\mathsf{x}^{(i)T}}P) \odot ({\mathsf{x}^{(i)T}}P)]{L}}\} \frac{\partial{\dot{L}}}{\partial{{L}}} \big\}  \\ 
& -4\gamma(\mathbb{1}_{m \times l}+\cos(2L)) \odot \sin(2L) \\
\end{split}
\end{equation}

and on $P$ is

\begin{equation}
\begin{split}
\nabla_{P} {\mathcal{L}_{\mathbb{R}}} = & -2\sum_{i=1}^{n} \big\{ [\mathsf{x}^{(i)T} \odot ({\mathsf{x}^{(i)T}}P)]^T \{\mathsf{y}^{(i)T} \odot \frac{1}{[({\mathsf{x}^{(i)T}}P) \odot ({\mathsf{x^{(i)T}}}P)]{L}}\}{L}^T \\
& + [\mathsf{x}^{(i)T} \odot ({\mathsf{x}^{(i)T}}P)]^T ( \mathbb{1}_{l \times 1} - \mathsf{y^{(i)T}}) \odot \frac{1}{[({\mathsf{x}^{(i)T}}P) \odot ({\mathsf{x^{(i)T}}}P)]{L}}\} ( \mathbb{1}_{m \times l} - \dot{L} ) \big\} \\
& -4\xi(I - P^T P) \odot P.
\end{split}
\end{equation}

Besides, 

\begin{equation}
\begin{split}
\frac{\partial{\dot{L}}}{\partial{{L}}} 
= & \chi \omega \cos(\omega L) \odot \dot{L} \odot (\dot{L} - 1) \odot L.\\
\end{split}
\end{equation}




\subsection{Gradient rule on complex ECA}


Since $\mathcal{L}_{\mathbb{C}}$ is a real function of real matrix $L$,  the gradient on $L$ is the same as the real ECA. As $\mathcal{L}_{\mathbb{C}}$ is a real function of a complex matrix $P$, the gradient on $P$ could be taken on the real part and imaginary part of $P$ individually, such that

\begin{equation}
\begin{split}
\nabla_P {\mathcal{L}_{\mathbb{C}}}
= \frac{\partial{\mathcal{L}}}{\partial{\text{Re}(P)}} + i\frac{\partial{\mathcal{L}}}{\partial{\text{Im}(P)}}.
\end{split}
\end{equation}







\section{Experiment of complex implementation on MNIST data set}
We implemented one complex version of this model. As we pointed out, all the projection only happen in real coordinate space, thus it showes no edge over the real model. All the source code could be download from \url{https://github.com/chenmiaomiao/eca/}





\section{Separating amplitude and phase together}

  \subsection{Raising dimension operator on projections}

  For a \gls{rado} $A$, it could be defined as 
  \begin{equation}
  A_{l}x = \begin{bmatrix} 1\\x\\x^2\\\vdots\\x^l\end{bmatrix}
  \label{eq:one_proj_rasing_dim}
  \end{equation}


  \subsubsection{Fragment neural networks (FNNs)}


  These amplitudely varied features would map on more than one classes. We define $m$ \glspl{fnn} $\Gamma(\cdot)$ each of which would output a $l$ length vector with input one or no more than $m$ projections of original vector on each eigenfeature. The input of \gls{fnn} should be based upon the chracteristics of the data set. If there are some remained amplitude difference and those difference are relatively independent, one projection is enough. Otherwise, those amplitude difference are corelated or nonlinearly related. Then an fine and intricate \gls{fnn} should be designed. One could use the \glspl{pe} as input or $2^o$ or higher overlapping eigenfeatures as input. The simplest \gls{fnn} asuume these projections are independent with one projection as input. For one projection, we use aforementioned \gls{rado} (see \Cref{eq:one_proj_rasing_dim}), such that 

  \begin{equation}
  \Gamma_{jk} = \vartheta_k(A(\bra{\mathsf{x}^{(i)}} P_j))=\vartheta_{k0}+\sum_{s=1}^{l} \vartheta_{ks}(\bra{\mathsf{x}^{(i)}} P_j)^s.
  \end{equation}


  \subsection{LoR or SR (Bernoulli or categorical distribution)}
  Each class label of the amplitude of each eigenfeature could be assumed drawn from $l$ independent bernoulli distribution. We could also assume the class label of each eigenfeature follows categorical distribution. With the former assumption we would use \gls{lor}. The latter assumption could be estimated with \gls{sr}.
  
  \subsection{Model fusion}

  Firstly, we define \glspl{fnn} $\Gamma(\cdot)$ on each projection of a vector on eigenfeature. 

  To tell the difference on amplitude, when one eigenvalue has been mapped onto several classes, we could add a softmax function $\Xi$ on each output of $\Gamma(\cdot)$. 

  \begin{equation}
  \mathbf{p}(\mathsf{y}^{(i)}|\ket{\mathsf{x}^{(i)}}; \theta) = [({\bra{\mathsf{x}^{(i)}}} P)^* \odot ({\bra{\mathsf{x}^{(i)}}} P)] [\dot{\Gamma} \odot \dot{L}].
  \end{equation}

  The amplitude difference could also be assumed as independent Bernoulli distribution, such that the objective is

  \begin{equation}
  \begin{split}
  \argmin_{P, L} \quad & - \sum_{i=1}^{n} \big\{ \mathsf{y^{(i)T}} \log \{ [({\bra{\mathsf{x}^{(i)}}} P)^* \odot ({\bra{\mathsf{x}^{(i)}}} P)] [\dot{\Gamma} \odot \dot{L}] \}^T \\
  & + (\mathbb{1}_{1 \times l} - \mathsf{y^{(i)T}}) \log \{ [({\bra{\mathsf{x}^{(i)}}} P)^* \odot ({\bra{\mathsf{x}^{(i)}}} P)] [\mathbb{1}_{m \times l} - \dot{\Gamma} \odot \dot{L}] \}^T \big\} \\
  & + {\xi \|I - (P^{\dagger}P)^* \odot (P^{\dagger}P)\|^2_F} \\ 
  & + \gamma \|\mathbb{1}_{m \times l}++\cos(2L)\|^2_F
  \end{split}
  \end{equation}

  To train this model, it's tricky to balance the $\dot{\Gamma}$ and $\dot{L}$. One solution is trainning $P$ and $\dot{L}$ firstly and then set them fixed. Then we train the $\dot{\Gamma}$ part. Instead learn $\Gamma$ and $L$ separately, we could learn $\Gamma$ and $L$ together, such that the objective is

  \begin{equation}
  \begin{split}
  \argmin_{P, L} \quad & - \sum_{i=1}^{n} \big\{ \mathsf{y^{(i)T}} \log \{ [({\bra{\mathsf{x}^{(i)}}} P)^* \odot ({\bra{\mathsf{x}^{(i)}}} P)] \dot{\Gamma} \}^T \\
  & + (\mathbb{1}_{1 \times l} - \mathsf{y^{(i)T}})  \log \{ [({\bra{\mathsf{x}^{(i)}}} P)^* \odot ({\bra{\mathsf{x}^{(i)}}} P)] ( \mathbb{1}_{m \times l} - \dot{\Gamma} \}^T \big\} ) \\
  & + {\xi \|I - (P^{\dagger}P)^* \odot (P^{\dagger}P)\|^2_F} \\ 
  & + \gamma \|\mathbb{1}_{m \times l}+\cos(2L)\|^2_F
  \end{split}
  \end{equation}

  Besides, we could assume the class label of each eigenfeature follows categorical distribution. 
  To tell the difference on amplitude, when one eigenvalue has been mapped onto several classes, we could add a row-wise softmax function $\Xi$ on each row of the output of $\Gamma(\cdot)$. 

  \begin{equation}
  \mathbf{p}(\mathsf{y}^{(i)}|\ket{\mathsf{x}^{(i)}}; \theta) = [({\bra{\mathsf{x}^{(i)}}} P)^* \odot ({\bra{\mathsf{x}^{(i)}}} P)] [\Xi(\Gamma) \odot \dot{L}].
  \end{equation}

  Then our objective is 

  \begin{equation}
  \begin{split}
  \argmin_{P, L} \quad & -\sum_{i=1}^{n} \big\{ \mathsf{y^{(i)T}} \log \{ [({\bra{\mathsf{x}^{(i)}}} P)^* \odot ({\bra{\mathsf{x}^{(i)}}} P)] [\Xi(\Gamma) \odot \dot{L}] \}^T \\
  & + ( \mathbb{1}_{1 \times l} - \mathsf{y^{(i)T}} ) \log \{ [({\bra{\mathsf{x}^{(i)}}} P)^* \odot ({\bra{\mathsf{x}^{(i)}}} P)] [\mathbb{1}_{m \times l} - \Xi(\Gamma) \odot \dot{L}] \}^T \big\} \\
  & + {\xi \|I - (P^{\dagger}P)^* \odot (P^{\dagger}P)\|^2_F} \\ 
  & + \gamma \|\mathbb{1}_{m \times l}+\cos(2L)\|^2_F
  \end{split}
  \end{equation}

  Actually, the row-wise softmax function could be designed more flexiblely. \ignore{Instead of imposing this function on one, several or all elements of the input, it could be designed to dynamicly. } We could conduct a two-step training. Firstly, we find the \gls{efm} $P$ and \gls{ecmm} $\dot{L}$. Then to distinguish these amplitudely varied features we could package these overlapped (2 or more degree of overlapping) eigenfeatures which has the same eigenvalues. Lastly, these packaged features are then fed into an \gls{fnn} following by a softmax function. Meantime, we could raise dimension on these amplitude varied features. 






\section{Kernel eigen component analysis (KECA), working with nonlinear models}
\label{app:ext_eca}

  \subsection{KECA}
  \Gls{keca} could be easily implemented as \gls{eca} are composed of a series of inner product.

  For a polynomial kernel, 

  \begin{equation}
  K_p(\mathsf{x}, \mathsf{x}') = (1 + \braket{\mathsf{x}|\mathsf{x}'})^d.
  \end{equation}

  For a \gls{rbf} kernel, 

  \begin{equation}
  K_{RBF}(\mathsf{x}, \mathsf{x}') = e^{-\|\mathsf{x}-\mathsf{x}'\|_2^2}.
  \end{equation}

  The the real \gls{keca} objective could be
  \begin{equation}
  \begin{split}
  \argmin_{P, {L}} & - \sum_{i=1}^{n} \big\{ \mathsf{y}^{(i)T} \log \{[K({\mathsf{x}^{(i)}}, P) \odot K({\mathsf{x}^{(i)}}, P)] \dot{L} \}^T \\
  & + (\mathbb{1}_{1 \times l} - \mathsf{y}^{(i)T}) \log \{[K({\mathsf{x}^{(i)}}, P)  \odot K({\mathsf{x}^{(i)}}, P) ] (\mathbb{1}_{m \times l} - \dot{L}) \}^T \big\} \\
  & + {\xi \|I - K(P,P)\|^2_F} \\ 
  & + \gamma \|\mathbb{1}_{m \times l}+\cos(2L)\|^2_F.
  \end{split}
  \end{equation}















\section[Non-matrix form of H, ininfite dimensional x, ceca]{Non-matrix form of $H$, ininfity dimensional $\ket{x}$ and continuous eigen component analysis (CECA)}
\label{app:ceca}


We could also build a generic \gls{eca} model which the state $\ket{x}$ and measurable and $H$ could be matrix or function. A generic \gls{eca} model could work on continuous domain. 

\begin{figure}
  \centering
  \includegraphics[width=\linewidth]{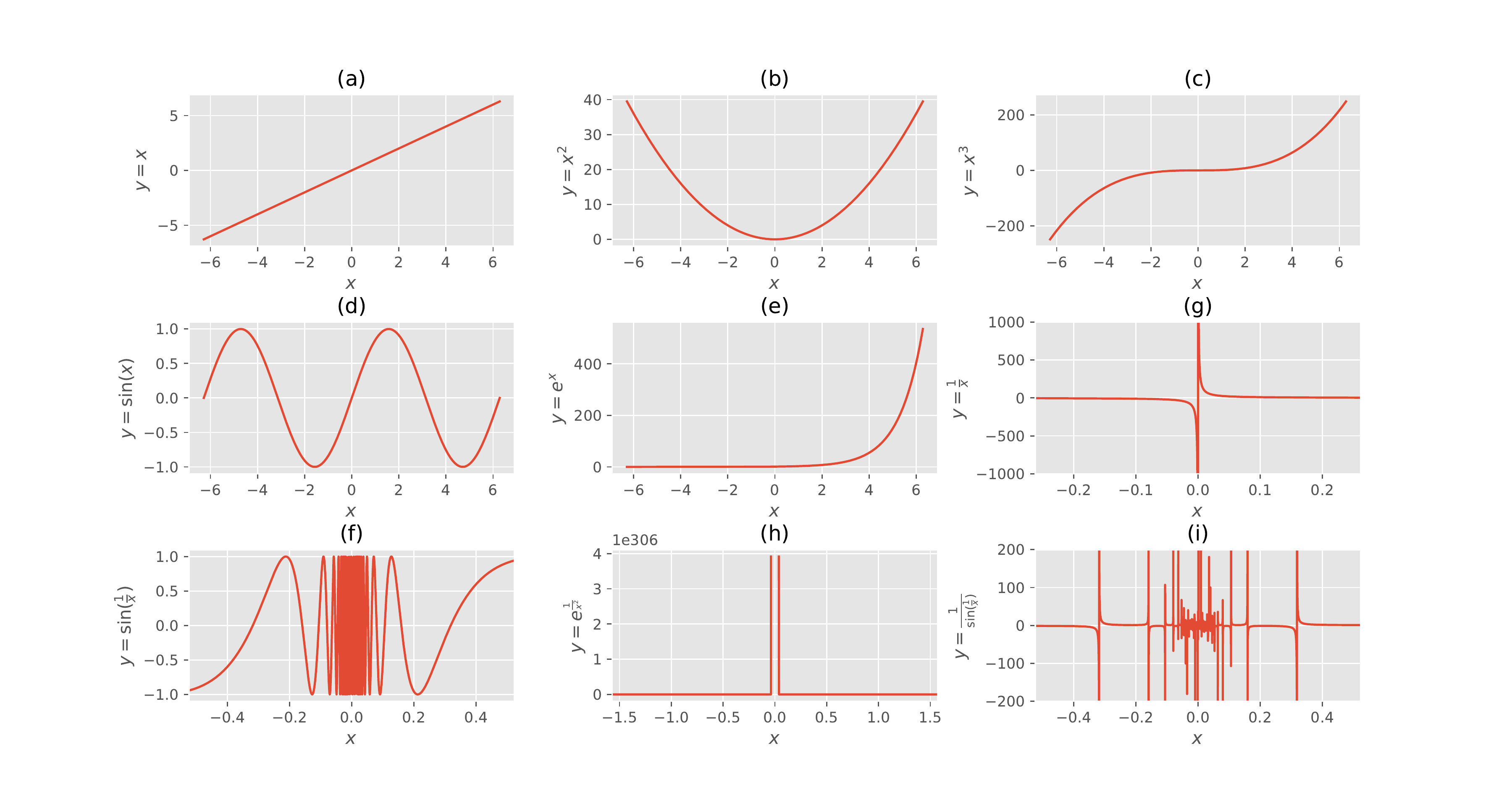}
  \caption{Function to be learnt by CECA}
  \label{fig:ceca_to_learn}
\end{figure}

Unlike the development of vanlila \gls{eca}, most of the output are trival except some limited spikes (\Cref{fig:ceca_to_learn} (i)) that we care about in \gls{ceca}. For learning these \gls{1d} continuous function ((\Cref{fig:ceca_to_learn}) (a)-(g)) on state $\ket{x}$ (which is an infintity dimension vector), it's actually an infinite dimension \gls{eca} problem. The outputs or eigenvalues of these problems are infinite in continuous domain.


For a $q$-dimensional data, the \gls{ceca} assumes a $\infty$-dimensional transformation $A_{\infty}$, such that

\begin{equation}
HA_{\infty \times q}\mathsf{x}=yA_{\infty \times q}\mathsf{x}
\end{equation}

And the observed value is the expectation of observed value. The expectation of $H$ on a given state is

\begin{equation}
\hat{y}=\braket{H} = \braket{\mathsf{x}|A^{\dagger}HA|\mathsf{x}}
\end{equation}

\section{More remarks, tips and tricks of eigen component analysis (ECA)}
  \subsection{Introducing degeneracy as redundancy}

  Two sum up all the probabilities on all these degenerated eigenfeatures, we avoid transform these features nonlinearly. This could be used in any model to increase the linearity and avoid overfitting. For a one layer neural network $N$ with $\Sigma$ units, it could be defined as 

  \begin{equation}
  N_{\Sigma \times 1} = \overline{\sigma{(W_{\Sigma\times m}\mathsf{x}_{m\times1} + b)}}
  \end{equation} 

  with probability normalization operator $\overline{(\cdot)}$ or 

  \begin{equation}
  N_{\Sigma \times 1} = S{(W_{\Sigma\times m}\mathsf{x}_{m\times1} + b)}
  \end{equation} 

  in which  $\sigma(\cdot)$ and $S(\cdot)$ are the sigmoid function and softmax function respectively. 

  Then we could view $N_{\Sigma \times 1}$ as probability. With a degeneracy operator $\mathfrak{L}_{\Sigma \times l}$, the combined \gls{pmf} of the class label given $\mathsf{x}$ could be

  \begin{equation}
  \mathbb{p}(\mathsf{y}|\mathsf{x}; \theta)= \mathfrak{L}_{l \times \Sigma}^T N_{\Sigma \times 1}
  \end{equation} 

  The crowdedness (i.e. class degeneracy) on each class could determine the stability of a model. If one class is more crowded, it implys a complex structure for the pattern in that class. The lower crowdedness might imply a simpler structure of that class which also might lead to higher accuracy in prediction. If the variance of crowdedness on every class is smaller, then the model would be more stable. 

  \subsection[Learning a sparse or a dense L]{Learning a sparse or a dense $\dot{L}$}
  From the development of \gls{veca}, we know that it should learn a sparse $\dot{L}$. More spase of the $\dot{L}$, the meaning of the eigenfeature becomes more abstract. If $\dot{L}$ is dense, then the higher overlapping will degenerate the performance of the model. Using \gls{aeca}, we could learn a balanced sparsity of $\dot{L}$.

  To learn a more sparse $\dot{L}$ for \gls{eca}, especially \gls{aeca}, we could also make our model be less sensitive to weak eigenfeatures or weak mapping between eigenfeature and class and ignore it. We could use a $\ell_0$ regularization in our objective. As the elements of $L$ are all positive, the $\ell_1$ is equivalent to the expectation of the elements, which is

  \begin{equation} 
  \mathbb{E}[\dot{L}_{ij}]
  \end{equation}

  As the elements of $L$ being concentrated on 0 or 1, $\ell_2$ regularization is enough to guarantee sparsity which is

  \begin{equation} 
  \|\dot{L}\|^2_F
  \end{equation}

  and we found $\ell_2$ works better than $\ell_1$ in practice. 


  Also, we could assume the class label of eigenfeatures follow a categorical distribution. Then each row of $\mathfrak{L}$ would have at most one $1$. But the round operation on $\dot{L}$ might be risky. It would be harder to make these probabilities be asymptote to $0$ or $1$. Anyway, it's still possible to work for classification but tricky to explain. Without overlapping, it becomes less consistent with the real world scenario. 

  \subsection{Relaxing the constraint on L}
  The constraints on $\dot{L}$ assume an eigenfeature unambiguously belongs to one class. Once we discard these constraints, it gives our model flexibility. We might obtain greater performance but less interpretable model. Meantime, it would influence the stability of the whole model and overfit the model. 

  In practice, we found that even without the constraint of $L$ in \gls{veca} or \gls{aeca}, the model could still learn a $\dot{L}$ whose elements concentrate on $0$ or $1$.

  \subsection{Relaxing the constraint on P}

  Once we abandon the orthogonal constraints on $P$, these extracted features no longer unambiguously distinguishable. Furthermore, if we abandon the constraint on normality, these states lose the interpretation of probability. Then this model collapsed onto a common or somehow crippled neural network. However, because of the redundancy introduced by $\dot{L}$, the model still could obtain higher performance. 

  Furthermore, the dimensions of $P$ could be more flexible once we drop the orthogonality constraint. We could define a non-orthogonal feature matrix 

  \begin{equation}
  P_{m \times {M}}, {M} > m
  \end{equation}

  with normalized column vector and its corresponding degeneracy operator

  \begin{equation}
  \mathfrak{L}_{{M} \times l}, {M} > m.
  \end{equation}

  \subsection{Learning on super-imbalanced data set}
  With maximizing the variance on the projections on eigenfeatures and learning a sparse $\dot{L}$, we could learn on a super-imbalanced data set. The variance could be calculated as 

  \begin{equation}
  \text{Var}(\braket{\mathsf{x}|P_j}) = \mathbb{E}[\braket{\mathsf{x}|P_j}^2] - \mathbb{E}^2[\braket{\mathsf{x}|P_j}].
  \end{equation}  


  \subsection{Multimodal distribution}
  The distributions of projections on eigenfeature (see \Cref{fig:mnist_proj_dist_ef}) are multimodal in the MNIST data set. However, with \gls{eca}, we don't have to learn these multimodal distribution explicitly. In \gls{geca}, we assume these distributions still be normal distribution to approximate these multimodal distributions. 

  \subsection{Eigen component analysis (ECA), dictionary learning (DictL) and and dimension reduction}
  With \gls{eca}, especially \gls{veca}, we could learn a dictionary. The atoms of this dictionary are these \glspl{pe}. 

  \subsection{Deep neural networks (DNNs) and eigen component analysis (ECA)}
  Actually, the softmax function together with \gls{dnn} could be interpreted as having an implicit \gls{ecmm} (\Cref{equ:dnn_softmax}) which  is the identity matrix

  \begin{equation}
      \overset{\asymp}{\mathfrak{L}}_{implicit} = 
      I_{l\times l}.
      \label{equ:dnn_softmax}
  \end{equation}

  For a state $\mathsf{x}$, the goal of \gls{dnn} is to find a nonlinear function $N$ and a softmax function $S$ which could transform the $\mathsf{x} \in \mathbb{C}^m$ into probabilities $\mathsf{z} \in \mathbb{R}^l$. That could be described as  

  \begin{equation*}
  \mathsf{z} = S(N(\mathsf{x}))
  \end{equation*}

  and the vector with combined probabilities is

  \begin{equation*}
  \mathbf{p}(\mathsf{y}|\mathsf{x}; \theta) = (\mathsf{z}^T \overset{\asymp}{\mathfrak{L}}_{implicit})^T.
  \end{equation*}

  In the development of \gls{ecan}, we found that \glspl{dnn} could be generalized \glspl{ecan}.



\printglossaries

\end{document}